\documentclass[preprint,10pt,3p,authoryear]{elsarticle}
\usepackage{geometry}
\usepackage{times}
\usepackage{amssymb}
\usepackage{amsmath}
\usepackage{bm}
\usepackage{dsfont}   %
\usepackage{amsthm}
\usepackage{booktabs}
\usepackage{tabularx}
\usepackage{algorithm}
\usepackage{algpseudocode}
\usepackage{graphicx}
\usepackage[section]{placeins}   %
\usepackage{xcolor}
\usepackage[normalem]{ulem}
\DeclareRobustCommand{\rev}[1]{\ifvmode\leavevmode\fi{\color{red}#1}}   %
\DeclareRobustCommand{\revdel}[1]{\ifvmode\leavevmode\fi{\color{red}\sout{#1}}}   %
\usepackage[hidelinks]{hyperref}

\theoremstyle{plain}
\newtheorem{theorem}{Theorem}
\newtheorem{proposition}{Proposition}
\newtheorem{lemma}{Lemma}
\newtheorem{corollary}{Corollary}
\theoremstyle{definition}
\newtheorem{definition}{Definition}
\newtheorem{assumption}{Assumption}
\theoremstyle{remark}
\newtheorem{remark}{Remark}

\newcommand{\acrinit}[1]{\underline{\textbf{#1}}}

\newcommand{\obs}{o}
\newcommand{\bev}{\bm{m}}
\newcommand{\ego}{\bm{s}}
\newcommand{\act}{\bm{a}}
\newcommand{\lat}{\bm{z}}
\newcommand{\latp}{\hat{\bm{z}}}
\newcommand{\mem}{\mathcal{M}}
\newcommand{\sg}{\operatorname{sg}}
\newcommand{\symlog}{\operatorname{symlog}}
\newcommand{\symexp}{\operatorname{symexp}}
\newcommand{\twohot}{\operatorname{twohot}}

\newcommand{\clip}{\operatorname{clip}}
\newcommand{\relu}{\operatorname{relu}}
\newcommand{\horizon}{H}
\newcommand{\ctx}{C}
\newcommand{\E}{\mathbb{E}}
\newcommand{\R}{\mathbb{R}}

\journal{Transportation Research Part C: Emerging Technologies}

\begin{document}

\begin{frontmatter}

\title{SAFER: Navigating the Exploration--Exploitation Dilemma in\\ World-Model-Based Planning for Autonomous Driving}

\author[a]{Yansong Qu}
\ead{qu120@purdue.edu}
\author[b]{Zilin Huang}
\ead{zilin.huang@wisc.edu}
\author[b]{Zihao Sheng}
\ead{zihao.sheng@wisc.edu}
\author[b]{Sikai Chen\corref{cor1}}
\ead{sikai.chen@wisc.edu}
\cortext[cor1]{Corresponding author.}
\author[a]{Samuel Labi}
\ead{labi@purdue.edu}
\affiliation[a]{organization={Lyles School of Civil and Construction Engineering, Purdue University},
            citysep={}, city={West Lafayette}, postcode={47907}, country={USA}}
\affiliation[b]{organization={Department of Civil and Environmental Engineering, University of Wisconsin--Madison},
            citysep={}, city={Madison}, postcode={53706}, country={USA}}

\begin{abstract}
An autonomous vehicle must foresee the consequences of its actions, especially in the rare,
time-critical situations where crashes happen. A world model offers this foresight, but only within the coverage of its training
data, which poses an exploration--exploitation dilemma. Within this coverage a planner can hardly improve on the
data, and beyond it the model's errors mislead it. Logs of competent driving rarely contain the failures that would
widen the coverage, and agents that learn from interaction gain them only by failing themselves, without being
shown what to do instead. We propose \acrinit{SAFER}, a framework that learns offline from failures that
others make, judged by an expert, and keeps learning online from its own driving. Offline, partially trained agents and scripted, recoverable
perturbations \acrinit{S}tage failures in simulation, and an expert \acrinit{A}nnotates every visited state with
its action and value, so that much of what the planner explores is covered before it drives. In planning, a
chunked latent world model \acrinit{F}oresees the consequences fast enough for real-time use, and a risk-aware
 sampling-based planner \acrinit{E}valuates candidate maneuvers by their return minus a bounded collision penalty. Online, a
loop continuously \acrinit{R}efines the model wherever this driving leaves the coverage of the offline data. In closed-loop
evaluation against the compared methods, SAFER reaches the highest route completion and reward per step on the CARLA Leaderboard routes,
$0.95$ and $0.77$ from a privileged bird's-eye view, and the highest episode reward in MetaDrive, $53.4$ in regular and
$49.0$ in safety-critical traffic. Project page: \url{https://ys-qu.github.io/safer-website}.
\end{abstract}


\end{frontmatter}

\section{Introduction}
\label{sec:intro}

Autonomous vehicles must handle rare, time-critical
interactions in which a late or wrong decision cannot be undone, the situations in which crashes happen. Learning-based driving systems have made rapid progress in
closed-loop driving, yet these situations remain the main obstacle to deploying them, because the data they learn
from contain few such situations and fewer of their consequences~\citep{li2026worldengine}.

Handling such a situation requires anticipating what an action will lead to before taking it. An
experienced driver does this before every maneuver, imagining how the next seconds will unfold and choosing
accordingly. A policy trained by imitation or by model-free reinforcement learning does not. It maps the current
observation directly to an action. A world model, in contrast, predicts how the scene will evolve under a candidate
sequence of actions, so that a planner can compare the imagined futures before committing to
one~\citep{ha2018worldmodels,hafner2020dreamer,hansen2024tdmpc2,levy2026simdist}.

\begin{figure}[t]
  \centering
  \includegraphics[width=\linewidth]{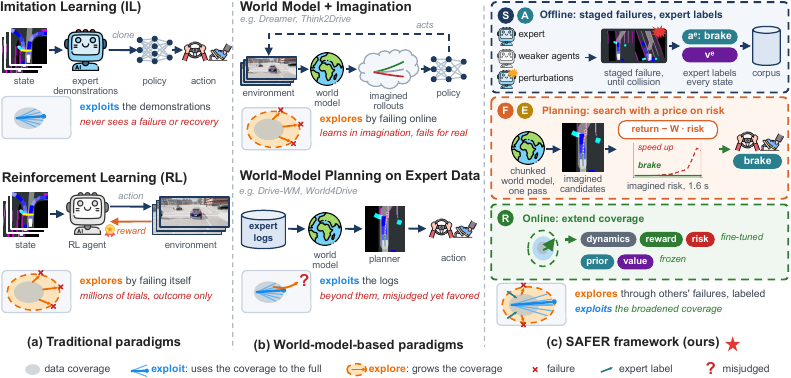}
  \caption{Learning paradigms for autonomous driving, compared by where the experience beyond competent driving
  comes from. The glyph under each paradigm shows the coverage of its data (gray) and how the paradigm
  treats it. Exploiting (blue) spreads the candidates over the coverage and takes the best one up to its edge, but
  no further. Exploring (orange) grows the coverage at the price of failures (red crosses), the paradigm's own or,
  for SAFER, failures that others make, labeled by the expert (teal). A step out of the coverage that does not
  grow it is misjudged (?). (a)~Imitation learning copies expert demonstrations and never observes a failure; reinforcement
  learning discovers consequences by failing in the environment. (b)~A world model learned
  online~\citep{hafner2020dreamer,li2024think2drive} still explores by failing. A world model learned from logged
  expert driving~\citep{wang2024drivewm,zheng2025world4drive} covers competent driving only, and a planner that
  searches beyond this coverage favors the actions that the model misjudges. (c)~SAFER broadens the coverage
  offline with staged failures that an expert annotates, plans with a fast chunked world model that prices
  collision risk, and extends the coverage online from the planner's own driving. The BEV frames, candidates and
  risk curves in (c) are from a recorded collision of the corpus (Fig.~\ref{fig:planner}a).}
  \label{fig:teaser}
\end{figure}

A world model, however, can only foresee what its training data contain; we call the states and actions that these
data represent its \emph{coverage}. This poses an exploration--exploitation dilemma for any planner that relies on a
world model (Fig.~\ref{fig:teaser}b). The dilemma is the choice between exploiting known options, whose outcomes are predictable, and
exploring new ones, which may pay off in the long run but whose outcomes are
uncertain~\citep{bergertal2014exploration,hills2015exploration,wilson2021balancing}. A planner that exploits what the model knows, choosing among the actions that the data cover, can rely
on the predictions but can hardly do better than the driving in the data. To do better, it can explore actions that
the data seldom contain, such as braking later or pulling out earlier than competent drivers do, and the states that
its own decisions lead to. There the predictions are unreliable, and because the planner selects the candidate with
the highest predicted value, it is drawn to exactly the candidates whose value the model
overestimates~\citep{chua2018pets,janner2019mbpo,yu2020mopo}. Experience of these actions is what corrects the model. Gathering such experience is the role of exploration in reinforcement
learning~\citep{curi2020hucrl,sekar2020plan2explore}, but in driving the experience beyond competent driving is the
experience of failure, which cannot be gathered on public roads.

Existing approaches differ in where this experience comes from (Fig.~\ref{fig:teaser}a,b). Logs and
demonstrations contain little of it. World models trained on logs of human or expert
driving~\citep{hu2023gaia1,wang2023drivedreamer,wang2024drivewm,zheng2025world4drive,hu2022mile} see what competent
drivers do and rarely what follows from a mistake of the ego vehicle, and asked what would have followed an action
other than the logged one, they can miss what would actually have happened~\citep{zhang2026counterfactual};
behavior cloning on such data does not explore at all, and an expert who never  makes mistakes cannot show how to
recover~\citep{ross2011dagger,codevilla2019limitations}. Methods that learn from interaction gather the experience
by failing: reinforcement-learning agents and world models learned online fail in simulation, typically over millions
of steps~\citep{zhang2021roach,toromanoff2020affordances,li2024think2drive}, RaWMPC exposes its world model to
hazardous behavior through a risk-aware interaction strategy~\citep{sun2026rawmpc}, and closed-loop training methods
generate the scenarios in which the current policy is likely to
fail~\citep{zhang2023cat,sheng2026curricuvlm,li2026worldengine}. These failures are the learner's own. The learner
has to cause them, over many trials that would be unacceptable on public roads, and each failure teaches it only the
outcome, a collision or a low reward, not what should have been done instead.

We let the world model learn
instead from failures that others make, judged by an expert. We call the resulting framework
SAFER, after its five stages, which
Fig.~\ref{fig:teaser}c groups into three parts. Offline (top), safety-critical failures are staged
deliberately in simulation and annotated by an expert. Partially trained agents and scripted, recoverable
perturbations drive until their failures run to their consequences, so that the world model learns what mistakes
lead to, while the expert records at every visited state what should have been done instead. In planning (middle), a
chunked world model foresees the consequences of candidate maneuvers, and a risk-aware planner
evaluates them by their predicted return minus a price on collision risk. Online (bottom), the world model
is refined on the planner's own driving. SAFER thus
broadens the coverage of the world model offline, so that much of what the planner explores is already covered, and
extends it online, gradually,
wherever the planner moves beyond it (Fig.~\ref{fig:teaser}c).

\begin{figure}[t]
  \centering
  \includegraphics[width=\linewidth]{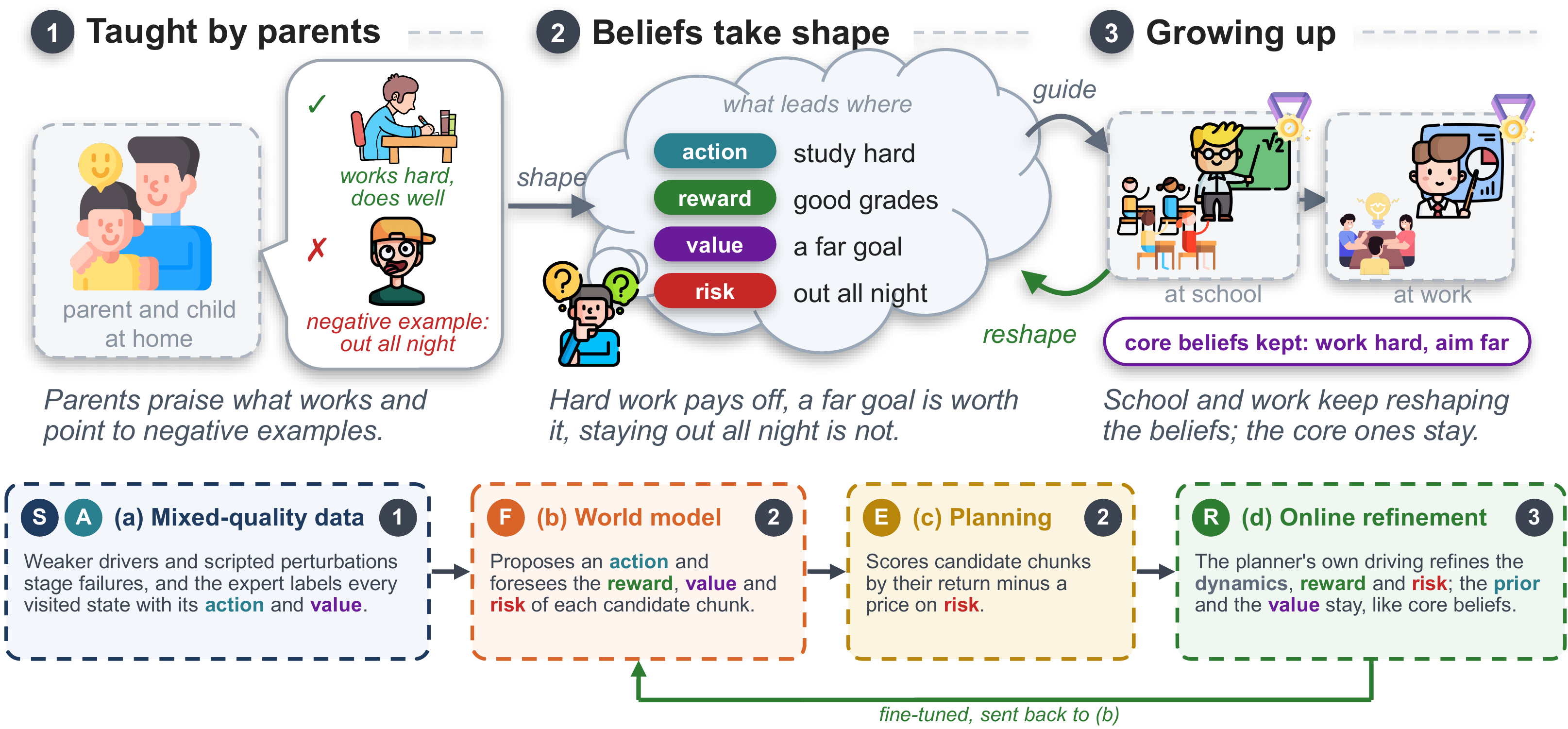}
  \caption{An everyday analogy for SAFER (top) and the corresponding stages of the method (bottom, detailed in
  Fig.~\ref{fig:overview}); the numbered badges link each stage to its scene. (1)~Parents praise what works and
  point to negative examples, so that the child need not fail first. (a)~Likewise, weaker drivers and scripted
  perturbations stage failures in simulation, and the expert labels every visited state with its action and
  value. (2)~The child learns what each choice leads to. Hard work (action) brings good grades (reward), a far
  goal is worth pursuing (value), and staying out all night is risky (risk). (b)~Likewise, the world model
  foresees these quantities for each candidate action chunk, and (c)~the planner scores the candidates by their
  return minus a price on risk. (3)~Away from home, school and work keep reshaping the beliefs while the core
  ones stay. (d)~Likewise, the planner's own driving refines the dynamics and the reward and risk heads, while
  the prior and the value, whose targets only the expert provides, stay fixed.}
  \label{fig:analogy}
\end{figure}

SAFER is inspired by how people resolve the same dilemma as they grow
up (Fig.~\ref{fig:analogy}, top). A child is not left to
discover the consequences of every choice alone. Parents praise what works and point to negative examples, so that the child learns what leads where without having to fail first. The beliefs formed in this way,
which choices pay off now, which goals are worth pursuing and which behavior is dangerous, let the child weigh a
choice before making it. Later, school and work keep reshaping these beliefs in situations the parents never
showed, while the core ones stay. Such learning does need failures, but failures that others make, where they can
be afforded, judged by someone who knows what should have been done. SAFER follows the same course
(Fig.~\ref{fig:analogy}, bottom). Like parenting, its offline data, staged and annotated by the expert, provide a
wealth of experience to grow up with. Like the child, the world model foresees what leads where and forms the same
beliefs, as predictions of reward, value and collision risk, by which the planner evaluates each choice. Like school
and work, the online stage keeps the planner learning from its own driving. It refines the dynamics and the reward
and risk predictions, while the prior and the value stay fixed, like core beliefs. Driving already uses failures of this kind to
widen the data of imitation learning. Logged trajectories are perturbed and the learner is trained to recover from
them~\citep{bansal2019chauffeurnet}, noise is injected into an expert's driving while the expert keeps
labeling~\citep{laskey2017dart}, and the interventions of safety drivers are recorded.

Behavior cloning on the
expert's labels learns to recover rather than to repeat the mistakes that produced
the data, and the resulting prior is a head of a chunked latent world model (Section~\ref{sec:method:wm}) of the kind introduced by
SimDist~\citep{levy2026simdist}, which imagines a whole chunk of the future in one pass rather than step by step and so
keeps planning fast. A collision-risk head (Section~\ref{sec:method:risk}) learns from the recorded
collisions of this corpus to predict the probability that a candidate action sequence leads to a collision, and it
is trained so that braking harder does not raise the predicted risk. The risk-aware model predictive path integral (MPPI)
planner~\citep{williams2017mppi} (Section~\ref{sec:method:planner}) scores candidate action chunks by their
predicted return minus a bounded collision penalty. The online loop (Section~\ref{sec:method:online}) adapts the
dynamics, the reward scorers and the risk head on the planner's own driving and keeps the prior and the value
fixed, while the planner scores with slowly moving copies of the adapted heads.
Observations and evaluation follow the privileged bird's-eye-view (BEV) setting of Roach~\citep{zhang2021roach} in
CARLA~\citep{dosovitskiy2017carla} and the closed-loop benchmark of CAT~\citep{zhang2023cat} in
MetaDrive~\citep{li2023metadrive}.

Our contributions are summarized as follows:
\begin{itemize}
  \item A formulation of the exploration--exploitation dilemma of world-model-based planning in driving, in which
        searching beyond the coverage of the world model can improve on the data, but there the errors
        of the model bias the choice of action, and the experience that would correct them consists of failures. The
        bias and the regret it causes are stated formally in Section~\ref{sec:method:planner}.
  \item A data recipe in which the world model learns from failures that others make, judged by an expert. Partially
        trained agents and scripted, recoverable perturbations drive in simulation until
        their failures run to their consequences, and the expert annotates every visited
        state with its action and value, so
        that the data cover much of what the planner explores.
  \item A fast chunked latent world model with a collision-risk head that
        is trained to be monotonic in braking and foresees collisions in recorded driving before they happen, without
        false alarms, and a risk-aware MPPI
        planner that scores candidate maneuvers by their predicted return minus a bounded collision penalty
        and plans in real time on a single GPU; the imagination of the world model carries the information on
        which driving decisions depend, such as a traffic light that is about to change.
  \item An online adaptation rule that extends the coverage from the planner's own driving by fine-tuning the
        reward scorers and the risk head together with the dynamics, while the planner scores against slowly
        moving copies of the adapted heads; adapting these heads, and not only the dynamics, makes every type of
        collision rarer.
  \item A closed-loop evaluation that isolates the contribution of each stage in CARLA, including towns without
        training data, and compares SAFER with the baselines, against which it reaches the highest route completion and reward per step in CARLA and the
        highest episode reward in MetaDrive, in regular and safety-critical traffic; one turn of a data flywheel,
        which retrains the system on data targeted at the scenarios that the offline
        corpus covers least, makes it safer still.
\end{itemize}

The remainder of the paper reviews related work (Section~\ref{sec:related}), presents the method
(Section~\ref{sec:method}), reports the experiments (Section~\ref{sec:experiments}) and concludes
(Section~\ref{sec:conclusion}).

\section{Related work}
\label{sec:related}

\subsection{Learning-based driving policies}
\label{sec:related:driving}

Learning-based driving policies are trained mainly by imitation learning and by reinforcement learning.
Imitation learning maps observations to the actions of an expert, from conditional imitation of human
driving~\citep{codevilla2018cil} to privileged experts that see the simulator state and teach sensorimotor
students~\citep{chen2020lbc,zhang2021roach}. A policy trained on the expert's states compounds its own errors once it
leaves them, and demonstrations that contain no mistakes contain no corrections
either~\citep{ross2011dagger,codevilla2019limitations}; imitation is also prone to causal confusion, in which the
policy keys on correlates of the expert's action rather than on its causes~\citep{dehaan2019causal}. DAgger lets the
student drive and has the expert label the states it visits, which is how \citet{chen2020lbc} and
\citet{zhang2021roach} train their students, and DART injects noise into the expert's
actions~\citep{laskey2017dart}. Both widen the states around the expert, but neither shows what follows from driving
badly.

Reinforcement learning learns from outcomes instead. Model-free agents reach strong closed-loop performance in
CARLA~\citep{toromanoff2020affordances,zhang2021roach}, and the PPO agent of \citet{zhang2021roach} is the expert we
use in CARLA, but such agents are shaped by millions of simulator steps under a hand-designed reward. Vision-language models
have recently been used to enrich this reward and to guide training~\citep{huang2025vlmrl,qu2026foundrl,huang2026drivevlmrl}.
\citet{chen2021worldonrails} gain sample efficiency by assuming that the world does not react to the ego vehicle,
which removes exactly the interactions that matter for safety. Safety-critical scenarios are generated by perturbing
other road users~\citep{ding2020learningtocollide,wang2021advsim} or interactively with a multimodal large language
model~\citep{sheng2025talk2traffic}, and closed-loop training methods train the policy on such
scenarios~\citep{zhang2023cat,niu2024clic,sheng2026curricuvlm,li2026worldengine}. All of these learn from the
failures of the policy that is being trained.

The corpus of SAFER (Section~\ref{sec:method:data}) takes DAgger one step further. The states are visited by drivers that are deliberately worse than the expert, partially trained checkpoints of an RL agent, and by perturbations of the ego's own actions that run until a consequence occurs, and the expert
annotates each state with its action and value. A policy can learn recoveries from these labels, and a world model
can, in addition, learn what the mistakes lead to.

\subsection{Model-based reinforcement learning and planning}
\label{sec:related:wm}

A world model predicts how the environment evolves under the agent's actions, so that the agent can learn or
plan in imagination~\citep{ha2018worldmodels}. Agents use it in two ways. The Dreamer family trains an actor and a
critic on imagined trajectories~\citep{hafner2020dreamer,hafner2021dreamerv2,hafner2023dreamerv3}, and its most recent
member trains agents entirely inside a world model of complex environments~\citep{hafner2025dreamer4}. PlaNet,
TD-MPC and SimDist instead plan online with the model~\citep{hafner2019planet,hansen2022tdmpc,hansen2024tdmpc2,levy2026simdist},
searching over action sequences with sampling-based optimizers such as the cross-entropy method and model predictive
path integral control~\citep{williams2017mppi,pinneri2020icem}. Online planning keeps the objective explicit at
decision time, so that it can be extended, for example with the conditional value-at-risk of the sampled
returns~\citep{yin2023cvarmppi}.

Both uses share a weakness. The agent optimizes against the model, and the optimizer is drawn to the errors of
the model in regions the model was not trained on. One line of work keeps the agent where the model is reliable,
through probabilistic ensembles~\citep{chua2018pets}, short model rollouts branched from real
data~\citep{janner2019mbpo} or uncertainty-penalized rewards~\citep{yu2020mopo}. This is the exploitation side of the
dilemma of Section~\ref{sec:intro}, and it gives up the improvement that lies beyond the coverage of the model.
Another line widens the coverage by exploring, acting optimistically on the epistemic uncertainty of the
model~\citep{curi2020hucrl} or planning towards expected novelty~\citep{sekar2020plan2explore}, which in driving
means failing. SAFER widens the coverage from the data side instead. The failures that a planner would explore are
staged offline and annotated by an expert, and the model is re-fitted online on the states that the planner actually
reaches.

\subsection{World models for autonomous driving}
\label{sec:related:drivingwm}

Driving world models differ mainly in what they learn from. Models trained on logs of competent driving generate
future scenes~\citep{hu2023gaia1,wang2023drivedreamer}, select among candidate trajectories by their predicted
futures~\citep{wang2024drivewm,zheng2025world4drive}, or, like MILE, learn a latent world model and a driving policy
jointly from expert demonstrations~\citep{hu2022mile}; conditioned on an action other than the logged one, such
models can miss what would actually have happened~\citep{zhang2026counterfactual}. Other models learn from the
agent's own interaction. Think2Drive trains a policy inside a world model that it learns
online~\citep{li2024think2drive}, and InDRiVE and ExploreVLA explore by curiosity, rewarding the disagreement or the
uncertainty of the world model's predictions~\citep{khanzada2025indrive,sheng2026explorevla}. Closest to our aim,
RaWMPC folds the predicted probabilities of driving events into the cost of model predictive control and exposes its
world model to hazardous behavior through a risk-aware interaction strategy, without expert
demonstrations~\citep{sun2026rawmpc}, and CorrRisk-WM predicts the hazard that surrounding agents pose to the
corridor of each candidate trajectory~\citep{guo2026corrriskwm}.

SAFER also learns from failures, but from failures that other drivers make offline and that an expert annotates,
so that its data record both what a mistake leads to and what should have been done at the states before it; logs
contain few such failures, and models that learn from their own interaction have to cause them. As in RaWMPC, the
predicted probability of an event enters the planning cost, but the risk head of SAFER is supervised by the staged
failures and trained so that braking does not raise its prediction, and the planner keeps an explicit objective
rather than a policy distilled from the model.

\subsection{Continual adaptation of world models}
\label{sec:related:adapt}

A world model pretrained on one data distribution meets another once its own planner drives, and it has to keep
learning from the planner's experience without losing what it learned before. SimDist adapts a world model
pretrained in simulation to a real robot by re-fitting only its latent dynamics on the planner's
rollouts~\citep{levy2026simdist}, and AdaWM first identifies whether a loss in performance on a new driving task
stems from the policy or from the dynamics and then updates the dominant part with low-rank
updates~\citep{wang2025adawm}. In driving, the closed-loop distribution also shifts the quantities that the planner
scores, such as speeds, gaps and near-misses. SAFER therefore adapts the reward scorers and the collision-risk head
together with the dynamics, since all three can be supervised from recorded outcomes, and keeps the prior and the
value fixed, since only the expert can teach them. Every update mixes offline and online windows, so that the corpus
is not forgotten, and the planner scores with slowly moving copies of the adapted heads, the target-network idea of
value learning~\citep{mnih2015dqn}, so that its objective does not change with every update.

\section{Methodology}
\label{sec:method}

\begin{figure}[tbp]
  \centering
  \includegraphics[width=\linewidth]{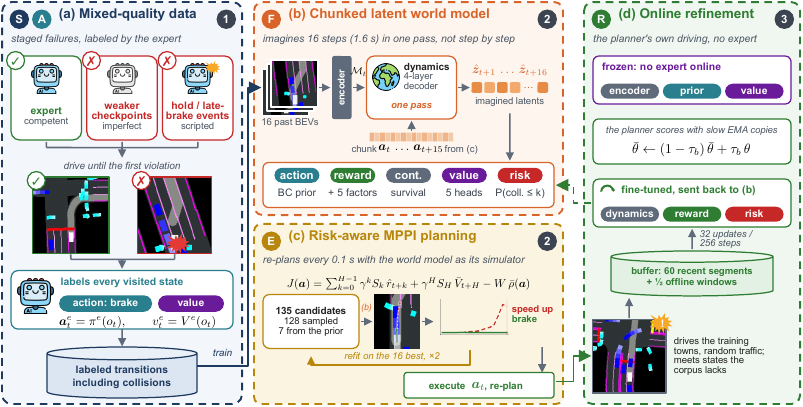}
  \caption{Technical overview of SAFER, illustrated with its CARLA instance. The numbered badges link each part
  to its scene of the analogy in Fig.~\ref{fig:analogy}. (a)~Stage and Annotate: the expert, weaker Roach
  checkpoints and scripted hold and late-brake events drive until the first violation, and the expert labels every
  visited state with its action and value (Section~\ref{sec:method:data}). (b)~Foresee: from the context $\mem_t$ of
  the past BEVs, the dynamics imagines the $16$ latents of a candidate chunk in one pass, and heads read them as the
  behavior-cloned prior, the reward and its five factors, survival, value and collision risk
  (Section~\ref{sec:method:wm}). (c)~Evaluate: every $0.1$\,s, the planner scores $135$ candidate chunks on their
  imagined futures (Eq.~\eqref{eq:score}), refits its sampling distribution to the $16$ best in each of two
  iterations and executes the first action (Section~\ref{sec:method:planner}). (d)~Refine: the planner's own driving
  fine-tunes the dynamics, the reward and the risk head online, while the encoder, the prior and the value stay
  frozen and the planner scores with slowly moving EMA copies (Section~\ref{sec:method:online}). The crash frame in
  (a), the context frames in (b) and the candidates and risk curves in (c) are from a recorded collision of the corpus
  (Fig.~\ref{fig:planner}a); the other two BEV frames are illustrative.}
  \label{fig:overview}
\end{figure}

This section describes SAFER. At every control step, the planner imagines the consequences of candidate action
sequences with a learned latent world model, searches for the sequence with the best predicted return, penalized by
the predicted probability that the sequence leads to a collision, and executes its first action. The five stages of SAFER map onto four parts,
the four panels of Fig.~\ref{fig:overview}:
\begin{enumerate}
  \item[(a)] \emph{Stage} and \emph{Annotate}: a corpus of \emph{mixed-quality} simulated experience, driven by
        a reinforcement-learning (RL) expert, by weaker checkpoints of an RL agent trained with the expert's
        recipe and by scripted perturbations, in which every step is labeled with the expert's action and value
        (Section~\ref{sec:method:data});
  \item[(b)] \emph{Foresee}: a \emph{chunked latent world model} that encodes an observation,
        predicts the latent states of a whole chunk of the future under a candidate action sequence
        in one forward pass, and reads the imagined trajectory with a physically decomposed reward, a
        continuation probability, an ensemble of value heads and a behavior-cloned action prior
        (Section~\ref{sec:method:wm}), together with a \emph{collision-risk head} that predicts the probability
        that a collision has occurred by each future step; the risk head is trained only on the simulator's
        recorded outcomes and regularized so that braking harder does not raise the predicted risk
        (Section~\ref{sec:method:risk});
  \item[(c)] \emph{Evaluate}: a \emph{risk-aware sampling-based planner} that runs model predictive path integral
        control (MPPI) on the world model, seeded with the behavior-cloned prior and penalized by the predicted
        collision risk (Section~\ref{sec:method:planner});
  \item[(d)] \emph{Refine}: an \emph{online adaptation} loop that keeps the dynamics, the reward scorers and the
        risk head consistent with the planner's own driving, while the planner scores against slowly moving
        exponential-moving-average (EMA) copies of them (Section~\ref{sec:method:online}).
\end{enumerate}

The design follows the decomposition of Simulation Distillation (SimDist) \citep{levy2026simdist}: an
observation encoder, a chunked latent dynamics model conditioned on an explicit history, reward and value
heads that read the imagined trajectory, and a base policy that warm-starts a sampling-based planner. The world
model of SAFER is trained on data that other policies generate and is then asked to plan on the states that its own
planner visits. A planner for safety-critical driving must also know what makes a candidate good or dangerous, and
its model must keep pace with the planner's own driving. SAFER therefore predicts the reward through its physical
factors and the value with an ensemble, prices the predicted collision probability in the planner's score with an
event-based risk head trained to be monotonic in braking, and adapts the scorers online while the planner scores
with their EMA copies.

\subsection{Problem formulation}
\label{sec:method:setting}

We formulate driving as a partially observable Markov decision process in discrete time, $t=0,1,\dots$, with
steps of fixed length. At step $t$ the agent receives an observation $\obs_t$, a partial view of the scene that
describes the surroundings of the ego vehicle and its own state, and applies an action $\act_t=(\alpha_t,\delta_t)$
made of two commands in $[-1,1]$: an acceleration command $\alpha_t$, applied as throttle $\alpha_t$ when it is
positive and as brake $-\alpha_t$ when it is negative, and a steering command $\delta_t$. The scene then evolves
under unknown, stochastic dynamics, which include the motion of the other road users, and the environment
returns a reward $r_t$, a termination flag $d_t\in\{0,1\}$ and a vector $\bm{e}_t\in\{0,1\}^{n_e}$ of safety
events, which marks the safety-critical outcomes of the step, such as a collision. The aim is a controller that
maximizes the expected discounted return $\E\big[\sum_{t=0}^{T}\gamma^{t}r_t\big]$ of an episode, with discount
$\gamma\in(0,1)$ and termination step $T$, while keeping the probability of a collision low.

We assume that the reward is a known function of quantities that the simulator records at every step,
\begin{equation}
  r_t=f\big(\bm{\phi}_t,\act_t,\act_{t-1}\big),
  \label{eq:reward}
\end{equation}
where the reward factors $\bm{\phi}_t\in\R^{n_\phi}$ are physical quantities, such as the speed of the ego
vehicle, or indicators of events, such as a contact, and $f$ is given in closed form. The world model uses
this structure by predicting the factors and evaluating $f$ on them (Section~\ref{sec:method:wm}).

Three sources of supervision are available. The first is an expert policy
$\pi^{e}$ with a value function $V^{e}$, both functions of the observation, so that they can be queried at any
observation, including those of states that the expert would never reach itself; we write
$\act^{e}_t=\pi^{e}(\obs_t)$ for its deterministic action and $v^{e}_t=V^{e}(\obs_t)$ for its value at step $t$.
The second is an offline corpus $\mathcal{D}$ of episodes driven by behavior policies that are allowed to fail,
in which every step carries the expert's labels $(\act^{e}_t,v^{e}_t)$ (Section~\ref{sec:method:data}). The third
is online interaction, in which the controller drives the environment itself and records
$(\obs_t,\act_t,r_t,\bm{\phi}_t,d_t,\bm{e}_t)$ but no expert labels, so that what it learns online does not depend
on the expert. The problem is to learn from $\mathcal{D}$ a controller that drives well and safely, and to keep
improving it from its own online experience.

SAFER solves this problem by receding-horizon planning with a learned world model. At step $t$ the controller
conditions on the last $\ctx$ observations and the actions between them, $(\obs_{t-\ctx+1:t},\act_{t-\ctx+1:t-1})$,
and searches over action chunks $\act_{t:t+\horizon-1}=(\act_t,\dots,\act_{t+\horizon-1})\in[-1,1]^{\horizon\times2}$
of $\horizon$ steps for one that maximizes a planning objective $J(\act_{t:t+\horizon-1})$; it executes the first
action of the chunk and plans again at step $t+1$. The objective, Eq.~\eqref{eq:score}, is the discounted return of
the chunk predicted by the world model, completed by a predicted value at the end of the horizon, minus a penalty
on the predicted probability of a collision. To evaluate it for a candidate chunk, the world model must predict the
latent states that the chunk leads to, the rewards, the continuation and the value along them
(Section~\ref{sec:method:wm}), and the probability of a collision (Section~\ref{sec:method:risk});
Section~\ref{sec:method:planner} describes the search and Section~\ref{sec:method:online} the online adaptation. We
use $\ctx=\horizon=16$ steps, i.e.\ $1.6$\,s of history and of horizon at the $0.1$\,s steps of both simulators.

The formulation is instantiated in two simulators. In CARLA~\citep{dosovitskiy2017carla}, steps are $0.1$\,s
long and $\obs_t=(\bev_t,\ego_t)$ (Fig.~\ref{fig:env}c), where $\bev_t$, a $15\times192\times192$ array with values
in $\{0,\dots,255\}$, is the privileged BEV mask stack of Roach~\citep{zhang2021roach} and $\ego_t\in\R^6$ is the
ego state. The BEV covers a $38.4\,\mathrm{m}\times38.4\,\mathrm{m}$ window at $5$ pixels per meter with the ego
drawn $40$ pixels from the bottom edge, and contains three static channels (drivable road, the planned route
corridor, lane markings with broken markings at a lower intensity) and three dynamic channel groups (vehicles,
pedestrians, and traffic-light stop lines together with stop signs) rasterized at four time offsets
$\{0,0.5,1.0,1.5\}$\,s in the past, giving $3+3\times4=15$ channels. The dynamic groups therefore encode $1.5$\,s of
motion history of every road user. The ego state is
$\ego_t=(\text{throttle},\text{steer},\text{brake},\text{gear}/5,v_x,v_y)$, i.e.\ the last applied control and the
planar velocity. The safety events are a collision, a red-light violation and a stop-sign violation ($n_e=3$). The
reward is the dense reward of Roach, whose factors are the ego speed, a desired speed that decreases towards zero
as the ego approaches a vehicle, a pedestrian, a red light or a stop sign in its path, the lateral distance to the
route, the heading error and the magnitude of the terminal penalty ($n_\phi=5$, Section~\ref{sec:exp:setup}); $f$
also penalizes changes of the steering command. In MetaDrive~\citep{li2023metadrive}, on the benchmark of CAT~\citep{zhang2023cat}, steps
are also $0.1$\,s long and $\obs_t\in[0,1]^{101}$ is a vector (Fig.~\ref{fig:env}f) that holds distances along $30$ lidar
beams to other vehicles, $30$ beams to the road edges and $12$ beams to the lane lines, $6$ values of the ego state,
$10$ navigation values, the lateral offset from the logged route and $4$ values for each of the three nearest
vehicles. The safety events are a contact with a vehicle, a contact with an object and leaving the road
($n_e=3$). CAT's reward is the progress along the route, except at the step of an arrival, an exit from the road or
a contact, where it takes a fixed value; its factors are the progress and the indicators of these four events
($n_\phi=5$).

\subsection{Data collection: staged failures annotated by an expert}
\label{sec:method:data}

The corpus $\mathcal{D}$ is generated in the simulator by behavior policies that are allowed to fail, and
every state they visit is labeled by the expert.

\paragraph{Behavior policies and perturbations}
Besides the expert, the behavior policies are $M$ checkpoints $\pi_1,\dots,\pi_M$ of a reinforcement-learning
agent trained from scratch with the expert's recipe and saved at increasing numbers of training steps, so that the
corpus contains the systematic mistakes of drivers at several levels of competence, such as late braking, lane
departures and running red lights, together with their consequences. Two perturbations of the action
$\act^{b}_t$ proposed by a behavior policy make the driving worse in a controlled way. The first is action noise. An
episode alternates between clean phases and corrupt phases of random lengths, and during a corrupt phase the
executed action is
\begin{equation}
  \act_t=\clip\big(\act^{b}_t+\bm{\xi}_t,-1,1\big),\qquad
  \bm{\xi}_t\sim\mathcal{N}\big(\bm{0},\operatorname{diag}(\sigma_\alpha^2,\sigma_\delta^2)\big),
  \label{eq:noise}
\end{equation}
with severities $(\sigma_\alpha,\sigma_\delta)$ that are drawn at random and redrawn at random intervals, while
$\act_t=\act^{b}_t$ in a clean phase; a small share of the episodes stays clean throughout, and half of the others
start in a corrupt phase, following SimDist's mixed-quality recipe. The second is a scripted event. Triggered by the
simulator state, it overrides the expert's longitudinal command for a few steps with the opposite decision, braking
where the expert would drive on or keeping the throttle where the expert brakes for a hazard, and is followed by a
recovery in which the expert's action is executed. An event that withholds braking is triggered only while the
expert could still stop before the hazard, so that every staged mistake is recoverable. These events place the two
situations that a planner most often gets wrong, waiting when it should go and going when it should brake, into the
data together with their labeled corrections.

The corpus is the union of three sets: $\mathcal{D}_{\mathrm{exp}}$, driven by the expert without
perturbation; $\mathcal{D}_{\mathrm{int}}$, driven by the checkpoints with action noise; and
$\mathcal{D}_{\mathrm{tgt}}$, driven by the expert with action noise and scripted events. The checkpoints, the
parameters of the perturbations and the events of each environment are given in Section~\ref{sec:experiments}.

\paragraph{Recorded quantities}
Episodes are not reset when a mistake happens; they run until the environment terminates them, so that the
consequences of the mistakes, the safety events among them, are recorded. Whichever policy drives, every step
stores
\begin{equation}
  \big(\obs_t,\;\act_t,\;r_t,\;\bm{\phi}_t,\;d_t,\;\bm{e}_t,\;\act^{e}_t,\;v^{e}_t\big),
  \label{eq:record}
\end{equation}
that is, the transition with its reward factors and safety events and, as labels, the expert's action and value at
that state. The labels are what make mixed-quality data useful. A world model learns from the transitions what
each mistake leads to, and a policy learns from the labels what should have been done at the states that the
mistakes reach. In this work the failures are staged in simulators; the recipe itself needs only a driver that is
allowed to fail recoverably and an expert that can be queried at the visited states, so a proving ground with a
safety driver would serve the same purpose.

\subsection{Chunked latent world model}
\label{sec:method:wm}

\paragraph{Encoder}
An observation encoder maps every observation to a latent state,
\begin{equation}
  \lat_t = E_\theta(\obs_t)\in\R^{256}.
  \label{eq:encoder}
\end{equation}
 It has the architecture of the expert's feature extractor, the part of the expert's network that
processes the observation; it is initialized from that extractor (Section~\ref{sec:experiments}) and fine-tuned
with the rest of the model. Initializing from the expert gives the world model, from the
first step, the features that were sufficient to drive well.

\paragraph{Context tokens}
 The context of
the last $\ctx$ observations and the actions that led to them is turned into a set of $2\ctx$ context
tokens,
\begin{equation}
  \begin{aligned}
  \mem_t&=\big\{\,\bm{u}^{z}_i,\;\bm{u}^{a}_i\,\big\}_{i=0}^{\ctx-1},\\
  \bm{u}^{z}_i&=\lat_{t-\ctx+1+i}+\bm{\eta}^{z}+\mathrm{PE}(i),\qquad
  \bm{u}^{a}_i=A_\theta(\act_{t-\ctx+i})+\bm{\eta}^{a}+\mathrm{PE}(i),
  \end{aligned}
  \label{eq:memory}
\end{equation}
where $A_\theta$ is a two-layer MLP action encoder, $\bm{\eta}^{z},\bm{\eta}^{a}$ are learned token-type
embeddings and $\mathrm{PE}(i)$ is a sinusoidal position code. The action token $\bm{u}^{a}_i$ carries the action
$\act_{t-\ctx+i}$ that led to $\lat_{t-\ctx+1+i}$; the first one is set to zero, because the action that led to the
first context observation lies outside the window. The context tokens depend on the history alone, not on
the candidate, so they are computed once per control step and shared by all candidates.

\paragraph{Action-conditioned latent rollout}
A candidate action chunk $\act_{t:t+\horizon-1}$ is encoded into $\horizon$ query tokens
$\bm{q}_k=A_\theta(\act_{t+k})+\mathrm{PE}(k)$, $k=0,\dots,\horizon-1$. A four-layer transformer decoder $D_\theta$
\citep{vaswani2017attention} with width $256$ and four heads attends from the queries to the context tokens and
among the queries with a causal mask, and predicts all future latents in one pass,
\begin{equation}
  \big[\latp_{t+1},\dots,\latp_{t+\horizon}\big]=D_\theta\big(\bm{q}_{0:\horizon-1};\,\mem_t\big),
  \label{eq:dynamics}
\end{equation}
where, here and below, a hat marks a quantity predicted by the model or computed from its
predictions. Because of the causal mask, $\latp_{t+k+1}$ depends on $\act_{t:t+k}$ only. Predicting the chunk jointly rather than by
step-wise recursion makes the cost of scoring a candidate one decoder pass instead of $\horizon$ sequential
passes, which is what makes sampling-based planning affordable in real time; the causal mask preserves the
semantics of an autoregressive model.

\paragraph{Reward, continuation and value scorers}
The scalar heads do not read the decoder states directly. Each group of heads is preceded by a
small temporal transformer, a scorer that maps a sequence over the $\horizon$ slots to one feature vector per slot
(a two-layer input MLP and one encoder layer without a causal mask), so that the prediction at slot $k$ can use
the whole imagined trajectory. The reward scorer $T^{r}_\theta$ reads, for every transition, the state \emph{before}
it and the action taken,
\begin{equation}
  \begin{aligned}
  \big(\bm{f}^{r}_{0},\dots,\bm{f}^{r}_{\horizon-1}\big)&=T^{r}_\theta\Big(\big(\lat^{-}_{t+k},\act_{t+k}\big)_{k=0}^{\horizon-1}\Big),\\
  \lat^{-}_{t}&=\lat_t,\qquad \lat^{-}_{t+k}=\latp_{t+k}\quad(k\ge1),
  \end{aligned}
  \label{eq:reward_scorer}
\end{equation}
while the value scorer $T^{v}_\theta$ reads the state \emph{after} it,
$(\bm{f}^{v}_{0},\dots,\bm{f}^{v}_{\horizon-1})=T^{v}_\theta\big((\latp_{t+k+1})_{k=0}^{\horizon-1}\big)$. The reward
and value heads are distributional~\citep{hafner2023dreamerv3}. Each outputs a probability vector $p\in\Delta^{B}$,
where $\Delta^{B}$ is the set of probability vectors over $B=101$ bins whose centers $c_1<\dots<c_B$ are uniformly
spaced in $\symlog$ space, $\symlog(x)=\operatorname{sign}(x)\ln(1+|x|)$, and the prediction is decoded as
$\hat{y}=\symexp\big(\sum_b p_b c_b\big)$ with $\symexp(y)=\operatorname{sign}(y)(e^{|y|}-1)$; the reward bins span
$[-3,3]$ and the value bins $[-10,10]$ in $\symlog$ units. Four
quantities are predicted per slot, the reward and its factors from the reward features and the continuation and
the values from the value features:
\begin{align}
  \hat{p}^{\,r}_{t+k} &= \operatorname{softmax}\big(R_\theta(\bm{f}^{r}_{k})\big) \in\Delta^{B},
  \label{eq:reward_heads}\\
  \hat{\bm{\phi}}_{t+k} &= g_\phi\big(\Phi_\theta(\bm{f}^{r}_{k})\big)\in\R^{n_\phi},
  \label{eq:factor_head}\\
  \hat{c}_{t+k} &= \operatorname{sigmoid}\big(C_\theta(\bm{f}^{v}_{k})\big),
  \label{eq:cont_head}\\
  \hat{p}^{\,v,i}_{t+k+1} &= \operatorname{softmax}\big(V^{(i)}_\theta(\bm{f}^{v}_{k})\big),\qquad i=1,\dots,K.
  \label{eq:value_heads}
\end{align}
$\hat{p}^{\,r}$ is a distribution over reward bins;  $\hat{\bm{\phi}}$ predicts the
reward factors of Eq.~\eqref{eq:reward} through an output map $g_\phi$ that is fixed for each factor. For a
continuous quantity it multiplies the output by a fixed scale, after a softplus if the quantity is non-negative,
and for an event indicator it applies a sigmoid, so that the prediction is a probability; $\hat{c}$ is the probability
that the episode continues after the transition; and $\{V^{(i)}_\theta\}_{i=1}^{K}$ are $K=5$ independently
initialized value heads on the shared value trunk, in the spirit of the value ensembles of
TD-MPC2~\citep{hansen2024tdmpc2}. The factor head allows the reward to be reconstructed from the predicted
factors exactly as the simulator computes it,
\begin{equation}
  \hat{r}^{\,\phi}_{t+k}=f\big(\hat{\bm{\phi}}_{t+k},\,\act_{t+k},\,\act_{t+k-1}\big),
  \label{eq:reward_from_factors}
\end{equation}
 where the actions are those of the candidate itself and
$\act_{t-1}$ is the last executed action; when some factors are event indicators, $\hat{r}^{\,\phi}_{t+k}$ is the
expectation of $f$ under independent indicators with the predicted probabilities.
Two reward estimates are thus available for every
imagined step, a direct one $\hat{r}^{\,s}_{t+k}=\symexp\big(\sum_b \hat{p}^{\,r}_{t+k,b}c_b\big)$ and a
factorized one $\hat{r}^{\,\phi}_{t+k}$. The planner uses their average
(Section~\ref{sec:method:planner}). The factorized path exists because the scalar reward compresses very
different situations into one number, whereas the factors force the model to represent, for every candidate,
the quantities from which the
simulator computes the reward, such as how fast the vehicle will be going and how fast it should be going given the
hazards ahead; these are the quantities on which the correct action depends.

\paragraph{Behavior-cloned action prior}
A separate three-layer transformer decoder $P_\theta$ attends from $\horizon$ learned queries to the same
context tokens $\mem_t$ and outputs, for each future step, the parameters of a squashed Gaussian,
\begin{equation}
  \begin{aligned}
  \pi_\theta(\act_{t+k}\mid\mem_t):\qquad &\act_{t+k}=\tanh(\bm{x}_k),\\
  &\bm{x}_k\sim\mathcal{N}\big(\bm{\mu}^{\pi}_k,\operatorname{diag}((\bm{\sigma}^{\pi}_k)^2)\big),\qquad
  \log\bm{\sigma}^{\pi}_k\in[-2,1],
  \end{aligned}
  \label{eq:policy}
\end{equation}
for $k=0,\dots,\horizon-1$. Its deterministic chunk $\act^{\mathrm{bc}}_{t+k}=\tanh(\bm{\mu}^{\pi}_k)$, clipped entrywise
to $[-\bar{a},\bar{a}]$ with $\bar{a}=0.999$, is the base policy that warm-starts the planner; executed on its own, it
is the BC-prior arm of the ablation in Section~\ref{sec:exp:ablation}. It reads
only the context, never the candidate actions, so it costs one decoder pass per step regardless of the number of
candidates.

\paragraph{Training windows}
The model is trained on windows of $\ctx+\horizon=32$ consecutive steps sampled uniformly from the corpus. The first $\ctx$ observations form the context and the following $\horizon$ transitions
the imagination targets, with the keyframe $t$ being the last context step. Windows that run into the end
of an episode are padded after the terminal transition, and the padded steps are masked. When the risk head is
trained on a corpus in which safety events end the episodes, windows are also started at every step before a terminal, so that each terminal event appears at every
horizon position between $1$ and $\horizon$. Without this, a collision that ends an episode would appear only at
the last position of a window, and the risk head would never see one at short range.

\paragraph{Objective}
Let $M^{v}_k\in\{0,1\}$ mark the valid (non-padded) transitions $k=0,\dots,\horizon-1$ of a window,
$M^{s}_k=M^{v}_k(1-d_{t+k})$ the transitions whose successor observation exists, and $\langle x\rangle_{M}$ the
mean of $x_k$ over the slots with $M_k=1$. With the Huber loss $\ell_{\mathrm{H}}$, the stop-gradient operator
$\sg(\cdot)$, the two-hot encoding $\twohot(y)\in\Delta^{B}$ of a scalar target $y$, which puts all its mass on the two
bin centers that bracket $\symlog(y)$, split so that its mean is $\symlog(y)$, the cross-entropy
$\mathrm{CE}(q,p)=-\sum_b q_b\log p_b$ and the binary cross-entropy $\mathrm{BCE}(y,p)=-y\log p-(1-y)\log(1-p)$,
each with the target first, the per-window losses are
\begin{align}
  \mathcal{L}_{\mathrm{lat}} &= \big\langle\,\ell_{\mathrm{H}}\big(\latp_{t+k+1},\,\sg(E_\theta(\obs_{t+k+1}))\big)\,\big\rangle_{M^{s}},
  \label{eq:loss_latent}\\
  \mathcal{L}_{\mathrm{rew}} &= \big\langle\,\mathrm{CE}\big(\twohot(r_{t+k}),\,\hat{p}^{\,r}_{t+k}\big)\,\big\rangle_{M^{v}},
  \label{eq:loss_reward}\\
  \mathcal{L}_{\phi} &= \big\langle\,\ell_{\phi}\big(\hat{\bm{\phi}}_{t+k},\,\bm{\phi}_{t+k}\big)\,\big\rangle_{M^{v}},
  \label{eq:loss_factor}\\
  \mathcal{L}_{\mathrm{cons}} &= \big\langle\,\ell_{\mathrm{H}}\big(\hat{r}^{\,\phi}_{t+k},\,r_{t+k}\big)\,\big\rangle_{M^{v}},
  \label{eq:loss_cons}\\
  \mathcal{L}_{\mathrm{cont}} &= \big\langle\,\mathrm{BCE}\big(1-d_{t+k},\,\hat{c}_{t+k}\big)\,\big\rangle_{M^{v}},
  \label{eq:loss_cont}\\
  \mathcal{L}_{\mathrm{val}} &= \frac{1}{K}\sum_{i=1}^{K}
      \big\langle\,\mathrm{CE}\big(\twohot(v^{e}_{t+k+1}),\,\hat{p}^{\,v,i}_{t+k+1}\big)\,\big\rangle_{M^{s}},
  \label{eq:loss_value}\\
  \mathcal{L}_{\mathrm{bc}} &= \big\langle\,-\log\pi_\theta\big(\act^{e}_{t+k}\mid\mem_t\big)\,\big\rangle_{M^{v}},
  \label{eq:loss_bc}
\end{align}
where $\ell_{\mathrm{H}}$ is averaged over the entries of its arguments, 
$\ell_{\phi}$ is the Huber loss on the continuous factors, each divided by its scale, averaged over them, plus the
binary cross-entropy of the indicator factors, averaged over them, and the
log-density of the squashed Gaussian at $\act=\act^{e}_{t+k}$ is
$\log\mathcal{N}\big(\operatorname{atanh}\act;\bm{\mu}^{\pi}_k,\operatorname{diag}((\bm{\sigma}^{\pi}_k)^2)\big)-\sum_j\log(1-a_j^2)$.
Three choices in Eqs.~\eqref{eq:loss_latent}--\eqref{eq:loss_bc} deserve comment. First, the latent target is the
encoder's own output with a stop-gradient, so that the dynamics is trained to reproduce the
representation rather than the representation being trained to be predictable. Second, the value heads
regress the expert's value $V^{e}$ evaluated at the recorded future observations $\obs_{t+k+1}$, not a
Monte-Carlo return of the behavior policy, because the corpus is driven mostly by degraded policies whose returns would measure their
own mistakes, whereas $V^{e}$ measures how good a state is under competent driving, which is the quantity
the planner needs for bootstrapping. Third, the behavior-cloning target is the expert's action at the
visited state for \emph{every} valid step, including steps at which a degraded policy or a perturbation has taken
the vehicle off the expert's trajectory. The prior therefore learns recoveries, not only expert behavior, unlike
the base policy of SimDist~\citep{levy2026simdist}, which is trained only on steps the expert executed. The
consistency loss $\mathcal{L}_{\mathrm{cons}}$ ties the factor head to the scalar reward through the known
reward formula, and the continuation loss provides the survival probabilities used by the planner.

The losses have very different scales and would otherwise require hand-tuned weights that drift during
training. Each term is therefore divided by a running estimate of its own root-mean-square,
$\tilde{\mathcal{L}}_j=\mathcal{L}_j/\sqrt{\nu_j}$, where $\nu_j\leftarrow0.99\,\nu_j+0.01\,\mathcal{L}_j^2$ is updated
at every training step and treated as a constant in the gradient, and the weights express relative importance only:
\begin{equation}
  \begin{aligned}
  \mathcal{L}_{\mathrm{wm}}={}&\tilde{\mathcal{L}}_{\mathrm{lat}}+\tilde{\mathcal{L}}_{\mathrm{rew}}
  +\tilde{\mathcal{L}}_{\phi}+0.1\,\tilde{\mathcal{L}}_{\mathrm{cons}}+0.1\,\tilde{\mathcal{L}}_{\mathrm{cont}}\\
  &+\tilde{\mathcal{L}}_{\mathrm{val}}+4\,\tilde{\mathcal{L}}_{\mathrm{bc}}
  \;(+\,\tilde{\mathcal{L}}_{\mathrm{risk}}).
  \end{aligned}
  \label{eq:loss_total}
\end{equation}
The risk term is introduced in the next section.  The training
schedules, which in CARLA add the risk head after a first stage and in MetaDrive train all parts together from the
start, are given in Section~\ref{sec:experiments}.

\subsection{Learned collision-risk head}
\label{sec:method:risk}

The return in the planner's score, Eq.~\eqref{eq:score} below, already penalizes safety events, through
the reward and the value bootstrap. A collision, however, enters the return only through the reward of a single
imagined step, diluted by the discount and, when it ends the episode, by the survival probability, and its effect on the score depends on
how well the reward head extrapolates to a rare event. A dedicated head that answers a simpler question,
\emph{has a collision happened by imagined step $k$}, can be trained as a classifier with class re-weighting, is
bounded, and yields a penalty whose scale is under direct control.
Similar event heads are used by RaWMPC~\citep{sun2026rawmpc}; our head differs in its cumulative target and
in the monotonicity constraint below, and it is supervised only by the simulator's recorded safety events,
never by any hand-written rule.

\paragraph{Architecture and target}
The head $\rho_\theta$ is a three-layer MLP (LayerNorm, two hidden layers of width $256$ with GELU and
dropout $0.1$) applied to the concatenation of the keyframe latent and each imagined latent,
\begin{equation}
  \bm{\rho}_{k}=\rho_\theta\big([\lat_t;\,\latp_{t+k+1}]\big)\in\R^{n_e},\qquad k=0,\dots,\horizon-1,
  \label{eq:risk_logits}
\end{equation}
one logit $\rho_{k,j}$ per safety-event type
$j=1,\dots,n_e$ of Section~\ref{sec:method:setting}. The target is cumulative. With the
recorded events $\bm{e}_{t+k}$ of the window and the validity mask $M^{v}$ of Section~\ref{sec:method:wm},
\begin{equation}
  y_{k,j}=\max_{k'\le k}\big(M^{v}_{k'}\,e_{t+k',j}\big),\qquad
  M^{r}_{k,j}=\max\big(M^{v}_{k},\,y_{k,j}\big).
  \label{eq:risk_target}
\end{equation}
When an event
ends the episode, the steps after it are padding; the mask $M^{r}_{k,j}$ keeps
them supervised with target $1$ (``the event has happened''), so that the logit at every later slot is
trained as well, whereas a channel without an event is not supervised on padding, where its outcome is unknown.
The classification loss is a positively re-weighted binary cross-entropy averaged over the masked entries,
\begin{equation}
  \mathcal{L}_{\mathrm{bce}}=\frac{\sum_{k,j}M^{r}_{k,j}\,\mathrm{BCE}_{w^{+}}\big(y_{k,j},\,\operatorname{sigmoid}(\rho_{k,j})\big)}{\sum_{k,j}M^{r}_{k,j}},
  \qquad w^{+}=20,
  \label{eq:risk_bce}
\end{equation}
where $\mathrm{BCE}_{w^{+}}(y,p)=-w^{+}y\log p-(1-y)\log(1-p)$; the positive class weight compensates for the rarity
of events among the transitions of a window.

\paragraph{Braking monotonicity}
A classifier trained on logged data inherits its correlations. In driving data hard braking co-occurs with
danger, and a head trained with Eq.~\eqref{eq:risk_bce} alone can learn that a candidate which brakes is
\emph{more} likely to be in a collision window, which is the opposite of what a planner needs. We counter this
bias with a soft constraint that needs no counterfactual outcome label. For each window, a
braked copy of the recorded action sequence, $\act^{-}_{t+k}=(\clip(\alpha_{t+k}-\beta,-1,1),\,\delta_{t+k})$ with a
brake offset $\beta\sim\mathcal{U}[0.1,0.5]$ drawn once per window, is rolled out with the same context into
latents $\latp^{-}_{t+k+1}$, and the head is penalized whenever the braked rollout is predicted to be riskier:
\begin{equation}
  \begin{aligned}
  \mathcal{L}_{\mathrm{mono}}&=\frac{1}{n_e\horizon}\sum_{k=0}^{\horizon-1}\sum_{j=1}^{n_e}
  \relu\Big(\operatorname{sigmoid}\big(\rho^{-}_{k,j}\big)-\operatorname{sigmoid}\big(\rho_{k,j}\big)\Big),\\
  \bm{\rho}^{-}_{k}&=\rho_\theta\big([\lat_t;\,\latp^{-}_{t+k+1}]\big).
  \end{aligned}
  \label{eq:risk_mono}
\end{equation}
The term is evaluated with dropout disabled, since two stochastic passes would differ even for identical
inputs and the hinge would penalize that noise. Both latent inputs are detached, so that the term can only
change the head; if it could reach the dynamics, the cheapest way to satisfy it would be to make the executed
rollout look more dangerous. The complete risk objective is
$\mathcal{L}_{\mathrm{risk}}=\mathcal{L}_{\mathrm{bce}}+\mathcal{L}_{\mathrm{mono}}$, RMS-normalized like
the other terms in Eq.~\eqref{eq:loss_total}. When the risk head is
trained together with the world model, the classification term is
allowed to shape the latents, so that the imagined states carry the information needed to foresee an event.

\paragraph{Planning penalty}
For a candidate $\act$ the planner subtracts
\begin{equation}
  \bar{\rho}(\act)=\frac{1}{\horizon}\sum_{k=0}^{\horizon-1}\sum_{j=1}^{n_e}\omega_j\,\operatorname{sigmoid}\big(\rho_{k,j}(\act)\big),
  \label{eq:risk_penalty}
\end{equation}
where $\rho_{k,j}(\act)$ are the logits of Eq.~\eqref{eq:risk_logits} on the rollout of $\act$ and
the non-negative weights $\bm{\omega}\in\R^{n_e}$ select the events that the planner prices; when $\bm{\omega}$
selects the collision channel alone, $\bar{\rho}(\act)\in[0,1]$ is the mean over the horizon of the predicted
probability that a collision has occurred. Because the
target is cumulative, an event predicted early contributes at every later slot and weighs more than the same event
predicted late, and a candidate that is predicted to avoid the event pays almost nothing. Channels with zero weight are trained but do not
enter the score; the weights of each environment are given in Section~\ref{sec:experiments}.

\subsection{Risk-aware MPPI planning}
\label{sec:method:planner}

At every control step the planner searches over action chunks $\act_{t:t+\horizon-1}$ using the world
model as its simulator. The search is the model predictive path integral (MPPI) update of
\citet{williams2017mppi}, with a shifted warm start and with the risk penalty of
Eq.~\eqref{eq:risk_penalty} in the score.

\paragraph{Candidate generation}
The context tokens $\mem_t$ and the action prior $\act^{\mathrm{bc}}=\act^{\mathrm{bc}}_{t:t+\horizon-1}$ of
Eq.~\eqref{eq:policy} are computed once per control step. The sampling distribution is a diagonal Gaussian over
chunks with mean $\bm{\mu}\in\R^{\horizon\times2}$ and standard deviation $\bm{\sigma}\in\R^{\horizon\times2}_{>0}$.
Its mean is warm-started from the previous step's final mean shifted by one step, with the prior's last action
filling the vacated slot (at the first step of an episode, from $\act^{\mathrm{bc}}$ itself), and its standard
deviation is reset at every step to $\bm{\sigma}_0=(0.25,0.08)$, for the longitudinal and the steering command at
every horizon step. In each of $I=2$ iterations $i=0,1$ the candidate set is
\begin{equation}
  \begin{aligned}
  \mathcal{A}_i={}&\underbrace{\Big\{\clip\big(\bm{\mu}+\bm{\sigma}\odot\bm{\epsilon}^{(i,n)},-\bar{a},\bar{a}\big)\Big\}_{n=1}^{N}}_{\text{sampled}}
  \;\cup\;
  \underbrace{\Big\{\clip\big(\act^{\mathrm{bc}}+\bm{\sigma}_{\mathrm{bc}}\odot\bm{\epsilon}'^{(n)},-\bar{a},\bar{a}\big)\Big\}_{n=1}^{N_{\mathrm{bc}}}}_{\text{prior proposals}}
  \;\cup\;\big\{\act^{\mathrm{bc}}\big\},
  \end{aligned}
  \label{eq:candidates}
\end{equation}
with $N=128$ samples, $N_{\mathrm{bc}}=\operatorname{round}(0.05N)=6$ prior proposals with
$\bm{\sigma}_{\mathrm{bc}}=(0.10,0.04)$ and the exact prior chunk as the last candidate, $135$ candidates in total;
every entry is clipped to $[-\bar{a},\bar{a}]$, as for the prior. All entries of the noise
$\bm{\epsilon}^{(i,n)}$ and $\bm{\epsilon}'^{(n)}$ are drawn independently from a standard normal distribution; the
sampled part is redrawn in every iteration, whereas the prior proposals are drawn once per control step and reused in
both iterations. The prior proposals and the exact prior keep the
behavior-cloned chunk and its neighborhood in the candidate set whatever the mean has drifted to, and
including the exact prior chunk guarantees that the best candidate of every iteration scores at least as high
as the behavior-cloned chunk (Lemma~\ref{lem:incumbent}).

\paragraph{Scoring}
Each candidate is rolled out with Eq.~\eqref{eq:dynamics} and scored with
\begin{equation}
  \begin{aligned}
  J(\act)&=\sum_{k=0}^{\horizon-1}\gamma^{k}\,S_k\,\hat{r}_{t+k}
  \;+\;\gamma^{\horizon}\,S_{\horizon}\,\bar{V}_{t+\horizon}
  \;-\;W\,\bar{\rho}(\act),\\
  S_k&=\prod_{k'=0}^{k-1}\hat{c}_{t+k'},
  \end{aligned}
  \label{eq:score}
\end{equation}
where all predicted quantities are those of the rollout of $\act$, $\gamma$ is the
discount of Section~\ref{sec:method:setting}, the blended reward is
$\hat{r}_{t+k}=\tfrac12\hat{r}^{\,s}_{t+k}+\tfrac12\hat{r}^{\,\phi}_{t+k}$, the survival probability $S_k$ is accumulated
from the continuation head ($S_0=1$), the terminal value
$\bar{V}_{t+\horizon}=\frac1K\sum_{i=1}^{K}\symexp\big(\sum_b\hat{p}^{\,v,i}_{t+\horizon,b}\,c_b\big)$ is the ensemble
mean of the values predicted for the last imagined state, and the risk weight is $W=50$. The value
bootstrap extends the effective horizon far beyond $1.6$\,s, which is necessary in driving, where the
consequences of an action such as entering an intersection often arrive after the horizon. The scale of
$W$ is set so that a collision predicted with certainty from the first imagined step, $\bar{\rho}=1$,
costs about as much as the whole discounted return of a good candidate.

\paragraph{Planning under model error}
$J$ is the world model's estimate of the planner's objective. Writing $J^{\star}(\act)$ for the same objective
evaluated in expectation over the true environment, with the rewards, terminations and collisions that executing
$\act$ actually produces and the expert's value of the state it actually reaches, the cost of exploring beyond the
coverage of the model, the exploration side of the dilemma of Section~\ref{sec:intro}, can be stated precisely.

\begin{definition}[Model error and the planner's choice]\label{def:model_error}
At a decision state, let $\mathcal{A}$ be a set of candidate chunks on which $J$ and $J^{\star}$ attain their
maxima. The \emph{model error} of a candidate is $e_J(\act)=J(\act)-J^{\star}(\act)$, the planner's choice is
$\hat{\act}\in\arg\max_{\act\in\mathcal{A}}J(\act)$, and the best candidate is
$\act^{\star}\in\arg\max_{\act\in\mathcal{A}}J^{\star}(\act)$.
\end{definition}

\begin{proposition}[The planner overestimates its own choice]\label{prop:curse}
Let $\mathcal{A}$ be fixed and $J^{\star}$ deterministic, and let the model be unbiased at every fixed
candidate, $\E[e_J(\act)]=0$ for all $\act\in\mathcal{A}$, the expectation being over the randomness of the
training data and of the model. Then
\begin{equation*}
  \E\big[e_J(\hat{\act})\big]\;\ge\;\E\big[J^{\star}(\act^{\star})-J^{\star}(\hat{\act})\big]\;\ge\;0 .
\end{equation*}
\end{proposition}

\begin{proposition}[Regret is bounded by the error at two candidates]\label{prop:regret}
For every realization of the model,
\begin{equation*}
  0\;\le\;J^{\star}(\act^{\star})-J^{\star}(\hat{\act})\;\le\;e_J(\hat{\act})-e_J(\act^{\star})\;\le\;|e_J(\hat{\act})|+|e_J(\act^{\star})| .
\end{equation*}
\end{proposition}

\begin{remark}\label{rem:dilemma}
Proposition~\ref{prop:curse} is the optimizer's curse~\citep{smith2006optimizerscurse}. Even a model that is
unbiased at every candidate, once maximized, overestimates the candidate it selects, by at least the expected
regret of that selection. Proposition~\ref{prop:regret} shows that the regret is bounded by the errors at only two
candidates, the best one and the chosen one. Maximizing $J$ thus draws $\hat{\act}$ towards candidates whose
value the model overestimates, typically candidates that the corpus does not cover. Staging failures and
annotating them with the expert (Section~\ref{sec:method:data}) reduces the error on the candidates that competent
and failing drivers take, and online refinement (Section~\ref{sec:method:online}) reduces it where the planner's
own choices lead. Proofs of all formal statements are in \ref{app:proofs}.
\end{remark}

\paragraph{Update and execution}
Index the candidates of an iteration as $\act^{(n)}$, $n=1,\dots,N+N_{\mathrm{bc}}+1$, and let $\mathcal{E}$ be the
indices of the $N_e=16$ candidates with the highest scores, the elites. The distribution is moved towards them with
exponential weights,
\begin{equation}
  \begin{aligned}
  w_n&=\frac{\exp\big(\kappa\,(J(\act^{(n)})-J_{\max})\big)}{\sum_{n'\in\mathcal{E}}\exp\big(\kappa\,(J(\act^{(n')})-J_{\max})\big)},
  \qquad n\in\mathcal{E},\\
  \bm{\mu}&\leftarrow\sum_{n\in\mathcal{E}}w_n\act^{(n)},\qquad
  \bm{\sigma}\leftarrow\max\Big(\sqrt{\textstyle\sum_{n\in\mathcal{E}}w_n\big(\act^{(n)}-\bm{\mu}\big)^{2}},\;\bm{\sigma}_{\min}\Big),
  \end{aligned}
  \label{eq:mppi_update}
\end{equation}
with $J_{\max}=\max_{n\in\mathcal{E}}J(\act^{(n)})$, which cancels in the ratio and only keeps the exponentials
bounded, $\kappa=0.25$ and $\bm{\sigma}_{\min}=(0.05,0.02)$; the square, the root and the maximum are taken entrywise,
and $\bm{\sigma}$ is computed around the updated mean. After the last iteration the first action of the mean is
executed as $\act_t$, and the mean is stored for the next step's warm start. The
planner executes the weighted mean rather than the single best sample, which averages out the randomness of any
single sample. In the evaluation of Section~\ref{sec:experiments} the planner runs at $10$\,Hz on one GPU,
re-planning at every step with the full context.

\paragraph{Properties of the update}
Two facts of the implementation make the following statements exact. Scoring is deterministic at planning
time (all modules are in evaluation mode), and within a control step the fixed candidates, the prior proposals
and $\act^{\mathrm{bc}}$, are formed once and are the same in every iteration.

\begin{lemma}[The prior is a floor]\label{lem:incumbent}
Let $\mathcal{F}_t$ be the fixed candidates of control step $t$, the $N_{\mathrm{bc}}$ prior proposals and
$\act^{\mathrm{bc}}$, and let $J^{\mathrm{best}}_i=\max_{\act\in\mathcal{A}_i}J(\act)$ be the best score among
the candidates $\mathcal{A}_i$ of iteration $i$ of that step (Eq.~\eqref{eq:candidates}). Then for every iteration
$i=0,\dots,I-1$,
$J^{\mathrm{best}}_i\ge\max_{\act\in\mathcal{F}_t}J(\act)\ge J(\act^{\mathrm{bc}})$.
\end{lemma}

The planned chunk is the weighted mean $\bm{\mu}$ of the elites rather than the best candidate, and because $J$ is
in general not concave, $J(\bm{\mu})$ can be lower than $J^{\mathrm{best}}_{I-1}$. What holds for $\bm{\mu}$ is
asymptotic in the number of samples.

\begin{assumption}\label{ass:maximiser}
On the action box $\mathcal{C}=[-\bar{a},\bar{a}]^{\horizon\times2}$, which contains every candidate of
Eq.~\eqref{eq:candidates}, the score $J$ attains its maximum at a unique $\hat{\act}_{\mathcal{C}}$, is continuous at
$\hat{\act}_{\mathcal{C}}$, and the maximum is well separated:
$\sup\{J(\act):\act\in\mathcal{C},\ \|\act-\hat{\act}_{\mathcal{C}}\|\ge\epsilon\}<J(\hat{\act}_{\mathcal{C}})$ for
every $\epsilon>0$.
\end{assumption}

\begin{theorem}[Consistency of the elite update]\label{thm:consistency}
Under Assumption~\ref{ass:maximiser}, consider one iteration whose candidate set contains, besides any other
candidates in $\mathcal{C}$, $N_{\mathrm{bc}}$ prior proposals drawn independently from the law of
$\clip(\act^{\mathrm{bc}}+\bm{\sigma}_{\mathrm{bc}}\odot\bm{\epsilon}',-\bar{a},\bar{a})$, with $\bm{\epsilon}'$ standard
normal and $\bm{\sigma}_{\mathrm{bc}}>\bm{0}$. For any fixed number of elites $N_e$ and any $\kappa\ge0$, the updated
mean $\bm{\mu}$ of Eq.~\eqref{eq:mppi_update} and the best candidate converge to $\hat{\act}_{\mathcal{C}}$ almost
surely as $N_{\mathrm{bc}}\to\infty$.
\end{theorem}

\begin{corollary}[The remaining error is the model's]\label{cor:regret}
Let $J^{\star}$ attain its maximum over $\mathcal{C}$ at $\act^{\star}_{\mathcal{C}}$. In the limit of
Theorem~\ref{thm:consistency} the planned chunk is $\hat{\act}_{\mathcal{C}}$, and its regret is bounded by the model
error at two chunks,
$0\le J^{\star}(\act^{\star}_{\mathcal{C}})-J^{\star}(\hat{\act}_{\mathcal{C}})\le|e_J(\hat{\act}_{\mathcal{C}})|+|e_J(\act^{\star}_{\mathcal{C}})|$.
\end{corollary}

\begin{remark}\label{rem:consistency}
Theorem~\ref{thm:consistency} is a statement about the limit of many samples and says nothing about the rate. With
$N_{\mathrm{bc}}=6$ proposals and $N=128$ samples in $32$ dimensions the planner is far from this limit. Full
support comes from the prior proposals, whose Gaussian noise gives every neighborhood of
$\hat{\act}_{\mathcal{C}}$ positive probability. The samples around the mean have full support as well, but their
law depends on the warm start and, in the second iteration, on the candidates of the first, whereas the
proposals are drawn from one fixed law per control step; the statement therefore relies on the proposals alone.
A reward that is discontinuous in the actions, such as CARLA's penalty
for changing the steering, makes the factorized reward, Eq.~\eqref{eq:reward_from_factors}, and with it $J$
discontinuous, which is why Assumption~\ref{ass:maximiser} asks for continuity only at the
maximizer. Corollary~\ref{cor:regret} is where the two halves of the method meet. Sampling removes the error
of the optimizer, and what remains is the model error at the chosen and at the best chunk, the quantity that
the data (Section~\ref{sec:method:data}) and the online refinement (Section~\ref{sec:method:online}) reduce.
\end{remark}

\subsection{Online adaptation with EMA scoring targets}
\label{sec:method:online}

A world model trained on the corpus of Section~\ref{sec:method:data} has learned from the states that other
drivers visited, not from those that its own planner visits. SimDist addresses the analogous simulation-to-reality
gap by collecting rollouts with the planner and re-fitting only the latent dynamics on them. We adopt the same loop (Algorithm~\ref{alg:online})
with two changes. First, the reward scorers and the risk head are adapted as well, because the closed-loop
distribution of speeds, gaps and near-misses differs from the corpus and these heads, not only the dynamics,
determine which candidate wins. Second, the planner does not score with the live scorers but with EMA
copies that follow the live parameters slowly, so that the objective it optimizes does not move with every
gradient step.

\paragraph{Interleaved collection and updates}
 The planner drives in the environment to which
it is to be adapted, which need not be the one in which the corpus was collected; in our experiments these are the
training environments of each benchmark, so that its test routes and scenarios stay held out
(Section~\ref{sec:experiments}). The
planner's transitions are kept in an in-memory buffer of fixed size that holds only the most recent ones, and the
parts of episodes still being driven are added to it whenever windows are drawn (the sizes are given in
\ref{app:impl}).
After every burst of $n_b=256$ environment steps, $\lfloor 0.125\,n_b\rfloor=32$
gradient updates are made (an update-to-data ratio of $0.125$, with fractional credit carried over). Each update batch mixes offline and online windows. Half of
the windows of a batch are drawn uniformly from the offline corpus and half from the online buffer, and half
of the online windows come from the episodes that the most recent burst completed or extended. Online windows use the same
$\ctx+\horizon$ layout and terminal padding as offline ones. 

\paragraph{What is updated}
The encoder $E_\theta$, the action prior $\pi_\theta$ and the value ensemble stay frozen. The value ensemble is frozen because its targets, the expert's values, are not
available online, and a value learned from the planner's own returns would bootstrap on its own mistakes.
Besides the risk head, two parameter groups are trained, with the scorer losses computed on detached rollouts so
that they cannot move the dynamics:
\begin{align}
  \mathcal{L}^{\mathrm{on}}_{\mathrm{dyn}}&=\tilde{\mathcal{L}}_{\mathrm{lat}},
  \label{eq:online_dyn}\\
  \mathcal{L}^{\mathrm{on}}_{\mathrm{sc}}&=\tilde{\mathcal{L}}_{\mathrm{rew}}+\tilde{\mathcal{L}}_{\phi}
  +\tilde{\mathcal{L}}_{\mathrm{cons}}+0.1\,\tilde{\mathcal{L}}_{\mathrm{cont}},
  \label{eq:online_scorer}
\end{align}
where Eq.~\eqref{eq:online_dyn} trains the dynamics $D_\theta$ (learning rate $3\times10^{-5}$) and
Eq.~\eqref{eq:online_scorer} trains the reward scorer and the reward, factor and continuation heads
$(T^{r}_\theta,R_\theta,\Phi_\theta,C_\theta)$ (learning rate $10^{-5}$). The risk head is trained with
$\mathcal{L}_{\mathrm{bce}}+\mathcal{L}_{\mathrm{mono}}$ on the same mixed batches, with its latent inputs
detached, at learning rate $10^{-4}$. Online windows carry recorded safety events but no expert labels;
Eqs.~\eqref{eq:online_dyn}--\eqref{eq:online_scorer} and the risk losses need only rewards, reward factors, events and
observations, which is precisely why these modules, and not the prior or the value heads, can be adapted
without an expert in the loop.

\paragraph{EMA targets for planning}
The planner rolls candidates out with the live dynamics but scores them with EMA copies
$\bar{\theta}$ of the scorers and of the risk head. After each burst,
\begin{equation}
  \begin{aligned}
  \bar{\theta}&\leftarrow(1-\tau_b)\,\bar{\theta}+\tau_b\,\theta,\\
  \tau_b&=1-(1-\tau)^{\,n_b/N_c},\qquad \tau=0.1,\quad N_c=1000,
  \end{aligned}
  \label{eq:ema}
\end{equation}
so that the target moves by the fraction $\tau$ per $N_c$ collected steps regardless of the burst size
($\tau_b\approx0.027$ for $n_b=256$); the risk head uses $\tau_b=0.1$ per burst directly. The EMA copies are
what is saved and what the evaluation of Section~\ref{sec:experiments} uses. The reasoning is the one behind
target networks in value learning. The reward and risk scorers define the planner's objective, and an
objective that changes every $32$ updates would make consecutive plans inconsistent; the slow copy keeps
the objective stable while still tracking the closed-loop distribution over hundreds of bursts.

In the following statements $\theta_m$ and $\bar{\theta}_m$ denote the live parameters and the EMA copy after burst
$m=0,1,\dots$ of the loop, so that Eq.~\eqref{eq:ema} reads $\bar{\theta}_{m+1}=(1-\tau_b)\,\bar{\theta}_m+\tau_b\,\theta_{m+1}$.

\begin{lemma}[The target's rate does not depend on the burst size]\label{lem:ema_rate}
After $m$ bursts of $n_b$ steps, $\bar{\theta}_m$ is a convex combination of its initial value, with weight
$(1-\tau_b)^m=(1-\tau)^{m n_b/N_c}$, and of the live parameters $\theta_1,\dots,\theta_m$ after each burst. In
particular, the weight of the initial value depends only on the number $s=m n_b$ of collected steps, as
$(1-\tau)^{s/N_c}$, whatever the burst size $n_b$, and equals $1-\tau$ at $s=N_c$.
\end{lemma}

\begin{proposition}[Tracking]\label{prop:tracking}
Let $\tau_b\in(0,1]$ be constant and let the live parameters change by at most $\Delta_\theta$ per burst,
$\|\theta_{m+1}-\theta_m\|\le\Delta_\theta$. Then for every $m$
\begin{equation*}
  \|\bar{\theta}_m-\theta_m\|\;\le\;(1-\tau_b)^{m}\,\|\bar{\theta}_0-\theta_0\|
  \;+\;\frac{1-\tau_b}{\tau_b}\big(1-(1-\tau_b)^{m}\big)\,\Delta_\theta ,
\end{equation*}
and if $\theta_m$ converges, $\bar{\theta}_m$ converges to the same limit.
\end{proposition}

\begin{proposition}[Attenuation of update noise]\label{prop:attenuation}
Let $\tau_b\in(0,1]$ be constant and let the live parameters fluctuate around a fixed point,
$\theta_m=\theta_\infty+\bm{\zeta}_m$, with independent zero-mean $\bm{\zeta}_m$ of common covariance $\Sigma$, and
let the initial target be deterministic. Then $\E[\bar{\theta}_m]\to\theta_\infty$ and
$\operatorname{Cov}(\bar{\theta}_m)\to\frac{\tau_b}{2-\tau_b}\,\Sigma$ as $m\to\infty$.
\end{proposition}

\begin{remark}\label{rem:ema}
With $n_b=256$, $\tau_b=1-0.9^{0.256}\approx0.027$. At this value, Proposition~\ref{prop:attenuation} reduces the standard
deviation of the update noise in the scorers' parameters by the factor $\sqrt{\tau_b/(2-\tau_b)}\approx0.12$
(by $0.23$ for the risk head, $\tau_b=0.1$), while Proposition~\ref{prop:tracking} bounds the lag behind a
drifting live model by $(1-\tau_b)\Delta_\theta/\tau_b\approx37\,\Delta_\theta$, about $37$ bursts of drift. The EMA does
not slow a consistent drift of the live parameters, which in the steady state of
Proposition~\ref{prop:tracking} it follows at the same speed, only later; what it removes is the
burst-to-burst noise of the updates, which is what would make consecutive plans inconsistent.
\end{remark}

\begin{algorithm}[t]
\caption{Online adaptation loop.}
\label{alg:online}
\begin{algorithmic}[1]
\Require offline corpus $\mathcal{D}$, world model $\theta$ from Section~\ref{sec:method:wm}, EMA copies $\bar{\theta}\leftarrow\theta$ of the scorers and the risk head
\State buffer $\mathcal{B}\leftarrow\emptyset$, credit $\leftarrow0$
\While{fewer than $10^{6}$ environment steps collected}
  \State drive $n_b=256$ steps with the planner of Section~\ref{sec:method:planner} (live dynamics, EMA scorers, EMA risk head); append the transitions to $\mathcal{B}$, which keeps only the most recent ones
  \State credit $\leftarrow$ credit $+\,0.125\,n_b$;\; $U\leftarrow\lfloor\text{credit}\rfloor$;\; credit $\leftarrow$ credit $-U$
  \For{$U$ updates}
    \State sample a batch of windows, half from $\mathcal{D}$ and half from $\mathcal{B}$, half of the latter from the latest burst
    \State update $D_\theta$ with Eq.~\eqref{eq:online_dyn}, the scorers with Eq.~\eqref{eq:online_scorer}, the risk head with $\mathcal{L}_{\mathrm{bce}}+\mathcal{L}_{\mathrm{mono}}$
  \EndFor
  \State $\bar{\theta}\leftarrow(1-\tau_b)\bar{\theta}+\tau_b\theta$ \Comment{Eq.~\eqref{eq:ema}}
\EndWhile
\State \Return live dynamics, EMA scorers, EMA risk head, frozen encoder / prior / value ensemble
\end{algorithmic}
\end{algorithm}

\section{Experiments}
\label{sec:experiments}

After the setup in both simulators (Section~\ref{sec:exp:setup}), the experiments ask four questions. What do
planning with the world model, the risk term, online adaptation and the data flywheel each contribute in closed loop
(Section~\ref{sec:exp:ablation})? How does the final system compare with the baselines, in CARLA
(Section~\ref{sec:exp:comparison}) and on the safety-critical benchmark of CAT in MetaDrive
(Section~\ref{sec:exp:metadrive})? Which
of the states that the ego can reach within the horizon does the world model expect to be unsafe, and how reliable
is this expectation
(Section~\ref{sec:exp:risk})? Does the imagination of the world model carry the information on which driving
decisions depend (Section~\ref{sec:exp:coc})? Section~\ref{sec:exp:efficiency} then reports the cost of training and
planning.

\subsection{Experimental setup}
\label{sec:exp:setup}

SAFER is evaluated in CARLA and in MetaDrive (Fig.~\ref{fig:env}); this subsection gives the setup in each.

\subsubsection{CARLA}
\label{sec:exp:setup:carla}

\begin{figure}[tbp]
  \centering
  \includegraphics[width=\linewidth]{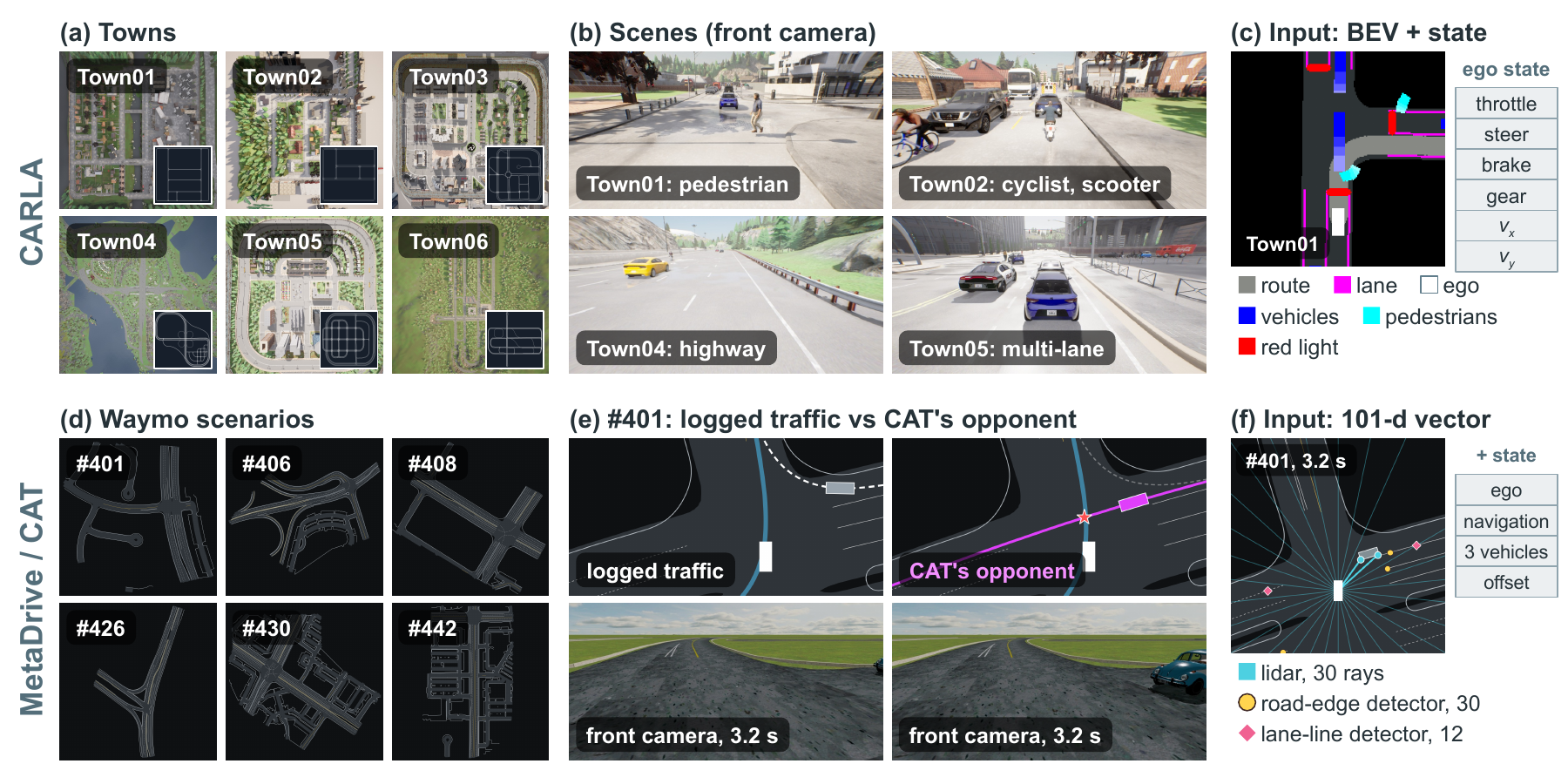}
  \caption{Experimental environments. Top: CARLA. (a)~The six towns of the evaluation, aerial view with the lane
  network inset; the corpus is collected in Town01, Town03, Town04 and Town06, while Town02 and Town05 contribute
  no training data. (b)~Scenes of evaluation episodes in a front camera added for illustration; the Town04 scene
  lies on a held-out test route. (c)~The agent's input in the Town01 scene of (b): the bird's-eye view, $38$\,m
  wide (route, lanes, vehicles and pedestrians over four past frames, fading, and traffic lights at their stop
  lines; the ego is the white box), and the ego state, i.e.\ the last applied control and the planar velocity.
  Bottom: MetaDrive with CAT. (d)~Road layouts of six Waymo test scenarios. (e)~Test scenario $401$ at $3.2$\,s in
  its logged traffic (left) and against CAT's opponent (right): top-down views, heading up, with the ego's logged
  route (blue), the opponent's logged path (white, dashed) and the path CAT generates for it (magenta), which
  crosses the ego's route (star); below, the front camera at the same moment. (f)~The agent's $101$-dimensional
  input at the right-hand moment of (e): $72$ distances, drawn as rays, to vehicles along $30$ directions (lidar,
  cyan), to road edges along $30$ (yellow) and to lane lines along $12$ (pink), and $29$ state entries: the ego's
  state, navigation features, the relative states of the three nearest vehicles and the lateral offset from the
  logged trajectory.}
  \label{fig:env}
\end{figure}

\paragraph{Simulator, expert and behavior policies}
We use CARLA \citep[version 0.9.15;][]{dosovitskiy2017carla}, and the agent observes only the privileged BEV mask stack and the ego state of
Section~\ref{sec:method:setting} (Fig.~\ref{fig:env}c). The expert is the publicly released Roach PPO agent~\citep{zhang2021roach}
after $11.8$\,M environment steps. Its actor provides the action labels $\act^{e}_t$, its critic the value
labels $v^{e}_t$, and its feature extractor gives the architecture of the
world-model encoder and initializes it. The intermediate drivers are
the $M=13$ checkpoints of a Roach agent that we trained from scratch with Roach's code and hyper-parameters, saved between $0.2$ and $4$\,M training steps, so that the bank spans
a wide range of competence rather than one level of it.

\paragraph{Offline corpus}
The three sets of Section~\ref{sec:method:data}, summarized in Table~\ref{tab:data}, are collected in the four training towns of Roach's benchmark, Town01, Town03, Town04 and Town06
(Fig.~\ref{fig:env}a), on unbounded random routes with randomized traffic density; every checkpoint
contributes the same number of transitions to $\mathcal{D}_{\mathrm{int}}$.  Two scripted events are used; their durations and the parameters of the action noise of
Eq.~\eqref{eq:noise} are listed in Table~\ref{tab:hparams} in \ref{app:impl}. A \emph{hold} fires when the ego has been stopped and the reward's desired
speed jumps up, as when a light turns green or a lead vehicle moves off, and holds full brake
while the expert would already be starting. A \emph{late brake} fires when the expert brakes for a hazard in the
ego's path and holds the previous throttle, but only if the expert could still stop before the hazard after the
override. An episode terminates at the first collision, red-light or stop-sign
violation, or when the ego is blocked or leaves its lane, so every safety event ends its episode in this corpus. The reward subtracts a terminal penalty, its fifth
factor, of $1$ at any termination and $1+v_t$, with the ego speed $v_t$ in m/s, at a safety violation, so that a fast
collision is penalized more than a slow one. The corpus contains the effect that a planner must be able to
foresee. Among the states in which the expert brakes for a hazard ahead, the collision rate within the next
$1.6$\,s is roughly half as high when the executed action brakes as when it does not.

\begin{table}[tbp]
  \centering
  \caption{Offline corpus in CARLA. Steps are transitions at $10$\,Hz. Episodes end on the first safety
  violation; the intermediate policies therefore produce many short episodes. ``Perturbed'' counts steps whose
  executed action differs from the behavior policy's proposal. The lower block lists the sets that one turn of the
  data flywheel adds (Section~\ref{sec:exp:ablation}); in $\mathcal{D}_{\mathrm{tgt}}^{+}$, scripted pedestrians step
  into the ego's path.}
  \label{tab:data}
  \footnotesize
  \setlength{\tabcolsep}{4pt}
  \begin{tabularx}{\linewidth}{l>{\raggedright\arraybackslash}Xrrrrr}
    \toprule
    Set & Driver & Steps & Episodes & Collisions & Red lights & Perturbed \\
    \midrule
    $\mathcal{D}_{\mathrm{exp}}$ & expert                    & $10.00$\,M & $1{,}901$  & $1{,}669$  & $133$     & $0\%$    \\
    $\mathcal{D}_{\mathrm{int}}$ & 13 intermediate checkpoints + noise & $10.05$\,M & $52{,}023$ & $28{,}629$ & $10{,}006$ & $21.6\%$ \\
    $\mathcal{D}_{\mathrm{tgt}}$ & expert + noise + events (two runs) & $2.00$\,M  & $338$      & $268$      & $46$      & $24.8\%$ \\
    \midrule
    total & & $22.06$\,M & $54{,}262$ & $30{,}566$ & $10{,}185$ & \\
    \midrule
    \multicolumn{7}{l}{\emph{added by one turn of the data flywheel}} \\
    $\mathcal{D}_{\mathrm{int}}^{+}$ & 13 intermediate checkpoints + noise & $4.00$\,M & $20{,}931$ & $11{,}439$ & $4{,}000$ & $21.8\%$ \\
    $\mathcal{D}_{\mathrm{tgt}}^{+}$ & expert + noise + events, crossing pedestrians (two runs) & $2.50$\,M & $2{,}640$ & $2{,}551$ & $54$ & $22.0\%$ \\
    \midrule
    total & & $28.56$\,M & $77{,}833$ & $44{,}556$ & $14{,}239$ & \\
    \bottomrule
  \end{tabularx}
\end{table}

\paragraph{Offline training}
Both offline stages train on the union of the three sets with uniform sampling, with no re-weighting between the expert, intermediate and targeted sets.  Stage~1 trains the world model, the reward, continuation and value
heads and the action prior for $1.0$\,M steps, together with the scene-factor head of
Section~\ref{sec:exp:coc} as an auxiliary task, whose loss is RMS-normalized like the others and added to
Eq.~\eqref{eq:loss_total} with weight $0.5$. Stage~2 adds the risk head, drops the scene-factor head and continues all
parameters jointly for $0.5$\,M steps, with the additional windows
before terminals of Section~\ref{sec:method:wm}. The remaining settings are
listed in Table~\ref{tab:hparams} in \ref{app:impl}.

\paragraph{Online adaptation}
The online loop (Algorithm~\ref{alg:online}) drives the four training towns of the
corpus under its termination rules, using the planner of
Section~\ref{sec:method:planner} with the risk penalty ($W=50$, pricing collisions only) and scoring against the EMA copies. It runs for $10^{6}$ environment steps, $27.8$\,h of driving.

\paragraph{Baselines}
Besides the published results of end-to-end driving methods (Section~\ref{sec:exp:comparison}), we train four
reinforcement learning methods offline on the whole corpus of Table~\ref{tab:data}, including the sets that the data flywheel adds: TD3+BC~\citep{fujimoto2021td3bc},
IQL~\citep{kostrikov2022iql}, TD-MPC2~\citep{hansen2024tdmpc2} and DreamerV3~\citep{hafner2023dreamerv3}. They
observe the BEV and the ego state through the frozen encoder of the Roach expert, are trained with the published
settings of each method, adapted to the corpus where necessary, and are evaluated under the protocol below.

\paragraph{Closed-loop evaluation protocol}
Evaluation follows the route-based protocol of the CARLA Leaderboard~\citep{carla2020leaderboard} as
implemented in Roach's benchmark. The ego must follow a predefined route through dense traffic and is scored
on how much of the route it completes and on the infractions it commits on the way. We use all $76$ routes of the benchmark, between $0.5$ and $4.9$\,km long. They are $50$ routes in the four training towns, the $10$
held-out test routes of Town04 and $16$ routes in Town02 and Town05, from which no training data come. Traffic
is the benchmark's fixed density per town. Unlike the corpus, an evaluation episode does not end
at a collision or a run red light. It ends when the route is completed, when the ego has been blocked for
$90$\,s, when it deviates from the route or after $1{,}500$\,s of simulation time, and every infraction on
the way is recorded. The \emph{driving score} of an episode is the product of the route completion $\mathrm{RC}$
($1$ for a completed route, otherwise the completed fraction of its length) and the infraction penalty
\begin{equation}
  P=\big(1-\ell_{\mathrm{off}}\big)\;0.65^{\,n_{\mathrm{lay}}}\;0.60^{\,n_{\mathrm{veh}}}\;0.50^{\,n_{\mathrm{ped}}}\;
    0.70^{\,n_{\mathrm{red}}}\;0.80^{\,n_{\mathrm{stop}}},
  \label{eq:driving_score}
\end{equation}
where $\ell_{\mathrm{off}}$ is the share of the completed distance driven outside the route's lane and
$n_{\mathrm{lay}}$, $n_{\mathrm{veh}}$, $n_{\mathrm{ped}}$, $n_{\mathrm{red}}$ and $n_{\mathrm{stop}}$ count
collisions with the static layout, with vehicles and with pedestrians, red lights run and stop signs
ignored. Besides the driving score we report the completion rate, infractions per $100$\,km by type, the
rate of blocked episodes, and two action-smoothness measures, the mean absolute change of the
executed acceleration and steering commands between consecutive steps, because a planner that re-optimizes
at every step can be jerky in ways the driving score does not see.  Every controller is evaluated on all $76$ routes under three seeds, $228$ episodes per controller. The seed sets the traffic and the planner's sampling noise, and all controllers start from the same
traffic.

\subsubsection{MetaDrive}
\label{sec:exp:setup:md}

\paragraph{Benchmark}
We use the closed-loop benchmark of CAT~\citep{zhang2023cat}, which consists of $500$ scenarios of the Waymo Open Motion
Dataset~\citep{ettinger2021womd} replayed in MetaDrive~\citep{li2023metadrive}, $400$ for training and $100$ for testing, observed through the $101$-dimensional vector of
Section~\ref{sec:method:setting} (Fig.~\ref{fig:env}d--f). Every test scenario is driven twice, first in its logged
traffic (\emph{regular}) and then with an opponent that CAT's generator places on the predicted trajectory most likely
to collide with the ego's regular drive (\emph{safety-critical}, Fig.~\ref{fig:env}e); the first vehicle contact ends
the episode. The metrics are those of CurricuVLM~\citep{sheng2026curricuvlm}: episode reward, road completion, total
distance, crash rate (the share of episodes that end in a vehicle contact) and average speed.

\paragraph{Data and model}
The offline corpus applies the recipe of Section~\ref{sec:method:data} to the $400$ training scenarios under
CAT's training rules, in which a vehicle contact costs $-1$ without ending the episode. The expert is a CAT agent, TD3 trained with CAT's closed-loop adversarial training and CAT's code, for $10^{6}$ steps, and eight of its checkpoints, saved between $2.5\times10^{4}$ and
$7.5\times10^{5}$ steps, are the intermediate drivers, perturbed by the noise process of
Section~\ref{sec:method:data} with the parameters of the CARLA corpus (Section~\ref{sec:exp:setup:carla}). Every scenario is driven in its logged traffic and again against CAT's opponent:
by the intermediate drivers, by the noise-free expert,
which also faces the opponents generated against the intermediate drivers, and by the expert
with targeted hazard brakes, in which the ego brakes for $0.5$--$1$\,s when a road user ahead on its route closes
in and the expert then resumes. CAT's expert seldom brakes for hazards, so this event takes the place of the
late brake of CARLA as the source of braking near hazards. The corpus holds $44{,}800$ episodes and $3.2$\,M
transitions; the expert labels every state with its action, and the value labels are the noise-free expert's
values, estimated by policy evaluation on the corpus with CAT's discount of $0.99$, which the planner
also uses as $\gamma$. The world model is that of
Section~\ref{sec:method:wm}, with an observation encoder initialized from the first two layers of the expert's
actor and CAT's reward components as reward factors; dynamics, heads, action prior and risk head are trained
together in a single stage of $2.5\times10^{5}$ steps.

\paragraph{Planner and online adaptation}
On this benchmark the risk head predicts three events, vehicle contact, contact with an object and leaving the
road, and the planner weights all three, $\bm{\omega}=(1,1,1)$ in Eq.~\eqref{eq:risk_penalty}, since leaving the road
also ends an episode; $W=50$ as in CARLA. Two settings differ from CARLA because of the benchmark. CAT's traffic
replays the logs and does not react to the ego, so slowing down can cause a rear-end contact, and the
braking-monotonicity term of Eq.~\eqref{eq:risk_mono} is not used during online adaptation. The action prior,
distilled from CAT's TD3 expert, often acts at the bounds of the action range, where a plan carried over from the
previous step lags behind it, so every MPPI step starts from the prior's current chunk instead of the shifted
previous plan of Section~\ref{sec:method:planner}. SAFER is evaluated after $10^{6}$ steps of online adaptation
on the training scenarios. We train three SAFER systems with different seeds, offline and online, evaluate each
three times, and report the mean and standard deviation over the three systems.

\subsection{Closed-loop ablation}
\label{sec:exp:ablation}

Five controllers share the stage-2 checkpoint and differ only in how it is used, and a sixth arm retrains it:
\begin{itemize}
  \item \textbf{BC prior}: the behavior-cloned prior $\pi_\theta$ of Eq.~\eqref{eq:policy} executed
        directly, no planning;
  \item \textbf{MPPI}: planning with the reward, continuation and value heads and no risk penalty ($W=0$),
        i.e.\ SimDist's planner on our model;
  \item \textbf{+ risk term}: the same planner with the risk penalty, $W=50$;
  \item \textbf{+ online, dynamics only}: the same planner on a model whose dynamics $D_\theta$ alone are adapted
        online (Eq.~\eqref{eq:online_dyn}), for $10^{6}$ steps and otherwise as in the full system, while the reward
        scorers and the risk head keep their offline weights, i.e.\ SimDist's adaptation rule on our model;
  \item \textbf{+ online adaptation}: the same planner on the model adapted online for $10^{6}$ steps, the full
        SAFER system. The snapshot of the same run after $5\times10^{5}$ steps is evaluated as well, for the effect of
        training length;
  \item \textbf{+ data flywheel}: one turn of a loop that aims new data at the scenarios that an analysis of the offline corpus finds covered least and retrains the
        system. These are encounters with pedestrians. Of the $21{,}287$ recorded collisions analyzed
        in Section~\ref{sec:exp:risk}, only $2{,}896$ involve a pedestrian, and the risk head predicts them at the step
        at which they happen less often than collisions with vehicles (Fig.~\ref{fig:risk_timeline}b). The corpus is
        therefore extended by $6.5$\,M transitions, collected like the rest of it on random routes in the
        training towns. Of these, $4.0$\,M come from the intermediate drivers, and $2.5$\,M from the
        expert with noise and late brakes, facing scripted pedestrians that step into the ego's path on straight
        roads and at the exits of left turns (Table~\ref{tab:data}). Both offline stages are retrained with the same
        recipe, and the new model is adapted online for $10^{6}$ steps with the full system's settings.
\end{itemize}

Table~\ref{tab:ablation} adds these components one at a time, from the behavior-cloned prior to the adapted
planner and one turn of the data flywheel. Differences between arms are paired over the same route and seed; their $95\%$ intervals come from a
bootstrap that resamples routes and keeps the three seeds of a route together. The
$\Delta$DS column of Table~\ref{tab:ablation} gives them for the driving score; a difference is called
significant when its interval excludes zero.

\begin{table}[tbp]
  \centering
  \caption{Closed-loop ablation on the $76$ Leaderboard routes: mean $\pm$ standard
  deviation over three seeds ($76$ episodes per seed, $228$ per arm). (a)~DS: driving score; RC: route
  completion; IP: infraction penalty; Compl.: routes completed; Clean: routes completed without a collision;
  $\Delta$DS: change in driving score from the row above (for all three online rows, from + risk term), paired over
  route and seed, with its $95\%$ bootstrap interval over routes. The last two online rows are the snapshots of one online
  run after $5\times10^{5}$ and $10^{6}$ environment steps; the first online row is a separate run of $10^{6}$ steps
  that adapts the dynamics alone. The last row is one turn of the data flywheel, in which the model is retrained on the corpus
  extended by the lower block of Table~\ref{tab:data} and adapted online for $10^{6}$ steps. (b)~Collisions by type, red lights run and stop signs run per $100$\,km; Blocked: episodes
  ended by blocking; smoothness: mean absolute step-to-step change of the executed acceleration $|\Delta a|$
  and steering $|\Delta\delta|$ commands.}
  \label{tab:ablation}
  \footnotesize
  \setlength{\tabcolsep}{3pt}
  (a)~Route level\\[2pt]
  \begin{tabular}{lcccccc}
    \toprule
    Arm & DS & RC & IP & Compl.\ (\%) & Clean (\%) & $\Delta$DS \\
    \midrule
    BC prior            & $0.647\pm0.021$ & $0.921\pm0.038$ & $0.701\pm0.026$ & $85.1\pm4.2$ & $38.6\pm6.5$ & -- \\
    MPPI                & $0.620\pm0.034$ & $0.965\pm0.009$ & $0.639\pm0.036$ & $91.7\pm1.5$ & $38.2\pm4.7$ & $-0.02$$8$ $[-0.078, +0.02$$1$$]$ \\
    + risk term         & $0.702\pm0.018$ & $0.978\pm0.019$ & $0.715\pm0.014$ & $94.7\pm3.5$ & $50.4\pm5.3$ & $+0.082$ $[+0.03$$4$$, +0.131]$ \\
    + online, dynamics only & $0.719\pm0.036$ & $0.959\pm0.004$ & $0.750\pm0.037$ & $90.8\pm1.3$ & $48.7\pm6.0$ & $+0.017$ $[-0.028, +0.061]$ \\
    + online, $5{\times}10^{5}$ steps & $0.740\pm0.036$ & $0.944\pm0.006$ & $0.788\pm0.040$ & $88.6\pm0.8$ & $54.4\pm9.0$ & $+0.038$ $[-0.013, +0.08$$9$$]$ \\
    + online, $10^{6}$ steps & $0.768\pm0.029$ & $0.967\pm0.008$ & $0.796\pm0.023$ & $93.4\pm1.3$ & $57.5\pm6.1$ & $+0.066$ $[+0.01$$3$$, +0.11$$9$$]$ \\
    + data flywheel & $0.801\pm0.038$ & $0.950\pm0.014$ & $0.843\pm0.031$ & $89.0\pm3.8$ & $61.8\pm7.3$ & $+0.033$ $[-0.010, +0.076]$ \\
    \bottomrule
  \end{tabular}\\[8pt]
  (b)~Infractions and smoothness\\[2pt]
  \begin{tabular}{lcccccccc}
    \toprule
    & \multicolumn{5}{c}{per $100$\,km} & & \multicolumn{2}{c}{Smoothness} \\
    \cmidrule(lr){2-6}\cmidrule(lr){8-9}
    Arm & Vehicle & Pedestrian & Static & Red light & Stop sign & Blocked (\%) & $|\Delta a|$ & $|\Delta\delta|$ \\
    \midrule
    BC prior            & $27.9\pm4.8$ & $19.5\pm3.7$ & $0.8\pm0.8$ & $1.3\pm0.4$ & $0.5\pm0.5$ & $14.9\pm4.2$ & $0.038\pm0.002$ & $0.004\pm0.000$ \\
    MPPI                & $31.7\pm4.4$ & $20.6\pm5.0$ & $0.3\pm0.4$ & $5.7\pm0.5$ & $8.0\pm1.2$ & $7.0\pm2.0$  & $0.244\pm0.002$ & $0.091\pm0.001$ \\
    + risk term         & $28.3\pm4.1$ & $12.7\pm0.3$ & $0.3\pm0.4$ & $4.3\pm1.6$ & $4.6\pm0.7$ & $4.8\pm4.2$  & $0.243\pm0.003$ & $0.092\pm0.002$ \\
    + online, dynamics only & $22.7\pm4.2$ & $14.7\pm3.4$ & $0.8\pm0.8$ & $1.0\pm0.5$ & $1.3\pm0.4$ & $8.3\pm1.5$  & $0.231\pm0.001$ & $0.074\pm0.001$ \\
    + online, $5{\times}10^{5}$ steps & $20.7\pm4.3$ & $10.8\pm3.7$ & $0.8\pm0.8$ & $2.1\pm0.5$ & $1.8\pm1.2$ & $11.0\pm0.8$ & $0.192\pm0.000$ & $0.069\pm0.000$ \\
    + online, $10^{6}$ steps & $20.6\pm3.5$ & $8.8\pm2.6$  & $0.0\pm0.0$ & $1.3\pm0.4$ & $3.1\pm0.8$ & $6.6\pm1.3$  & $0.194\pm0.002$ & $0.074\pm0.000$ \\
    + data flywheel & $16.0\pm2.4$ & $5.5\pm1.3$ & $0.5\pm0.9$ & $1.6\pm1.6$ & $1.8\pm0.4$ & $11.0\pm3.8$ & $0.172\pm0.001$ & $0.059\pm0.000$ \\
    \bottomrule
  \end{tabular}
\end{table}

Replacing the prior by the planner without a risk term does not change the driving score significantly. The
planner completes more of each route and is blocked less often ($7.0\%$ of the episodes against $14.9\%$), but it
runs more red lights ($5.7$ against $1.3$ per $100$\,km), so its infraction penalty falls, and its acceleration
command changes six times more from one step to the next.

Adding the risk penalty raises the driving score by $0.08$, and the gain comes from the infraction penalty. Collisions
with pedestrians fall from $20.6$ to $12.7$ per $100$\,km, and the share of routes completed without a
collision rises from $38\%$ to $50\%$.

The model adapted on the planner's own driving raises the driving score by a further $0.07$, again through
the infraction penalty, while route completion stays at the level of the offline planner
($0.967$ against $0.978$). Every type of infraction becomes
rarer. Per $100$\,km, vehicle collisions fall from $28.3$ to $20.6$, pedestrian collisions from $12.7$ to $8.8$ and
red lights run from $4.3$ to $1.3$, and the commands become smoother ($|\Delta a|$ from $0.243$ to $0.194$). The
gain is concentrated where the online loop drives. The driving score rises by $0.07$ on the $50$ routes of the four training towns and by $0.16$ on the ten held-out
Town04 routes, too few to be conclusive, but not in the two towns without training data. Online adaptation thus extends the coverage of the towns
in which the planner drives during adaptation; in the two towns where it never drives, the driving score is
unchanged.

Over the three steps, the full system raises the driving score of the behavior-cloned prior by $0.12$,
through both a higher route completion and a higher infraction penalty. The standard deviation of the driving
score over seeds lies between $0.018$ and $0.038$ for every row of Table~\ref{tab:ablation}, below the size of each
significant step.

Halfway through the online run, after $5\times10^{5}$ steps, the safety gain is largely in place. The infraction
penalty is already $0.07$ above the offline planner's, vehicle collisions are at $20.7$ per $100$\,km ($20.6$ after
$10^{6}$ steps) and red lights run at $2.1$. This snapshot is more cautious, though. It is blocked in $11.0\%$ of
the episodes ($4.8\%$ offline, $6.6\%$ after $10^{6}$ steps) and completes less of each route, so its gain in
driving score over the offline planner is smaller and not significant. In the second half of the run, route
completion rises again, from $0.944$ to $0.967$, while the infraction penalty holds, which gives the full
system's gain.

Besides how long the loop runs, the gain depends on what it adapts. When the loop adapts the dynamics alone, as
SimDist does, and leaves the reward scorers and the risk head at their offline weights, the driving score rises by
only $0.02$ over the offline planner, which is not significant, and stays $0.05$ below the full system's, a
significant difference that comes from the infraction penalty. The adapted dynamics alone make vehicle collisions and red lights run rarer ($22.7$ and $1.0$ per $100$\,km,
against $28.3$ and $4.3$ offline), but pedestrian collisions do not fall ($14.7$ against $12.7$) and the planner
is blocked more often ($8.3\%$ of the episodes against $4.8\%$). The significant gain of online adaptation
therefore requires adapting the reward scorers and the risk head along with the dynamics, as SAFER does
(Section~\ref{sec:method:online}).

One turn of the data flywheel makes the system safer. Its infraction penalty rises by $0.05$, a significant
gain. Pedestrian collisions fall from $8.8$ to $5.5$ per $100$\,km and vehicle collisions from $20.6$ to $16.0$.
The retrained system is also more cautious. It is blocked in $11.0\%$ of the episodes, against $6.6\%$, and
completes less of each route ($0.950$ against $0.967$), so its driving score rises by $0.03$, which is not
significant.

\begin{figure}[tbp]
  \centering
  \includegraphics[width=\linewidth]{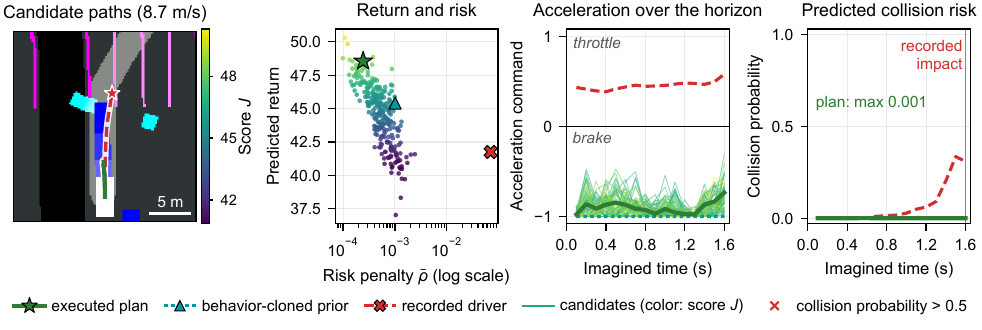}\\[1pt]
  {\small (a) recorded collision with the vehicle ahead}\\[8pt]
  \includegraphics[width=\linewidth]{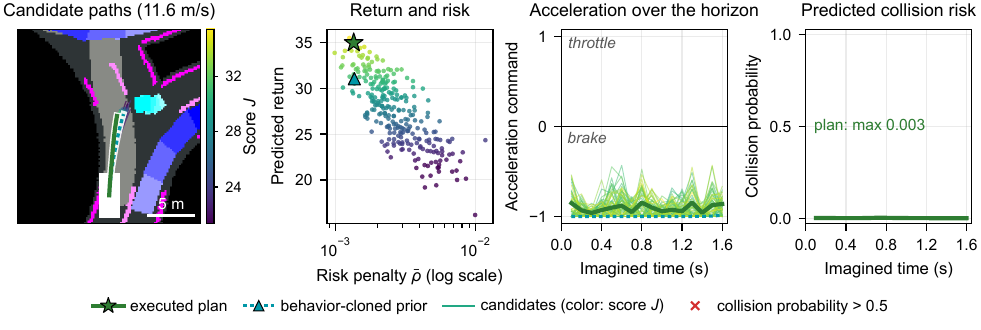}\\[1pt]
  {\small (b) dense traffic, no collision}
  \caption{Two planning decisions of the offline planner on recorded states from
  $\mathcal{D}_{\mathrm{int}}$. From left to right: the BEV at the decision time (route light gray, road dark gray, lane markings magenta, vehicles blue,
  pedestrians cyan, lighter shades are earlier frames; ego white)
  with the $270$ scored candidates drawn as paths and colored by their score $J$, the executed plan (green),
  the behavior-cloned prior (dotted) and, in (a), the recorded driver's actions up to the impact (dashed red,
  star), and a red cross where a candidate's predicted collision probability first exceeds $0.5$; paths are sketched from the action chunks with a kinematic bicycle model and the predicted speed, not
  decoded positions; predicted return against risk penalty $\bar{\rho}$ (log scale) for all candidates, with
  the plan (star), the prior (triangle) and, in (a), the recorded actions (cross); the acceleration
  command of the $60$ best candidates, the plan, the prior and the recorded
  actions over the imagined horizon (negative values brake);  and the predicted probability that a collision has occurred by each imagined step.}
  \label{fig:planner}
\end{figure}

Fig.~\ref{fig:planner} opens up two single decisions of the offline planner on recorded states. Each
of the $270$ candidate chunks scored in the planner's two iterations is decomposed into its predicted
return, the first two terms of Eq.~\eqref{eq:score}, and its risk penalty $\bar{\rho}$ of Eq.~\eqref{eq:risk_penalty}.

In the first state (Fig.~\ref{fig:planner}a) the ego drives at $8.7$\,m/s towards a vehicle ahead in its lane.
The recorded driver kept accelerating and hit the vehicle $1.6$\,s later, although the expert's label at that
moment was to brake. The planner brakes over the whole horizon. Its executed plan scores $J=48.5$, against
$45.4$ for the behavior-cloned prior and $38.2$ for the recorded actions, and its predicted collision
probability does not exceed $0.001$. For the recorded actions, by contrast, the risk head's probability rises
from about $0.8$\,s ahead and reaches $0.33$ just before the recorded impact. The same holds beyond this
example. Over $200$ recorded collisions of this kind, the executed plan brakes harder than the recorded driver
in $98.5\%$ of them and carries a lower predicted risk up to the impact in $98.0\%$ (mean $0.013$, against
$0.166$ for the recorded actions and $0.016$ for the prior).

In the second state (Fig.~\ref{fig:planner}b) the ego drives at $11.6$\,m/s through a junction with vehicles
on both sides and pedestrians ahead to the right, and no collision is imminent. The risk head stays quiet. The
plan's predicted collision probability never exceeds $0.003$, and no candidate reaches a risk penalty of
$0.025$. Candidates that turn towards the pedestrians or into the traffic receive the lowest scores; the plan
keeps its lane and slows down, and it improves on the prior's score ($34.9$ against $31.0$) at the same
predicted risk. The planner therefore responds to the situation rather than to the density of the scene. The
risk term rises before a real collision, changes the plan there, and stays silent where none threatens.

\subsection{Comparison on the CARLA Leaderboard}
\label{sec:exp:comparison}

Table~\ref{tab:comparison} compares SAFER with end-to-end driving baselines and with the four offline
reinforcement learning baselines of Section~\ref{sec:exp:setup:carla} on the CARLA Leaderboard. SAFER observes the privileged bird's-eye view of Roach and is
scored with the Leaderboard's metrics on the $76$ Leaderboard routes of Section~\ref{sec:exp:setup:carla}. The
comparison therefore places SAFER's driving decisions among those of the baselines; it does not rank complete
pipelines from sensors to control. The offline reinforcement learning baselines observe what SAFER observes and
are scored on the same routes under the same seeds, so they compare with SAFER directly.

SAFER reaches the highest route completion ($0.95$) and the second-highest driving score ($0.80$), after DreamerV3
($0.82$). The strongest end-to-end
baselines trade route completion for infraction penalty. TCP, InterFuser and TF++ have the highest infraction penalties among them
($0.84$--$0.87$) but complete $79$--$88\%$ of the route length, while LAV completes $94\%$ at an infraction
penalty of $0.64$. SAFER completes $95\%$ of each route at an infraction penalty of
$0.84$, that of InterFuser and TF++ and below TCP's,
so its lead in driving score over them comes from completing more of each route at a comparable infraction penalty. The ablation shows where this pairing comes from
(Table~\ref{tab:ablation}). Planning with the world model alone completes $96.5\%$ of each route at an infraction
penalty of $0.64$, the risk term and online adaptation raise the penalty to $0.80$ while completion stays
at $0.97$, and one turn of the data flywheel raises it to $0.84$ at a completion of $0.95$.

The four methods trained on SAFER's corpus reach driving scores of $0.82$ (DreamerV3), $0.73$ (IQL), $0.63$ (TD3+BC) and
$0.43$ (TD-MPC2), against
$0.80$ for SAFER. Paired over routes and seeds, DreamerV3 and SAFER do not differ significantly in any of the
three metrics. SAFER's lead over IQL is significant in driving score and in route
completion, while their infraction penalties do not differ significantly. SAFER's lead over TD-MPC2 is significant in all
three metrics. TD-MPC2 is blocked in $53\%$ of its episodes, against $11\%$ for SAFER.

Table~\ref{tab:sameinput} explores further metrics for the methods whose input is exactly that of SAFER, the
bird's-eye view and the ego state. SAFER has the best value in six of the eight measures: the highest reward per step ($0.77$, against at most $0.71$ for the
baselines), the highest average speed ($3.31$\,m/s) and the largest share of completed routes ($89\%$), and
the fewest collisions with vehicles, red lights run and blocked episodes. DreamerV3
collides with pedestrians less often than SAFER ($1.1$ against $5.5$ per
$100$\,km) and runs no stop signs, but it falls behind SAFER in the other six measures. IQL also collides with
pedestrians less often than SAFER ($4.2$ against $5.5$ per $100$\,km) and TD-MPC2 runs fewer stop signs.

\begin{table}[htbp]
  \centering
  \caption{Comparison with the baselines on the CARLA Leaderboard. The end-to-end methods (top) are given with their published results. TD3+BC, IQL, TD-MPC2 and DreamerV3 (middle) are trained by us on SAFER's corpus, with the frozen encoder of the expert and without its labels, and evaluated as SAFER.}
  \label{tab:comparison}
  \footnotesize
  \setlength{\tabcolsep}{4pt}
  \begin{tabularx}{\linewidth}{>{\raggedright\arraybackslash}Xlccc}
    \toprule
    Algorithm & Input & Driving & Route & Infraction \\
              &                & score $\uparrow$ & compl.\ $\uparrow$ & penalty $\uparrow$ \\
    \midrule
    CILRS \citep{codevilla2019limitations} & RGB & 0.05 & 0.14 & 0.55 \\
    LBC \citep{chen2020lbc} & RGB & 0.09 & 0.18 & 0.73 \\
    TransFuser \citep{prakash2021transfuser} & RGB + LiDAR & 0.17 & 0.52 & 0.42 \\
    NEAT \citep{chitta2021neat} & RGB & 0.22 & 0.42 & 0.65 \\
    MMFN \citep{zhang2022mmfn} & RGB + LiDAR + HD map & 0.23 & 0.47 & 0.63 \\
    IARL \citep{toromanoff2020affordances} & RGB & 0.25 & 0.47 & 0.52 \\
    Rails \citep{chen2021worldonrails} & RGB & 0.31 & 0.58 & 0.56 \\
    GRIAD \citep{chekroun2023gri} & RGB & 0.37 & 0.62 & 0.60 \\
    Latent TransFuser \citep{chitta2023transfuser} & RGB & 0.45 & 0.66 & 0.72 \\
    TransFuser \citep{chitta2023transfuser} & RGB + LiDAR & 0.61 & 0.87 & 0.71 \\
    LAV \citep{chen2022lav} & RGB + LiDAR & 0.62 & 0.94 & 0.64 \\
    TF++ \citep{jaeger2023hiddenbiases} & RGB + LiDAR & 0.66 & 0.79 & 0.84 \\
    TCP \citep{wu2022tcp} & RGB & 0.75 & 0.86 & 0.87 \\
    InterFuser \citep{shao2023interfuser} & RGB + LiDAR & 0.76 & 0.88 & 0.84 \\
    \midrule
    TD-MPC2 \citep{hansen2024tdmpc2} & BEV + ego state & 0.43 & 0.70 & 0.59 \\
    TD3+BC \citep{fujimoto2021td3bc} & BEV + ego state & 0.63 & 0.84 & 0.76 \\
    IQL \citep{kostrikov2022iql} & BEV + ego state & 0.73 & 0.88 & 0.82 \\
    DreamerV3 \citep{hafner2023dreamerv3} & BEV + ego state & \textbf{0.82} & 0.94 & \textbf{0.88} \\
    \midrule
    SAFER (ours) & BEV + ego state &  0.80 &  \textbf{0.95} &  0.84 \\
    \bottomrule
  \end{tabularx}
\end{table}

\begin{table}[htbp]
  \centering
  \caption{Further metrics of the methods of Table~\ref{tab:comparison} whose input is exactly that of SAFER,
  under the same protocol: mean $\pm$ standard deviation over three seeds. Reward of Eq.~\eqref{eq:reward} per
  step, average speed (route distance completed over the duration of the episode), share of the routes completed,
  share of the episodes in which the ego is blocked, and collisions and traffic-rule infractions per $100$\,km driven.}
  \label{tab:sameinput}
  \footnotesize
  \setlength{\tabcolsep}{2.5pt}
  \begin{tabularx}{\linewidth}{@{}>{\raggedright\arraybackslash}Xcccccccc@{}}
    \toprule
    & & & & & \multicolumn{4}{c}{Per $100$\,km $\downarrow$} \\
    \cmidrule(lr){6-9}
    & Reward & Speed & Completed & Blocked & \multicolumn{2}{c}{Collisions} & Red & Stop \\
    \cmidrule(lr){6-7}
    Algorithm & per step $\uparrow$ & (m/s) $\uparrow$ & (\%) $\uparrow$ & (\%) $\downarrow$ & Pedestrians & Vehicles & lights & signs \\
    \midrule
    TD-MPC2 \citep{hansen2024tdmpc2} & $-0.51\pm0.18$ & $2.61\pm0.22$ & $43.0\pm8.0$ & $52.6\pm6.0$ & $5.6\pm1.4$ & $27.2\pm0.2$ & $3.8\pm2.0$ & $0.7\pm1.2$ \\
    TD3+BC \citep{fujimoto2021td3bc} & $0.53\pm0.02$ & $3.08\pm0.09$ & $68.0\pm3.0$ & $30.3\pm2.6$ & $8.4\pm2.6$ & $27.9\pm3.3$ & $4.7\pm0.6$ & $2.6\pm0.8$ \\
    IQL \citep{kostrikov2022iql} & $0.47\pm0.09$ & $3.20\pm0.05$ & $75.4\pm2.7$ & $21.5\pm2.7$ & $4.2\pm2.5$ & $19.3\pm4.3$ & $3.9\pm2.1$ & $3.9\pm0.9$ \\
    DreamerV3 \citep{hafner2023dreamerv3} & $0.71\pm0.01$ & $3.26\pm0.05$ & $88.2\pm2.6$ & $11.8\pm2.6$ & $\mathbf{1.1\pm0.9}$ & $17.1\pm6.0$ & $2.1\pm1.6$ & $\mathbf{0.0\pm0.0}$ \\
    \midrule
    SAFER (ours) & $\mathbf{0.77\pm0.02}$ & $\mathbf{3.31\pm0.03}$ & $\mathbf{89.0\pm3.8}$ & $\mathbf{11.0\pm3.8}$ & $5.5\pm1.3$ & $\mathbf{16.0\pm2.4}$ & $\mathbf{1.6\pm1.6}$ & $1.8\pm0.4$ \\
    \bottomrule
  \end{tabularx}
\end{table}

\subsection{Comparison on the MetaDrive/CAT benchmark}
\label{sec:exp:metadrive}

To compare SAFER with a wider range of baselines under a common protocol, we evaluate it on the CAT benchmark
of Section~\ref{sec:exp:setup:md}. Table~\ref{tab:metadrive} compares SAFER with
reinforcement learning (SAC \citep{haarnoja2018sac}, PPO
\citep{schulman2017ppo}, TD3 \citep{fujimoto2018td3}), safe reinforcement learning (SAC-PID and TD3-PID
\citep{stooke2020pid}, TD3-Lag \citep{ray2019safetygym}), imitation learning (BC \citep{gleave2022imitation},
GAIL \citep{ho2016gail}, AIRL \citep{fu2018airl}, SQIL \citep{reddy2020sqil}) and closed-loop training, which
generates or selects training scenarios against the current policy (CAT, CLIC \citep{niu2024clic} and
CurricuVLM), and with DreamerV3~\citep{hafner2023dreamerv3}, which we train on SAFER's offline corpus of
Section~\ref{sec:exp:setup:md} as in CARLA (Section~\ref{sec:exp:setup:carla}), on the $101$-dimensional
observation and without online interaction.

In regular traffic SAFER has the highest episode reward, road completion and total distance in
Table~\ref{tab:metadrive} and shares the highest average speed with PPO; its crash rate ($18.6\%$) is lower than
that of every baseline except SAC ($18.3\%$) and the three closed-loop training methods ($15.1$--$16.0\%$). In
safety-critical traffic it has the highest episode reward, road completion and average speed, and its crash rate
($31.4\%$) is close to those of SAC and CAT ($30.5\%$ and $32.1\%$), while CurricuVLM, which builds its training
curriculum from generated safety-critical scenarios, keeps the lowest crash rate ($25.1\%$) and a slightly longer
total distance ($48.4$ against $48.2$). DreamerV3, trained on the same corpus as SAFER, stays below SAFER in
episode reward, road completion, total distance and average speed in both settings. It crashes about as often as
SAFER in regular traffic ($19.4\%$ against $18.6\%$) and more often in safety-critical traffic ($44.3\%$ against
$31.4\%$), and it drives at about $7$\,m/s, against $10$\,m/s for SAFER.

\begin{table}[tbp]
  \centering
  \caption{Closed-loop comparison on the MetaDrive/CAT benchmark ($100$ test scenarios): mean $\pm$ standard
  deviation over three seeds. (a)~Regular: logged traffic; (b)~safety-critical: CAT's opponent generated against
  the regular drive. Baseline results are from \citet{sheng2026curricuvlm}, except those of DreamerV3, which we trained.}
  \label{tab:metadrive}
  \footnotesize
  \setlength{\tabcolsep}{4pt}
  (a)~Regular test scenarios\\[2pt]
  \begin{tabular}{lccccc}
    \toprule
    Method & Episode reward $\uparrow$ & Road completion (\%) $\uparrow$ & Total distance $\uparrow$ & Crash rate (\%) $\downarrow$ & Average speed $\uparrow$ \\
    \midrule
    SAC \citep{haarnoja2018sac} & $43.3\pm1.61$  & $67.9\pm1.14$ & $44.9\pm1.07$ & $18.3\pm0.92$ & $9.35\pm0.06$ \\
    PPO \citep{schulman2017ppo} & $41.1\pm2.48$  & $65.9\pm2.51$ & $42.9\pm2.01$ & $22.6\pm3.75$ & $\mathbf{10.1\pm0.24}$ \\
    TD3 \citep{fujimoto2018td3} & $49.3\pm0.83$  & $72.9\pm0.44$ & $47.9\pm1.08$ & $18.9\pm1.72$ & $8.14\pm0.80$ \\
    \midrule
    SAC-PID \citep{stooke2020pid} & $7.57\pm3.20$  & $24.8\pm10.3$ & $12.9\pm2.91$ & $21.3\pm1.35$ & $5.19\pm0.57$ \\
    TD3-Lag \citep{ray2019safetygym} & $39.6\pm0.00$   & $46.9\pm0.00$ & $38.9\pm0.00$ & $23.0\pm0.00$ & $5.50\pm0.00$ \\
    TD3-PID \citep{stooke2020pid} & $-1.00\pm0.91$ & $6.48\pm0.44$ & $4.55\pm0.37$ & $22.8\pm1.77$ & $4.17\pm1.08$ \\
    \midrule
    BC \citep{gleave2022imitation} & $29.7\pm2.52$  & $58.6\pm2.09$ & $33.4\pm1.80$ & $20.6\pm1.18$ & $7.42\pm0.28$ \\
    GAIL \citep{ho2016gail} & $32.5\pm1.99$  & $57.0\pm1.50$ & $35.1\pm1.82$ & $23.4\pm1.94$ & $7.84\pm0.74$ \\
    AIRL \citep{fu2018airl} & $27.0\pm1.94$  & $54.8\pm3.13$ & $32.0\pm1.63$ & $20.1\pm3.06$ & $9.80\pm0.76$ \\
    SQIL \citep{reddy2020sqil} & $20.3\pm0.21$  & $44.7\pm0.78$ & $25.1\pm0.36$ & $26.9\pm1.30$ & $6.50\pm0.38$ \\
    \midrule
    CAT \citep{zhang2023cat} & $49.1\pm3.89$  & $73.0\pm3.94$ & $48.5\pm3.39$ & $15.5\pm1.72$ & $8.45\pm1.23$ \\
    CLIC \citep{niu2024clic} & $43.5\pm0.49$  & $68.3\pm0.60$ & $45.1\pm0.71$ & $\mathbf{15.1\pm1.39}$ & $9.31\pm0.28$ \\
    CurricuVLM \citep{sheng2026curricuvlm} & $52.3\pm2.28$  & $76.2\pm2.28$ & $51.0\pm1.65$ & $16.0\pm0.80$ & $9.52\pm0.14$ \\
    DreamerV3 \citep{hafner2023dreamerv3} & $46.3\pm0.24$ & $71.4\pm0.57$ & $45.8\pm0.43$ & $19.4\pm4.63$ & $7.05\pm0.32$ \\
    \midrule
    SAFER (ours)                & $\mathbf{53.4\pm1.60}$  & $\mathbf{79.7\pm0.79}$ & $\mathbf{51.3\pm1.35}$ & $18.6\pm0.57$ & $\mathbf{10.1\pm0.06}$ \\
    \bottomrule
  \end{tabular}\\[8pt]
  (b)~Safety-critical test scenarios\\[2pt]
  \begin{tabular}{lccccc}
    \toprule
    Method & Episode reward $\uparrow$ & Road completion (\%) $\uparrow$ & Total distance $\uparrow$ & Crash rate (\%) $\downarrow$ & Average speed $\uparrow$ \\
    \midrule
    SAC \citep{haarnoja2018sac} & $38.4\pm1.97$  & $63.2\pm1.21$ & $40.9\pm1.34$ & $30.5\pm2.33$ & $9.25\pm0.07$ \\
    PPO \citep{schulman2017ppo} & $38.4\pm0.86$  & $62.7\pm1.05$ & $40.0\pm0.70$ & $32.0\pm2.02$ & $9.94\pm0.30$ \\
    TD3 \citep{fujimoto2018td3} & $42.4\pm1.01$  & $65.2\pm1.40$ & $42.6\pm1.26$ & $39.7\pm1.04$ & $8.02\pm0.77$ \\
    \midrule
    SAC-PID \citep{stooke2020pid} & $7.46\pm3.05$  & $19.0\pm4.19$ & $12.7\pm2.82$ & $30.4\pm3.27$ & $5.22\pm0.54$ \\
    TD3-Lag \citep{ray2019safetygym} & $29.1\pm0.00$  & $39.0\pm0.00$ & $30.7\pm0.00$ & $51.0\pm0.00$ & $5.70\pm0.00$ \\
    TD3-PID \citep{stooke2020pid} & $-0.68\pm1.00$ & $5.50\pm0.90$ & $4.58\pm0.35$ & $29.4\pm5.83$ & $4.18\pm1.06$ \\
    \midrule
    BC \citep{gleave2022imitation} & $27.4\pm2.69$  & $53.2\pm1.81$ & $31.0\pm2.17$ & $33.7\pm0.27$ & $7.30\pm0.26$ \\
    GAIL \citep{ho2016gail} & $29.8\pm1.07$  & $52.3\pm1.42$ & $32.7\pm1.05$ & $33.7\pm1.78$ & $7.77\pm0.73$ \\
    AIRL \citep{fu2018airl} & $26.3\pm1.84$  & $52.8\pm3.16$ & $31.2\pm1.50$ & $27.8\pm3.51$ & $9.74\pm0.72$ \\
    SQIL \citep{reddy2020sqil} & $18.9\pm0.56$  & $41.4\pm1.01$ & $23.5\pm0.52$ & $38.5\pm1.67$ & $6.54\pm0.34$ \\
    \midrule
    CAT \citep{zhang2023cat} & $42.5\pm3.95$  & $66.6\pm4.37$ & $43.4\pm3.48$ & $32.1\pm2.08$ & $8.36\pm1.17$ \\
    CLIC \citep{niu2024clic} & $39.3\pm0.72$  & $64.3\pm0.40$ & $41.6\pm0.78$ & $26.2\pm1.17$ & $9.21\pm0.26$ \\
    CurricuVLM \citep{sheng2026curricuvlm} & $48.9\pm1.53$  & $73.4\pm1.66$ & $\mathbf{48.4\pm1.31}$ & $\mathbf{25.1\pm1.17}$ & $9.45\pm0.16$ \\
    DreamerV3 \citep{hafner2023dreamerv3} & $39.4\pm0.53$ & $63.0\pm1.00$ & $40.7\pm0.52$ & $44.3\pm6.61$ & $7.47\pm0.31$ \\
    \midrule
    SAFER (ours)                & $\mathbf{49.0\pm0.54}$  & $\mathbf{74.5\pm0.38}$ & $48.2\pm0.45$ & $31.4\pm1.23$ & $\mathbf{10.0\pm0.05}$ \\
    \bottomrule
  \end{tabular}
\end{table}

\subsection{A reachability view of the world model's risk prediction}
\label{sec:exp:risk}

\begin{figure}[tbp]
  \centering
  \includegraphics[width=\linewidth]{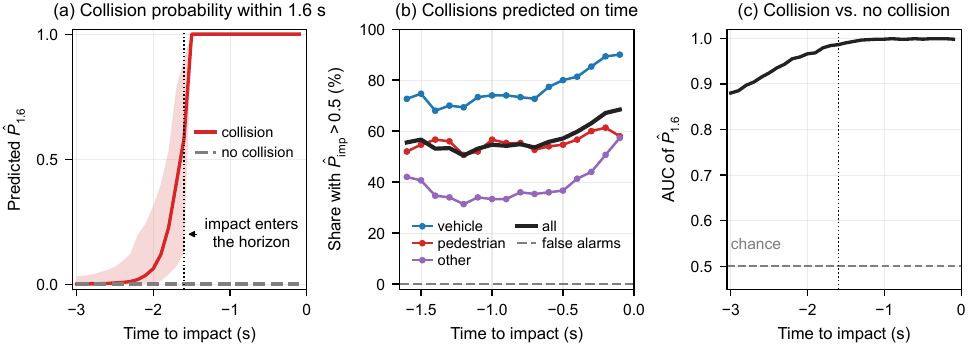}
  \caption{How early the risk head foresees recorded collisions ($150$ events per collision partner, $150$
  collision-free control segments; world model fed the recorded history and the driver's executed
  actions). (a)~Median and interquartile range of $\hat{P}_{1.6}$, the predicted probability of a
  collision within the $1.6$\,s horizon, against the time to impact, for the $450$ collision segments and the
  controls; the dotted line marks the moment the
  impact enters the horizon. (b)~Share of collisions with $\hat{P}_{\mathrm{imp}}>0.5$, i.e.\ predicted
  by the step at which they actually happen, per collision partner against the time to impact, and the false-alarm rate of the
  controls ($\hat{P}_{1.6}>0.5$ at any of their $4{,}500$ decisions). (c)~AUC of $\hat{P}_{1.6}$ between
  collision and control segments at each time to impact.}
  \label{fig:risk_timeline}
\end{figure}

\begin{figure}[tbp]
  \centering
  \includegraphics[width=\linewidth]{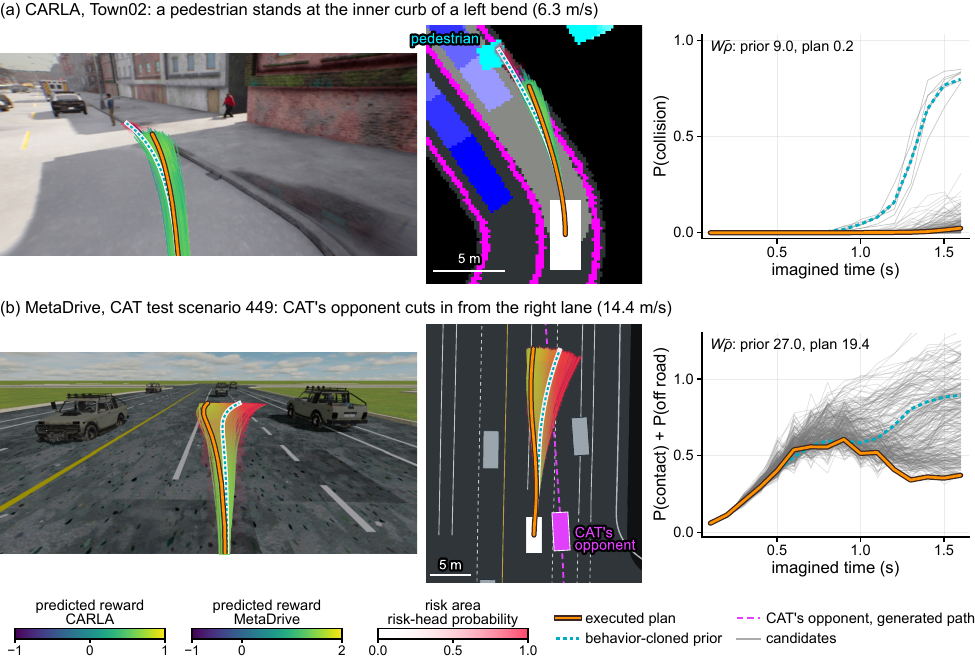}
  \caption{The imagined reachable set and its risk area in two closed-loop decisions of the full system. Left:
  a front camera added for visualization, which the agent does not observe. Middle: (a)~the BEV that the agent
  observes; (b)~a top-down view drawn from the simulator state, with CAT's opponent (magenta) and the trajectory
  that CAT generates for it (dashed), which the planner does not know. In both views the $270$ candidate chunks that
  the planner scored at that step are drawn as in \ref{app:examples}, as the paths that they trace under a
  kinematic bicycle model at the speeds that the world model predicts, colored by the predicted reward of each
  imagined step; together they sample the reachable set. Red: the risk area, i.e.\ the risk head's probability of a
  priced safety event (CARLA: collision; MetaDrive: vehicle contact plus leaving the road, object contact staying at
  zero), averaged at each place over the imagined states of all candidates that pass there and drawn with an
  opacity that grows with the probability; it is the same field in both views. Orange: executed plan; dotted:
  behavior-cloned prior. Right: the same probability along the horizon for every candidate (gray), the plan
  (orange) and the prior (dotted), with their risk terms $W\bar{\rho}$ of Eq.~\eqref{eq:risk_penalty}.}
  \label{fig:reachability}
\end{figure}

This subsection examines the world model's ability to predict risk from the perspective of reachability. Among
the states that the ego can reach within the horizon, which does the world model expect to be unsafe, and how
reliable is that expectation? At every step, the candidate chunks that the planner scores, each rolled out by the world model, sample the part
of the ego's reachable set that the planner explores within the horizon; we call it the imagined reachable set.
Along every candidate the world model predicts the reward of each imagined step, and the risk head the probability
that a priced safety event has occurred by then. Laid onto the ground, these probabilities mark a risk area, the
places of the reachable set where the risk head expects a collision or another priced
event. The planner does not threshold it. Eq.~\eqref{eq:risk_penalty} charges every candidate $W$ times the mean of
its probabilities over the horizon, so a candidate loses score in Eq.~\eqref{eq:score} the longer, and the deeper,
it runs through the risk area, while the predicted rewards enter its return. Read this way, the planner looks for a
path through the reachable set that earns a high predicted return and stays clear of the risk area. The risk area is
only as reliable as the risk head; we first test the head offline on recorded collisions
(Fig.~\ref{fig:risk_timeline}) and in closed loop on the system's own collisions, and then show the reachable set and its risk area in two closed-loop decisions
(Fig.~\ref{fig:reachability}).

The test replays recorded collisions and asks how early, and how confidently, the head predicts them. An
event is a collision that ends an episode of the corpus while the ego is still moving (at least $1$\,m/s
just before the impact) and whose episode contains the $3$\,s before the impact plus a full context. The
corpus contains $21{,}287$ such events. Each is assigned to a collision partner from the BEV at the impact
step: a pedestrian within $4$\,m, otherwise a vehicle within $6$\,m, otherwise \emph{other}, which covers
collisions with the static layout and contacts that these distances do not attribute ($12{,}961$ vehicle,
$2{,}896$ pedestrian and $5{,}430$ other events). We draw $150$ events per partner, so that the three are
weighted equally, and $150$ control segments from the same sets, in which the ego is moving and no episode
ends within the segment or the horizon after it. For each event the decision time slides from $3.0$\,s to
$0.1$\,s before the impact in steps of $0.1$\,s. At every decision the world model receives the recorded
$16$-step history and the $16$ actions the driver actually executed next, which are the actions that led to
the collision, and the head predicts the cumulative collision probabilities
$\operatorname{sigmoid}(\rho_{k,\mathrm{coll}})$ of Eq.~\eqref{eq:risk_logits} along the imagined
rollout. No planner is involved. Two readouts are reported: $\hat{P}_{1.6}$, the probability that a
collision has occurred by the end of the $1.6$\,s horizon, whose correct value is $0$ while the impact is
more than $1.6$\,s away and $1$ afterwards; and, once the impact lies inside the horizon,
$\hat{P}_{\mathrm{imp}}$, the probability that it has occurred by the step at which it actually happens,
which requires the collision to be timed as well as expected. Control segments go through the same
procedure, and their $\hat{P}_{1.6}$ measures false alarms. The events and controls come from the training
corpus, so this test measures what the head has learned to predict from the recorded situations; how it carries over
to new situations is shown by the closed-loop results of Sections~\ref{sec:exp:ablation}--\ref{sec:exp:metadrive}.

The controls never raise a false alarm. Over all $4{,}500$ control decisions the largest $\hat{P}_{1.6}$ is
$0.29$ and the $99$th percentile $0.04$ (Fig.~\ref{fig:risk_timeline}a). On the collision segments the
median $\hat{P}_{1.6}$ stays near zero until about $2.2$\,s before the impact, rises steeply as the impact
approaches the horizon and is $1$ once the impact lies inside it; from $1.0$\,s before the impact on,
$95$--$97\%$ of the events have $\hat{P}_{1.6}>0.5$ ($97$--$100\%$ of vehicle, $95$--$97\%$ of pedestrian and
$91$--$95\%$ of other collisions). The separation from the controls is already $0.88$ in AUC at $3.0$\,s,
when the impact is still $1.4$\,s beyond the horizon, $0.97$ at $2.0$\,s, and at least $0.998$ from $1.1$\,s
on (Fig.~\ref{fig:risk_timeline}c).

Requiring the collision to be predicted by the step at which it happens is stricter. $\hat{P}_{\mathrm{imp}}$
exceeds $0.5$ for $56\%$ of the events when the impact is $1.6$\,s away, $54$--$57\%$ between $1.0$ and
$0.5$\,s, and $68\%$ at $0.1$\,s (Fig.~\ref{fig:risk_timeline}b). Vehicle collisions are timed best
($73\%$ at $1.6$\,s, $90\%$ at $0.1$\,s), pedestrian collisions at $51$--$61\%$ and other collisions at
$31$--$57\%$. Since at least $92\%$ of these events have $\hat{P}_{1.6}>0.5$ from $1.4$\,s before the impact on, the gap
between the two readouts is a timing error, not a missed collision. The head expects the collision inside
the horizon but places it later than it happens.

In closed loop, the same test runs on the system's own collisions. In the $228$ evaluation episodes of the
final system, the SAFER of Table~\ref{tab:comparison}, the ego collides $68$ times while moving ($47$ times with a
vehicle, $20$ times with a pedestrian, once with the static layout); these events are compared with $300$
collision-free control segments of the same episodes, and the executed plan takes the place of the driver's
actions. The head of the final system still separates the two, less sharply than on recorded collisions and again
without a false alarm. No control decision reaches $\hat{P}_{1.6}=0.04$, the AUC of $\hat{P}_{1.6}$ is $0.89$ at
$2.0$\,s before the impact and $0.95$ at $0.1$\,s, and it is at least $0.95$ at every time to impact for the
pedestrian collisions. The probabilities that the head assigns to its own plan before these collisions are much
lower than those of the offline head on recorded collisions, though. Their median stays below $0.02$ up to the
impact, and within the last $1.6$\,s only $4\%$ of the vehicle and $5\%$ of the pedestrian collisions reach
$\hat{P}_{1.6}>0.5$ at any decision. Where the head does flag the behavior-cloned prior's chunk
($\hat{P}_{1.6}>0.5$, at $315$ moments), the planner executes a plan that the head expects to be safe in $98\%$ of
the cases, and $92\%$ of these moments pass without a collision in the next $3$\,s.

Fig.~\ref{fig:reachability} shows the reachable set and the risk area in two decisions of the full system, the
planner with the risk penalty on the model adapted online, while it drives in closed loop. (a)~In CARLA, at
$6.3$\,m/s, a pedestrian stands at the inner curb of a left bend, about $13$\,m ahead. The risk area is confined to
the inner edge of the reachable set next to the pedestrian, where the predicted collision probability reaches
$0.83$; most candidates do not come near it (median risk term $W\bar{\rho}=0.15$). The prior cuts the bend towards
the pedestrian and runs into the risk area (risk term $9.0$, higher than for $96\%$ of the candidates); the plan
follows the lane outside it ($0.2$). Of the plan's score lead of $19.8$ over the prior, $8.8$ come from the risk
term and $10.9$ from the predicted return, since the world model also predicts low rewards for the steps at the
curb. (b)~On CAT's benchmark, at $14.4$\,m/s, CAT's opponent drives alongside in the right lane at $8.5$\,m/s on a
trajectory that cuts into the ego's lane. With the opponent next to the ego, every candidate carries some risk
(risk terms between $19.7$ and $39.2$), and the risk area is densest on the right, where the candidates would cut in
front of the opponent. The plan, which MPPI refits from the best-scored candidates and is not itself one of them, keeps its lane at the left edge of the reachable set, where the risk is lowest (risk
term $19.4$, against $27.0$ for the prior). Plan and prior are predicted nearly the same reward ($1.50$ and $1.48$
per step on average, the progress along the route), and $7.6$ of the plan's score lead of $10.0$ come from the risk
term. In both decisions the plan runs through the part of the reachable set where the risk area is weakest,
whereas the prior, which the planner keeps among its candidates, runs into it.

\subsection{Explaining the imagination}
\label{sec:exp:coc}

The planner chooses among imagined futures, so the imagination has to carry the information that driving
decisions depend on. We check this with a scene-factor head, a two-layer MLP that reads a latent as the state of
the traffic light on the route (none, red light or stop sign, yellow, green) and the presence of a vehicle ahead,
a vehicle directly ahead, a pedestrian near the path, a stop line on the route and the ego on its route, with
targets from rule-based labels of the BEV frames. It is trained with the world model in stage~1, on the context and the imagined latents
(Section~\ref{sec:exp:setup}), and the planner never uses it, so this subsection reads the imagination of the stage-1 world model. For recorded windows
of the corpus, the model receives the $16$ recorded steps of history and the $16$ actions the driver executed
next, and the head reads each imagined latent.

\begin{figure}[tbp]
  \centering
  \includegraphics[width=\linewidth]{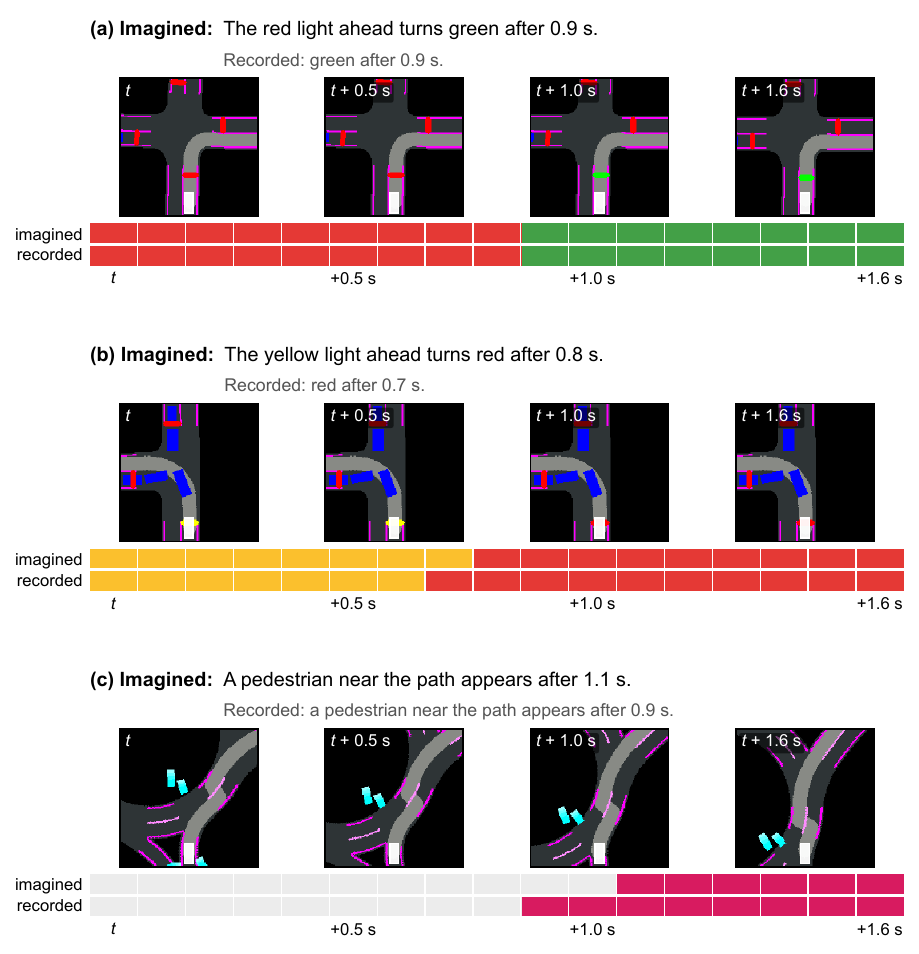}
  \caption{Explaining the imagination with the scene-factor head, on recorded windows of the corpus; the model
  receives the recorded history and the actions the driver executed. Each row: the sentence composed from the
  factors read in the imagined latents and the recorded outcome; the BEV at the decision time $t$ and the
  recorded frames $0.5$, $1.0$ and $1.6$\,s later (ego white); and the imagined and recorded state over the
  $1.6$\,s horizon in steps of $0.1$\,s: (a), (b)~the traffic light, (c)~a pedestrian near the path (magenta).}
  \label{fig:scene_imagination}
\end{figure}

Table~\ref{tab:scene} compares the state read $1.6$\,s ahead in the imagination with the label of the recorded
frame, and with copying the state read at the decision time. Over random windows the factors rarely change
within the horizon, and the imagination is as accurate as copying. It does not invent changes. Where the state
does change, copying is almost always wrong for traffic lights, while the imagined light is right in $77\%$ of
the windows; for pedestrians and the vehicle ahead the imagination also does better than copying ($57\%$ against
$51\%$ and $49\%$ against $30\%$). Fig.~\ref{fig:scene_imagination} turns the imagined factors of three
recorded situations into a sentence. The model foresees a red light turning green and a yellow light turning
red within $0.1$\,s of the recorded change, and pedestrians beside the road coming near the ego's path
($1.1$\,s imagined, $0.9$\,s recorded). The imagination thus carries the state of traffic lights and the
movement of other road users on which the planner's decisions depend.

\begin{table}[tbp]
  \centering
  \caption{Reading the imagination of the stage-1 world model with its scene-factor head: share of recorded
  windows (\%) in which the state read $1.6$\,s ahead in the imagined latents matches the rule label of the
  recorded frame, against copying the state the head reads at the decision time. All windows: $400$ drawn at
  random; changing windows: $150$ per factor in which the state $1.6$\,s ahead differs from the present one (the
  light changes color, a pedestrian comes near the path, a vehicle ahead appears or leaves).}
  \label{tab:scene}
  \footnotesize
  \begin{tabular}{lcccc}
    \toprule
    & \multicolumn{2}{c}{All windows} & \multicolumn{2}{c}{Changing windows} \\
    \cmidrule(lr){2-3}\cmidrule(lr){4-5}
    Factor & Imagined & Copied & Imagined & Copied \\
    \midrule
    Traffic light            & $93.5$ & $90.8$ & $76.7$ & $2.7$ \\
    Pedestrian near the path & $98.0$ & $98.3$ & $56.7$ & $51.3$ \\
    Vehicle ahead            & $94.8$ & $95.0$ & $49.3$ & $30.0$ \\
    \bottomrule
  \end{tabular}
\end{table}

\subsection{Computational efficiency}
\label{sec:exp:efficiency}

SAFER plans with a small model of $15.5$\,M parameters in CARLA ($15.3$\,M for the world model with its heads,
$0.2$\,M for the risk head) and $14.3$\,M in MetaDrive. Its training is also inexpensive on at most two Quadro
RTX 6000 GPUs. The offline training takes $66$\,h in CARLA and $7$\,h in MetaDrive, and the online adaptation
over $10^{6}$ environment steps $41$\,h and $8$\,h.
Table~\ref{tab:efficiency} gives the inference time per driving decision in CARLA, from the observation to the
action, on one of these GPUs with a batch of one, and the decision frequency it allows. The behavior-cloned prior
decodes one action chunk at $173$\,Hz. The planner rolls out and scores $270$ candidate chunks of $16$ steps
per decision, two iterations of $135$, and is therefore about four times slower, yet it still decides at
$43$--$45$\,Hz, more than four times the $10$\,Hz at which the agent acts; even the slowest of the $2{,}000$
decisions ($42$\,ms) would allow $24$\,Hz. The risk term and the online adaptation do not change this cost. A
small world model is thus enough. With $15.5$\,M parameters, SAFER reaches the highest route completion of
Table~\ref{tab:comparison} and plans in real time on a single GPU.

\begin{table}[tbp]
  \centering
  \caption{Inference time per driving decision (observation encoding and controller) on one Quadro RTX 6000 with
  a batch of one: mean $\pm$ standard deviation over $2{,}000$ decisions replayed from recorded closed-loop
  episodes, and the decision frequency it allows. All controllers use the same model of $15.5$\,M parameters.
  Rollouts: imagined $16$-step action chunks scored per decision. The agent acts at $10$\,Hz.}
  \label{tab:efficiency}
  \footnotesize
  \begin{tabular}{lccc}
    \toprule
    Controller & Rollouts & Time per decision (ms) & Frequency (Hz) \\
    \midrule
    BC prior                 & $0$   & $5.8\pm0.4$  & $173$ \\
    MPPI                     & $270$ & $23.1\pm3.2$ & $43$ \\
    + risk term              & $270$ & $22.3\pm2.3$ & $45$ \\
    + online, $10^{6}$ steps & $270$ & $22.0\pm2.3$ & $45$ \\
    \bottomrule
  \end{tabular}
\end{table}

\section{Conclusion}
\label{sec:conclusion}

A world model lets an autonomous vehicle weigh the consequences of its actions before it acts, but it is reliable
only within the coverage of its training data. A planner that exploits the model within this coverage can hardly
improve on the data, a planner that explores beyond it is misled by the errors of the model, and the experience
that would widen the coverage is the experience of failure. SAFER navigates this exploration--exploitation dilemma
by learning from failures that others make, where they can be afforded, judged by an expert. Partially trained
agents and scripted, recoverable perturbations fail in simulation, an expert annotates every visited state, and a
chunked latent world model learns from this corpus to foresee consequences as reward factors, expert values and
collision probabilities. A risk-aware MPPI planner scores candidate maneuvers by their predicted return minus a
bounded collision penalty, and an online loop refines the dynamics, the reward scorers and the risk head on the
planner's own driving while the planner scores against slowly moving copies of them.

In CARLA, the risk head foresees recorded collisions before they happen without raising a false alarm on
collision-free driving, and in closed loop each stage contributes where it is meant to. On $76$ routes, the risk
term raises the driving score of the planner by $0.08$, mainly through fewer collisions with pedestrians, and
online adaptation by a further $0.07$, with fewer collisions of every type, a gain that adapting the dynamics alone
does not reach; together with planning they raise
the driving score of the behavior-cloned prior from $0.65$ to $0.77$, and one turn of the data flywheel, which
adds targeted pedestrian data, makes the system safer at a driving score of $0.80$, with the highest route completion and reward per step in the comparison
with the baselines. In MetaDrive, SAFER reaches the highest episode reward and road
completion in the comparison with the baselines, in regular and in safety-critical
traffic. The imagination of its world model carries the
information on which driving decisions depend, such as a traffic light that is about to change, and with
$15.5$\,M parameters the planner decides at $43$--$45$\,Hz on a single GPU.

The recipe needs only a driver that is allowed to fail recoverably and an expert that can be queried at the
states it visits. A proving ground with a safety driver offers both, which makes the transfer of SAFER from
privileged bird's-eye-view observations in simulation to sensor-based driving its natural next step.

\section*{Acknowledgments}
{\emergencystretch=2em  %
This work was supported by Purdue University's Center for Connected and Automated Transportation (CCAT) which is
a part of the USDOT Region 5 University Transportation Center funded by the USDOT Award \#69A3551747105. Also,
general support was provided by Purdue's Next-generation Transportation Systems (NEXTRANS) Center, and the following
initiatives at Purdue University's College of Engineering: the Institute for Control, Optimization, \& Networks
(ICON), and the Autonomous \& Connected Systems (ACS) initiative. The contents of this paper reflect the views of the
authors, who are responsible for the facts and the accuracy of the data presented herein, and do not necessarily
reflect the official views or policies of the sponsoring organization.\par}

\appendix
\renewcommand{\theHfigure}{\Alph{section}.\arabic{figure}}
\renewcommand{\theHtable}{\Alph{section}.\arabic{table}}

\section{Proofs of the formal statements}
\label{app:proofs}

Norms are Euclidean on $\R^{\horizon\times2}$ for action chunks and on the parameter space for $\theta$. The
statements of Section~\ref{sec:method:planner} concern a single decision state and its candidate sets; those
of Section~\ref{sec:method:online} concern the sequence of bursts $m=0,1,\dots$ of the online loop, with
$\theta_m$ the live parameters and $\bar{\theta}_m$ the EMA target after burst $m$.

\subsection{Proof of Proposition~\ref{prop:curse}}
Because $\hat{\act}$ maximizes $J$ over $\mathcal{A}$, $J(\hat{\act})\ge J(\act^{\star})$ in every
realization (ties are broken by any fixed rule). Since $\mathcal{A}$ is fixed and $J^{\star}$ is
deterministic, $\act^{\star}$ is deterministic, and unbiasedness at this fixed candidate gives
$\E[J(\act^{\star})]=J^{\star}(\act^{\star})+\E[e_J(\act^{\star})]=J^{\star}(\act^{\star})$. Therefore
\begin{align*}
  \E[e_J(\hat{\act})]&=\E[J(\hat{\act})]-\E[J^{\star}(\hat{\act})]
  \;\ge\;\E[J(\act^{\star})]-\E[J^{\star}(\hat{\act})]\\
  &=\E\big[J^{\star}(\act^{\star})-J^{\star}(\hat{\act})\big]\;\ge\;0,
\end{align*}
where the last inequality holds because $\act^{\star}$ maximizes $J^{\star}$ over $\mathcal{A}$.
\hfill$\square$

\subsection{Proof of Proposition~\ref{prop:regret}}
The regret is non-negative because $\act^{\star}$ maximizes $J^{\star}$. Writing $J^{\star}=J-e_J$,
\begin{align*}
  J^{\star}(\act^{\star})-J^{\star}(\hat{\act})
  &=\big[J(\act^{\star})-J(\hat{\act})\big]+e_J(\hat{\act})-e_J(\act^{\star})\\
  &\le e_J(\hat{\act})-e_J(\act^{\star})\;\le\;|e_J(\hat{\act})|+|e_J(\act^{\star})|,
\end{align*}
since $J(\hat{\act})\ge J(\act^{\star})$ and by the triangle inequality. The argument uses only that
$\hat{\act}$ maximizes $J$ and $\act^{\star}$ maximizes $J^{\star}$ over the same set.
\hfill$\square$

\subsection{Proof of Lemma~\ref{lem:incumbent}}
The fixed candidates are formed once per control step and enter the candidate set of every iteration
(Eq.~\eqref{eq:candidates}), so $\mathcal{F}_t\subseteq\mathcal{A}_i$ for every $i$. Within a control step the
context is the same and the modules are evaluated deterministically, so every fixed candidate receives the same
score in every iteration. Hence $J^{\mathrm{best}}_i=\max_{\act\in\mathcal{A}_i}J(\act)\ge\max_{\act\in\mathcal{F}_t}J(\act)$,
and the second inequality holds because $\act^{\mathrm{bc}}\in\mathcal{F}_t$.
\hfill$\square$

\subsection{Proof of Theorem~\ref{thm:consistency}}
As usual for almost-sure statements in the number of samples, the $N_{\mathrm{bc}}$ proposals are the first
$N_{\mathrm{bc}}$ elements of one sequence of independent draws. Write $\Pi(\bm{x})=\clip(\bm{x},-\bar{a},\bar{a})$
for the entrywise clipping, fix $\epsilon>0$ and let
$U_\epsilon=\{\act\in\mathcal{C}:\|\act-\hat{\act}_{\mathcal{C}}\|<\epsilon\}$, which is convex as the intersection of
an open ball with the box $\mathcal{C}$.

\emph{Step 1: a superlevel set inside $U_\epsilon$.} By Assumption~\ref{ass:maximiser},
$\eta:=J(\hat{\act}_{\mathcal{C}})-\sup\{J(\act):\act\in\mathcal{C}\setminus U_\epsilon\}>0$. The set
$V=\{\act\in\mathcal{C}:J(\act)>J(\hat{\act}_{\mathcal{C}})-\eta\}$ satisfies $V\subseteq U_\epsilon$, because every
$\act\in\mathcal{C}\setminus U_\epsilon$ has $J(\act)\le J(\hat{\act}_{\mathcal{C}})-\eta$.

\emph{Step 2: a prior proposal lands in $V$ with positive probability.} By continuity of $J$ at
$\hat{\act}_{\mathcal{C}}$ there is $r>0$ such that every $\act\in\mathcal{C}$ with $\|\act-\hat{\act}_{\mathcal{C}}\|<r$
lies in $V$. The clipping $\Pi$ is non-expansive in every coordinate, $|\Pi(x)_j-\Pi(y)_j|\le|x_j-y_j|$, and fixes
$\hat{\act}_{\mathcal{C}}\in\mathcal{C}$; hence $\|\Pi(\bm{x})-\hat{\act}_{\mathcal{C}}\|\le\|\bm{x}-\hat{\act}_{\mathcal{C}}\|$,
and $\Pi$ maps the open ball $B(\hat{\act}_{\mathcal{C}},r)$ into $V$. A prior proposal is
$\Pi(\act^{\mathrm{bc}}+\bm{\sigma}_{\mathrm{bc}}\odot\bm{\epsilon}')$, where
$\act^{\mathrm{bc}}+\bm{\sigma}_{\mathrm{bc}}\odot\bm{\epsilon}'$ is Gaussian with the positive diagonal covariance
$\operatorname{diag}(\bm{\sigma}_{\mathrm{bc}}^{2})$, so the proposal lies in $V$ with probability at least
$q:=\Pr\big(\act^{\mathrm{bc}}+\bm{\sigma}_{\mathrm{bc}}\odot\bm{\epsilon}'\in B(\hat{\act}_{\mathcal{C}},r)\big)>0$, a
non-degenerate Gaussian assigning positive probability to every open ball.

\emph{Step 3: all elites lie in $U_\epsilon$.} The proposals are independent and identically distributed, so by
the strong law of large numbers the fraction of the $N_{\mathrm{bc}}$ proposals that lie in $V$ converges almost
surely to $\Pr(\text{proposal}\in V)\ge q>0$. Hence, almost surely, at least $N_e$ candidates lie in $V$ for all
$N_{\mathrm{bc}}$ large enough. On this event each of the $N_e$ best candidates scores at least as high as the
$N_e$-th best candidate in $V$, hence strictly above $J(\hat{\act}_{\mathcal{C}})-\eta$. All candidates lie in
$\mathcal{C}$, and every candidate in $\mathcal{C}\setminus U_\epsilon$ scores at most
$J(\hat{\act}_{\mathcal{C}})-\eta$; therefore the $N_e$ best candidates, and with them the best candidate, lie in
$V\subseteq U_\epsilon$.

\emph{Step 4: the mean lies in $U_\epsilon$.} For every $\kappa\ge0$ the weights of Eq.~\eqref{eq:mppi_update}
are non-negative and sum to one, so $\bm{\mu}$ is a convex combination of the elites and lies in the convex set
$U_\epsilon$; in particular $\|\bm{\mu}-\hat{\act}_{\mathcal{C}}\|<\epsilon$, and no clipping of $\bm{\mu}$ is needed.

\emph{Step 5.} Applying Steps 1--4 with $\epsilon=1,\tfrac12,\tfrac13,\dots$ and intersecting the countably many
almost-sure events shows that $\bm{\mu}\to\hat{\act}_{\mathcal{C}}$ and that the best candidate converges to
$\hat{\act}_{\mathcal{C}}$, almost surely, as $N_{\mathrm{bc}}\to\infty$.
\hfill$\square$

\subsection{Proof of Corollary~\ref{cor:regret}}
By Theorem~\ref{thm:consistency} the planned chunk $\bm{\mu}$ converges to $\hat{\act}_{\mathcal{C}}$, the maximizer
of $J$ over $\mathcal{C}$, and $\act^{\star}_{\mathcal{C}}$ maximizes $J^{\star}$ over $\mathcal{C}$. The proof of
Proposition~\ref{prop:regret} uses only these two maximality properties, so it applies with the candidate set
$\mathcal{A}=\mathcal{C}$, $\hat{\act}=\hat{\act}_{\mathcal{C}}$ and $\act^{\star}=\act^{\star}_{\mathcal{C}}$.
\hfill$\square$

\subsection{Proof of Lemma~\ref{lem:ema_rate}}
Unrolling Eq.~\eqref{eq:ema} over the first $m$ bursts gives
\begin{equation*}
  \bar{\theta}_{m}=(1-\tau_b)^{m}\,\bar{\theta}_0+\sum_{j=1}^{m}\tau_b(1-\tau_b)^{m-j}\,\theta_{j},
\end{equation*}
a convex combination, since the weights are non-negative and
$(1-\tau_b)^{m}+\tau_b\sum_{i=0}^{m-1}(1-\tau_b)^{i}=1$. With $\tau_b=1-(1-\tau)^{n_b/N_c}$,
$(1-\tau_b)^{m}=(1-\tau)^{m n_b/N_c}=(1-\tau)^{s/N_c}$ with $s=m n_b$, which equals $1-\tau$ when $s=N_c$.
\hfill$\square$

\subsection{Proof of Proposition~\ref{prop:tracking}}
Let $\bm{d}_m=\bar{\theta}_m-\theta_m$. By Eq.~\eqref{eq:ema},
$\bm{d}_{m+1}=(1-\tau_b)(\bar{\theta}_m-\theta_{m+1})=(1-\tau_b)\bm{d}_m+(1-\tau_b)(\theta_m-\theta_{m+1})$, so
$\|\bm{d}_{m+1}\|\le(1-\tau_b)\|\bm{d}_m\|+(1-\tau_b)\Delta_\theta$. By induction,
\begin{align*}
  \|\bm{d}_m\|&\le(1-\tau_b)^{m}\|\bm{d}_0\|+(1-\tau_b)\,\Delta_\theta\sum_{i=0}^{m-1}(1-\tau_b)^{i}\\
  &=(1-\tau_b)^{m}\|\bm{d}_0\|+\frac{1-\tau_b}{\tau_b}\big(1-(1-\tau_b)^{m}\big)\,\Delta_\theta .
\end{align*}
If $\theta_m\to\theta_\infty$, then $\Delta_m:=\|\theta_{m+1}-\theta_m\|\to0$. Given $\varepsilon>0$, choose
$m_0$ with $\Delta_m\le\varepsilon$ for all $m\ge m_0$; the same recursion started at $m_0$ gives
$\|\bm{d}_m\|\le(1-\tau_b)^{m-m_0}\|\bm{d}_{m_0}\|+(1-\tau_b)\varepsilon/\tau_b$, so
$\limsup_m\|\bm{d}_m\|\le(1-\tau_b)\varepsilon/\tau_b$ for every $\varepsilon>0$ and $\bm{d}_m\to\bm{0}$.
Hence $\|\bar{\theta}_m-\theta_\infty\|\le\|\bm{d}_m\|+\|\theta_m-\theta_\infty\|\to0$.
\hfill$\square$

\subsection{Proof of Proposition~\ref{prop:attenuation}}
By the unrolled form in the proof of Lemma~\ref{lem:ema_rate},
\begin{equation*}
  \bar{\theta}_m=(1-\tau_b)^{m}\bar{\theta}_0+\sum_{j=1}^{m}\tau_b(1-\tau_b)^{m-j}(\theta_\infty+\bm{\zeta}_j).
\end{equation*}
Taking expectations,
$\E[\bar{\theta}_m]=(1-\tau_b)^{m}\bar{\theta}_0+\big(1-(1-\tau_b)^{m}\big)\theta_\infty\to\theta_\infty$. The
initial target is deterministic and the $\bm{\zeta}_j$ are independent with covariance $\Sigma$, so
\begin{align*}
  \operatorname{Cov}(\bar{\theta}_m)&=\sum_{j=1}^{m}\tau_b^{2}(1-\tau_b)^{2(m-j)}\,\Sigma
  =\tau_b^{2}\,\frac{1-(1-\tau_b)^{2m}}{1-(1-\tau_b)^{2}}\,\Sigma\\
  &\to\frac{\tau_b^{2}}{\tau_b(2-\tau_b)}\,\Sigma=\frac{\tau_b}{2-\tau_b}\,\Sigma ,
\end{align*}
using $0<\tau_b\le1$.
\hfill$\square$

\section{Closed-loop planning examples}
\label{app:examples}
\setcounter{figure}{0}

\begin{figure}[p]
  \centering
  \includegraphics[width=\linewidth]{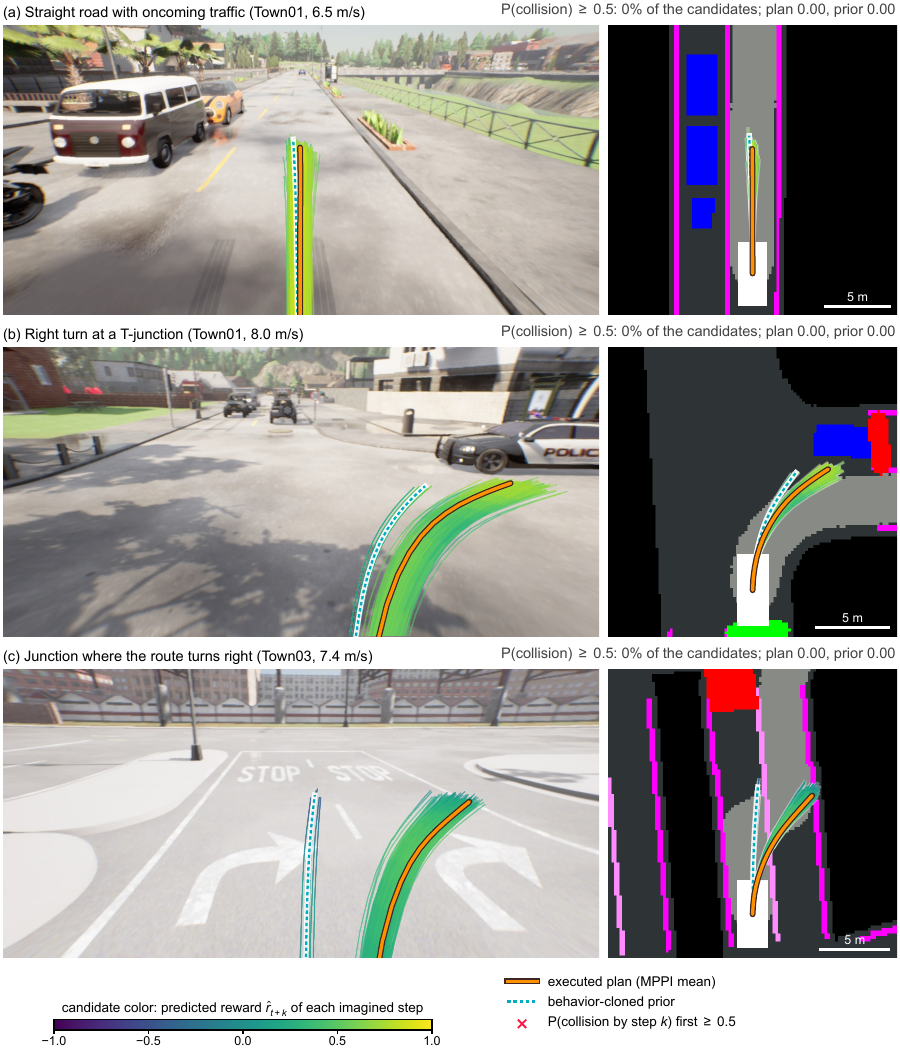}
  \caption{Planning steps of the full system in normal driving (\ref{app:examples}). Left: a front camera
  added for visualization (not observed by the agent); right: the BEV that the planner receives, cropped around the
  candidates. Paths: the $270$ candidate chunks scored at that step, sketched with a kinematic bicycle model at the
  predicted speeds and colored by the predicted reward of each imagined step; orange: executed plan; dotted:
  behavior-cloned prior; crosses: the first imagined step at which the predicted collision probability reaches
  $0.5$. (a)~Straight road with oncoming traffic; (b)~right turn at a T-junction; (c)~junction where the route turns
  right while a few chunks, the prior among them, continue straight towards a red light.}
  \label{fig:closed_loop_normal}
\end{figure}

\begin{figure}[p]
  \centering
  \includegraphics[width=\linewidth]{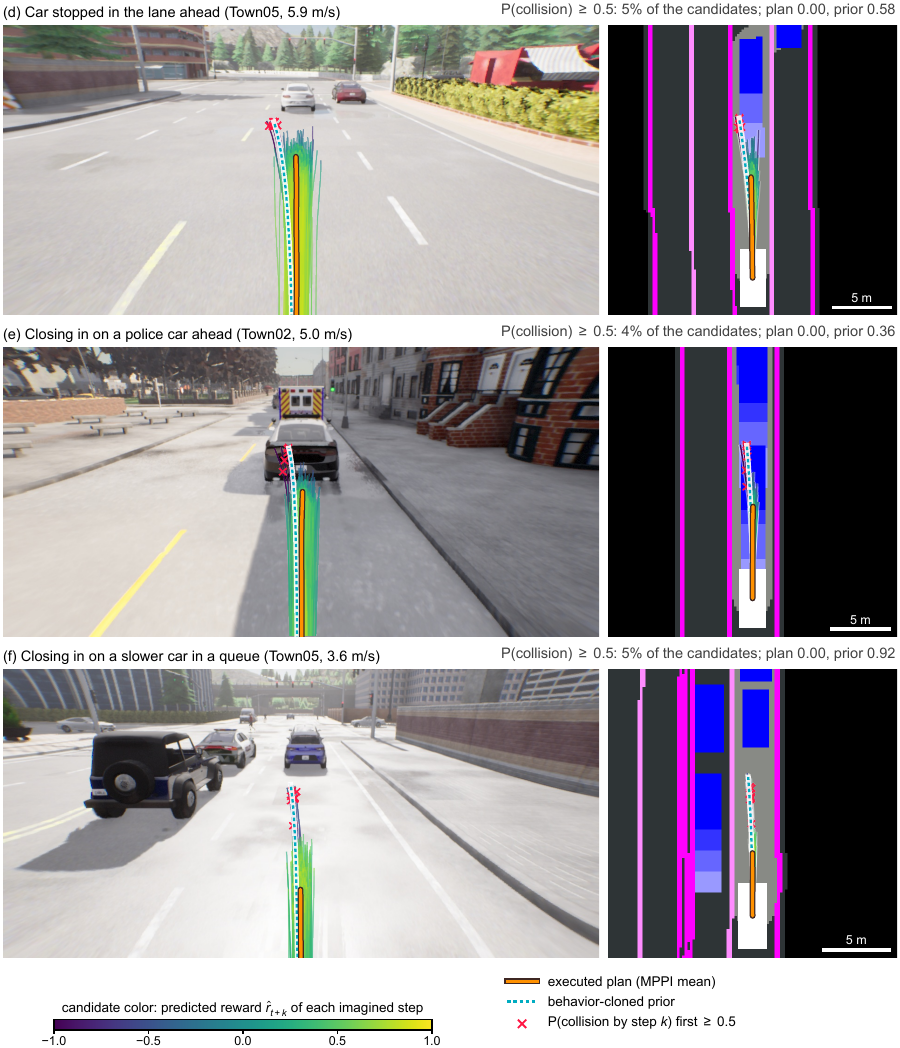}
  \caption{Planning steps of the full system in safety-critical scenes, drawn as in
  Fig.~\ref{fig:closed_loop_normal}. (d)~A car stopped in the lane ahead; (e)~a police car ahead;
  (f)~a slower car ahead in a queue. In each scene the chunks that the risk head predicts to collide (in (d) and (f), the behavior-cloned prior among them) carry the crosses, and the executed plan stays clear of them.}
  \label{fig:closed_loop_critical}
\end{figure}

Figs.~\ref{fig:closed_loop_normal} and~\ref{fig:closed_loop_critical} show single planning steps of the full
system, the planner with the risk penalty ($W=50$) on the model adapted online for $10^{6}$ steps, while it drives
Leaderboard routes of Section~\ref{sec:exp:setup:carla} in closed loop (seed $2021$). Each row pairs the BEV
that the planner receives with the image of a front camera that was added for these figures only; the agent never
observes it. Both views show the $270$ candidate chunks that the planner scored at that step in its two iterations
(Section~\ref{sec:method:planner}). As in Fig.~\ref{fig:planner}, a candidate is drawn as the path that its action
chunk traces under a kinematic bicycle model at the speeds predicted by the world model, not as decoded positions;
in the camera image the paths are projected onto the road, taken as flat. Every segment is colored by the blended
reward $\hat{r}_{t+k}$ of Eq.~\eqref{eq:score} that the world model predicts for that imagined step, a red cross
marks the step at which the risk head's probability that a collision has occurred first reaches $0.5$, the executed
plan, the MPPI mean whose first action is applied, is drawn in orange and the behavior-cloned prior dotted. The row
titles give the share of the candidates whose predicted collision probability reaches $0.5$ within the $1.6$\,s
horizon, and the probabilities of the plan and of the prior.

\paragraph{Normal driving}
In the scenes of Fig.~\ref{fig:closed_loop_normal} no candidate approaches a collision, and the candidates differ
only in their predicted rewards. On a straight road with oncoming traffic (a) they end within $0.5$\,m of the lane
center and are predicted high rewards (mean $0.78$, of a maximum of $1$). Entering a right turn at $8.0$\,m/s
(b), the plan brakes first (acceleration command $-0.87$) and slows to about $6$\,m/s through the turn, while the
prior turns less sharply and is predicted a lower reward ($0.35$ against $0.57$ on average over the horizon). At a
junction where the ego's lane turns right (c), a few chunks, the prior among them, continue straight towards a red
light $15$--$19$\,m ahead. Their predicted rewards are lower ($-0.08$ on average, against $0.26$ for the chunks
that turn), and the plan turns with the route ($J=49.9$ against $36.8$ for the prior).

\paragraph{Safety-critical scenes}
In each scene of Fig.~\ref{fig:closed_loop_critical}, $4$--$5\%$ of the candidates are predicted to collide within
the horizon, between $1.2$ and $1.6$\,s ahead, and in (d) and (f) the behavior-cloned prior is one of them; in each the prior
also accelerates harder than the plan (mean command over the first $0.8$\,s $0.99$, $0.97$ and $1.00$, against
$0.50$, $0.58$ and $0.32$). 
(d)~A car has stopped in the ego's lane, $16$\,m ahead. At $5.9$\,m/s, the chunks that keep accelerating
(mean command $0.96$ against $0.48$, and a predicted $8.3$\,m/s at the end of the horizon) reach its rear; their
predicted reward turns negative $0.8$--$0.9$\,s ahead, before their collision probability crosses $0.5$ at
$1.5$--$1.6$\,s. The prior is one of them ($0.58$, $J=42.0$ against $58.6$); the plan keeps its speed for
$1.2$\,s and then brakes (below $0.001$). (e)~Behind a police car about $8$\,m
ahead, at $5.0$\,m/s, the chunks that accelerate hardest (mean command $0.95$ against $0.59$, and a predicted
$12$\,m/s at the end of the horizon) reach its rear, with a mean predicted reward of $-0.33$ against $0.52$. The
prior is one of them ($0.36$, $J=18.5$ against $55.7$); the plan keeps the speed and the gap (below $0.001$).
(f)~In a queue, at $3.6$\,m/s, the prior keeps full throttle towards the slower car ahead and is predicted to
collide with probability $0.92$ ($J=41.6$); the plan accelerates less ($J=56.2$, below $0.001$), and within the
next $1.5$\,s the ego slows to $0.5$\,m/s behind the car. The colliding chunks are scored down twice, once by the risk
penalty and once by their predicted returns, which are also lower (mean $R$ of $42.9$, $7.6$ and $49.1$ against
$56.2$, $53.0$ and $55.0$ for the other candidates). Of the score gap between the plan and the prior, the risk term
accounts for $3.5$ of $16.6$ in (d), $2.4$ of $37.3$ in (e) and $7.8$ of $14.5$ in (f). In all three scenes the plan
that MPPI refits from the best-scored candidates stays clear of the colliding chunks, although in (d) and (f) the prior, which the
planner keeps among its candidates, is one of them.

\begin{figure}[p]
  \centering
  \includegraphics[width=\linewidth]{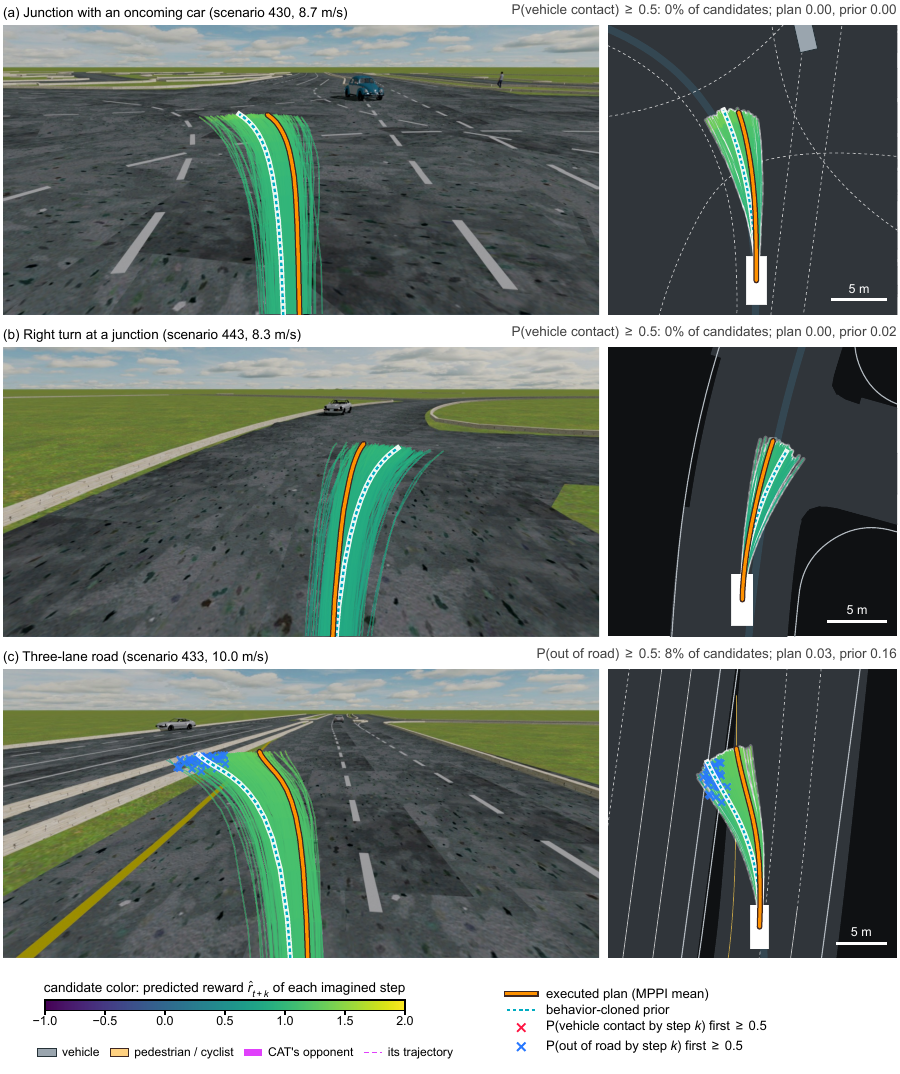}
  \caption{Planning steps of the full system on CAT's MetaDrive benchmark in regular traffic
  (\ref{app:examples}). Left: a front camera placed as in Fig.~\ref{fig:closed_loop_normal} and rendered for
  visualization (the agent observes a $101$-dimensional vector); right: a top-down view drawn from the simulator's
  state, cropped around the candidates. Paths: the $270$ candidate chunks, colored by the predicted reward of each
  imagined step (CAT's reward: progress in meters per step); orange: executed plan; dotted: behavior-cloned prior;
  red crosses: the first imagined step at which the predicted probability of a vehicle contact reaches $0.5$;
  blue crosses: the same for leaving the road. (a)~Junction with an oncoming car; (b)~right turn at a junction at $8.3$\,m/s; (c)~three-lane road on which $8\%$ of the candidates are predicted to leave the road on the
  left.}
  \label{fig:md_closed_loop_normal}
\end{figure}

\begin{figure}[p]
  \centering
  \includegraphics[width=\linewidth]{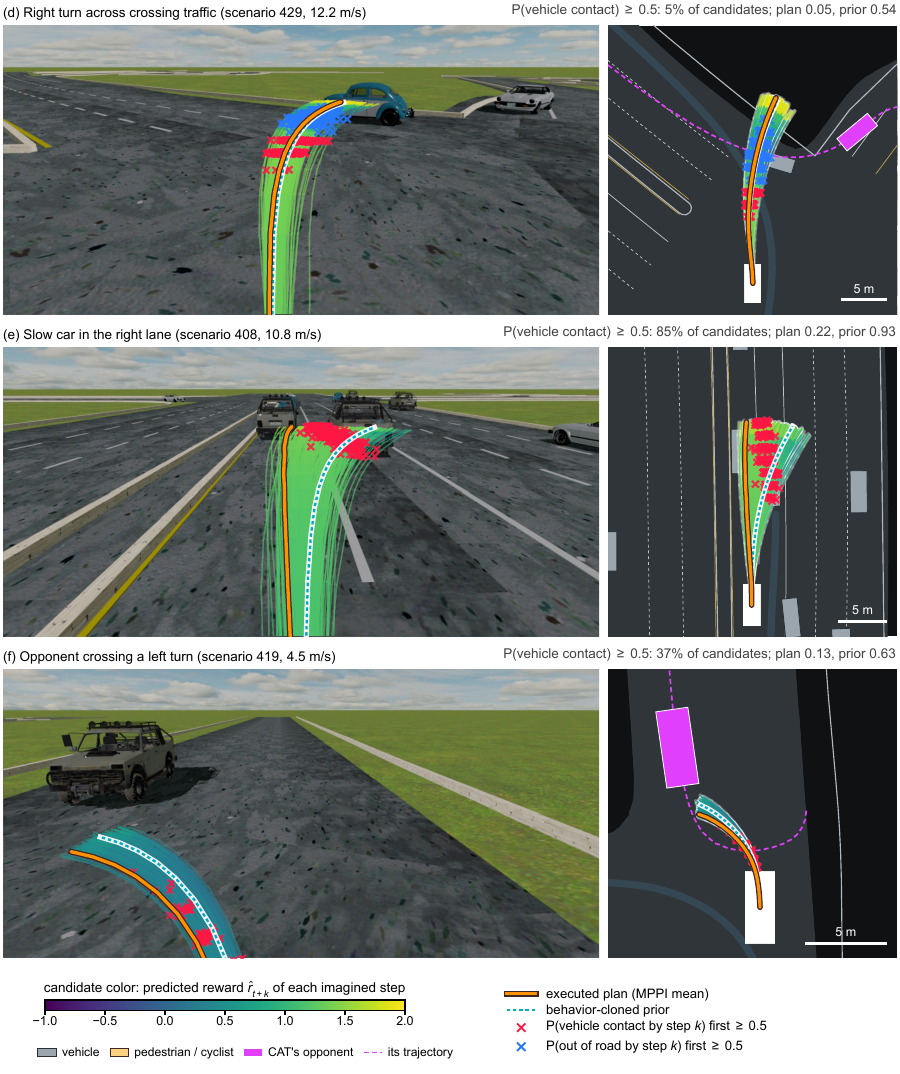}
  \caption{Planning steps of the full system on CAT's MetaDrive benchmark in safety-critical scenes, drawn as in
  Fig.~\ref{fig:md_closed_loop_normal}; magenta: CAT's opponent in the top-down view, and its generated
  trajectory. (d)~Right turn across crossing traffic; (e)~a slow car in the lane to the right, in regular traffic; (f)~the
  opponent turning across a left turn. In each scene the behavior-cloned prior is among the candidates predicted
  to make contact, and the executed plan is not.}
  \label{fig:md_closed_loop_critical}
\end{figure}

\paragraph{MetaDrive, regular traffic}
Figs.~\ref{fig:md_closed_loop_normal} and~\ref{fig:md_closed_loop_critical} show planning steps of the first
of the three systems of Table~\ref{tab:metadrive} on test scenarios of CAT's benchmark (evaluation seed $0$), in
regular traffic and in safety-critical scenes. The agent observes the $101$-dimensional vector of Section~\ref{sec:method:setting}; the front camera,
placed as in CARLA, and the top-down view, drawn from the simulator's state, are added for these figures only.
Candidates are drawn as in Fig.~\ref{fig:closed_loop_normal}, with two differences. The color scale spans $-1$ to
$2$, since CAT's reward is the progress along the route in meters per $0.1$\,s step, and blue crosses mark the
step at which the predicted probability of leaving the road, the risk head's third event, first reaches $0.5$; red
crosses refer to vehicle contact, the event of CAT's crash rate. In the scenes of
Fig.~\ref{fig:md_closed_loop_normal} no candidate approaches a vehicle contact. At a junction with an oncoming car
(a) the candidates differ little, with predicted rewards between $0.95$ and $1.05$ per step. 
In a right turn at a junction at $8.3$\,m/s (b), the candidates spread over $3.9$\,m at the end of the horizon
and all stay at least $4.9$\,m from the road edges; the plan follows the route, ending $0.4$\,m from it, while the
prior ends $1.2$\,m further to the right and is predicted a slightly lower return ($26.3$ against $27.6$). On a three-lane road (c), $8\%$ of the
candidates are predicted to leave the road about $1.5$\,s ahead; they end further to the left than the others
($5.2$ against $3.9$\,m on average). The prior also bends to the left, ending $5.4$\,m to the left against
$2.4$\,m for the plan, with a probability of leaving the road of $0.16$ against $0.03$ ($J=17.4$ against
$27.0$).

\paragraph{MetaDrive, safety-critical scenes}
In each scene of Fig.~\ref{fig:md_closed_loop_critical}, the behavior-cloned prior is predicted to make contact
with a vehicle within the horizon (probability $0.54$, $0.93$ and $0.63$) and the plan is not ($0.05$, $0.22$ and
$0.13$); in each, the prior keeps full throttle and the plan accelerates less (mean command over the first
$0.8$\,s $0.85$, $0.90$ and $0.85$). (d)~In a right turn at $12.2$\,m/s, a car crosses from the right $14$\,m
ahead and CAT's opponent approaches from the right, $21$\,m away; $5\%$ of the candidates are predicted to make
contact, about $0.7$\,s ahead. 
(e)~In regular traffic, at $10.8$\,m/s, a slow car ($3.8$\,m/s) drives in the lane to the right, $13$\,m
ahead, and a second car $17$\,m ahead in the ego's lane. $85\%$ of the candidates, those that drift into the lane
to the right, are predicted to make contact $1.0$--$1.6$\,s ahead, where the slow car will be; the prior ends
$4.5$\,m to the right, level with the slow car, while the plan keeps its lane behind the second car. (f)~During a left turn at $4.5$\,m/s, CAT's opponent
approaches from ahead on a trajectory that turns across the ego's path; $37\%$ of the candidates are predicted to
make contact. Of the score gap between the plan and the prior, the risk term accounts for $11.9$ of $16.1$ in
(d), $10.6$ of $38.8$ in (e) and $13.4$ of $15.3$ in (f). All three episodes reach the end of the route.

\section{Implementation details}
\label{app:impl}
\setcounter{table}{0}

Table~\ref{tab:hparams} lists the hyper-parameters of the evaluated CARLA system: the parameters of the data
collection, the architecture of the world model, the two offline training stages, the planner and the online
adaptation. The MetaDrive system uses the same settings except those listed in Table~\ref{tab:hparams_md}.

\begin{table}[tbp]
  \centering
  \caption{Hyper-parameters of the evaluated CARLA system.}
  \label{tab:hparams}
  \small
  \begin{tabularx}{\linewidth}{lX}
    \toprule
    \textbf{Data collection} & \\
    action noise & clean phases $25$--$100$ steps, corrupt phases $5$--$30$ steps; $\sigma_\alpha\sim\mathcal{U}[0.03,1.0]$, $\sigma_\delta\sim\mathcal{U}[0.005,0.30]$, redrawn every $500$--$1000$ steps; $2.5\%$ clean episodes \\
    scripted events & hold $30$--$80$ steps, late brake $5$--$10$ steps, recovery $30$--$60$ steps \\
    \midrule
    \textbf{World model} & \\
    latent width / decoder & $256$ / 4 layers, 4 heads, MLP ratio 4 \\
    context $\ctx$, horizon $\horizon$ & $16$, $16$ steps ($1.6$\,s at $10$\,Hz) \\
    reward, value bins & $B=101$ bins, uniform in $\symlog$ space on $[-3,3]$ and $[-10,10]$ \\
    reward factors & scaled softplus; scales $6$, $6$, $1$, $1$, $6$ (speed, desired speed, lateral distance, heading error, terminal penalty) \\
    value heads $K$ & $5$ \\
    policy decoder & 3 layers; $\log\bm{\sigma}^{\pi}$ clamped to $[-2,1]$ \\
    risk head & MLP $512\to256\to256\to3$, dropout $0.1$, $w^{+}=20$, $\beta\sim\mathcal{U}[0.1,0.5]$ \\
    \midrule
    \textbf{Offline training} & \\
    stage 1 (world model) & $1.0$\,M steps, batch $32$ windows, lr $2\times10^{-4}\to10^{-4}$ (10k warm-up, cosine over 800k), dropout $0.05$ (value heads $0.01$) \\
    stage 2 (joint, + risk head) & $0.5$\,M steps, lr $10^{-4}$ (2k warm-up), risk head lr $10^{-3}$, world-model dropout $0$ \\
    loss weights & Eq.~\eqref{eq:loss_total}; RMS decay $0.99$ \\
    optimizer & AdamW, weight decay $10^{-4}$, gradient clip $10$, linear warm-up and cosine decay \\
    \midrule
    \textbf{Planner} & \\
    samples $N$ / prior proposals $N_{\mathrm{bc}}$ / elites $N_e$ & $128$ / $6$ / $16$; iterations $I=2$ \\
    $\bm{\sigma}_0$, $\bm{\sigma}_{\min}$, $\bm{\sigma}_{\mathrm{bc}}$ & $(0.25,0.08)$, $(0.05,0.02)$, $(0.10,0.04)$ \\
    sampling noise & independent per step and dimension \\
    $\kappa$, $\gamma$ & $0.25$, $0.985$ \\
    risk weight $W$, event weights $\bm{\omega}$ & $50$, $(1,0,0)$ (collision only) \\
    \midrule
    \textbf{Online adaptation} & \\
    environment steps, burst, UTD & $10^{6}$, $256$, $0.125$ \\
    buffer, batch mix & the $60$ most recent trajectory segments, an episode being cut at every burst (at most $3{,}840$ transitions); $32$ windows, half offline and half online, half of the online from the latest burst \\
    learning rates (dynamics / scorers / risk) & $3\times10^{-5}$ / $10^{-5}$ / $10^{-4}$, gradient clip $1$, weight decay $10^{-5}$ (risk head $10^{-4}$) \\
    EMA & $\tau=0.1$ per $1{,}000$ steps (scorers), $0.1$ per burst (risk head) \\
    frozen & encoder, action prior, value ensemble \\
    \bottomrule
  \end{tabularx}
\end{table}

\begin{table}[tbp]
  \centering
  \caption{Settings of the evaluated MetaDrive system that differ from those of the CARLA system; all other
  settings are those of Table~\ref{tab:hparams}.}
  \label{tab:hparams_md}
  \small
  \begin{tabularx}{\linewidth}{lX}
    \toprule
    \textbf{Data collection} & \\
    scripted events & hazard brake instead of hold and late brake: $5$--$10$ steps at a brake level drawn from $\mathcal{U}[0.3,1]$, triggered when a road user ahead on the route is within $20$\,m and at most $3$\,s from collision, the ego drives at $2$--$10$\,m/s and the expert's acceleration command is above $-0.3$; recovery $30$--$60$ steps \\
    value labels & policy evaluation of the noise-free expert: its two Q-networks, initialized from its critic, fitted on all transitions of the corpus to one-step targets with the expert's action at the next state; $10^{5}$ steps, batch $1{,}024$, lr $3\times10^{-4}$, target-network rate $0.005$, discount $0.99$ \\
    \midrule
    \textbf{World model} & \\
    encoder & two-layer MLP, $101\to256\to256$, initialized from the first two layers of the expert's actor \\
    reward factors & progress along the route (linear, scale $2$\,m) and the indicators of arrival, leaving the road, vehicle contact and object contact (sigmoid) \\
    \midrule
    \textbf{Offline training} & \\
    schedule & one stage of $0.25$\,M steps with the settings of stage 1 (cosine decay over 200k), in which the risk head is trained from the start at its stage-2 learning rate; no scene-factor head \\
    training windows & a keyframe at every step, at the start of an episode with the missing context padded; no additional windows before terminals \\
    \midrule
    \textbf{Planner} & \\
    $\gamma$, event weights $\bm{\omega}$ & $0.99$, $(1,1,1)$ \\
    warm start & none: every step starts from the prior's current chunk \\
    \midrule
    \textbf{Online adaptation} & \\
    episodes & pairs on randomly drawn training scenarios: logged traffic, then CAT's opponent generated against that drive \\
    buffer, batch mix & the most recent finished episodes, about $3{,}840$ transitions, and the episodes being driven; $16$ windows, mixed as in CARLA \\
    risk-head loss & without the braking-monotonicity term \\
    \bottomrule
  \end{tabularx}
\end{table}

\bibliographystyle{elsarticle-harv}
\bibliography{refs}

\begin{thebibliography}{79}
\expandafter\ifx\csname natexlab\endcsname\relax\def\natexlab#1{#1}\fi
\providecommand{\url}[1]{\texttt{#1}}
\providecommand{\href}[2]{#2}
\providecommand{\path}[1]{#1}
\providecommand{\DOIprefix}{doi:}
\providecommand{\ArXivprefix}{arXiv:}
\providecommand{\URLprefix}{URL: }
\providecommand{\Pubmedprefix}{pmid:}
\providecommand{\doi}[1]{\href{http://dx.doi.org/#1}{\path{#1}}}
\providecommand{\Pubmed}[1]{\href{pmid:#1}{\path{#1}}}
\providecommand{\bibinfo}[2]{#2}
\ifx\xfnm\relax \def\xfnm[#1]{\unskip,\space#1}\fi
\bibitem[{Bansal et~al.(2019)Bansal, Krizhevsky and
  Ogale}]{bansal2019chauffeurnet}
\bibinfo{author}{Bansal, M.}, \bibinfo{author}{Krizhevsky, A.},
  \bibinfo{author}{Ogale, A.}, \bibinfo{year}{2019}.
\newblock \bibinfo{title}{{ChauffeurNet}: Learning to drive by imitating the
  best and synthesizing the worst}, in: \bibinfo{booktitle}{Robotics: Science
  and Systems XV}.
\newblock \DOIprefix\doi{10.15607/RSS.2019.XV.031}.
\bibitem[{Berger-Tal et~al.(2014)Berger-Tal, Nathan, Meron and
  Saltz}]{bergertal2014exploration}
\bibinfo{author}{Berger-Tal, O.}, \bibinfo{author}{Nathan, J.},
  \bibinfo{author}{Meron, E.}, \bibinfo{author}{Saltz, D.},
  \bibinfo{year}{2014}.
\newblock \bibinfo{title}{The exploration-exploitation dilemma: A
  multidisciplinary framework}.
\newblock \bibinfo{journal}{PLoS ONE} \bibinfo{volume}{9},
  \bibinfo{pages}{e95693}.
\newblock \DOIprefix\doi{10.1371/journal.pone.0095693}.
\bibitem[{{CARLA Team}(2020)}]{carla2020leaderboard}
\bibinfo{author}{{CARLA Team}}, \bibinfo{year}{2020}.
\newblock \bibinfo{title}{{CARLA} autonomous driving leaderboard}.
\newblock \bibinfo{howpublished}{\url{https://leaderboard.carla.org}}.
\bibitem[{Chekroun et~al.(2023)Chekroun, Toromanoff, Hornauer and
  Moutarde}]{chekroun2023gri}
\bibinfo{author}{Chekroun, R.}, \bibinfo{author}{Toromanoff, M.},
  \bibinfo{author}{Hornauer, S.}, \bibinfo{author}{Moutarde, F.},
  \bibinfo{year}{2023}.
\newblock \bibinfo{title}{{GRI}: General reinforced imitation and its
  application to vision-based autonomous driving}.
\newblock \bibinfo{journal}{Robotics} \bibinfo{volume}{12},
  \bibinfo{pages}{127}.
\newblock \DOIprefix\doi{10.3390/robotics12050127}.
\bibitem[{Chen et~al.(2021)Chen, Koltun and
  Kr{\"a}henb{\"u}hl}]{chen2021worldonrails}
\bibinfo{author}{Chen, D.}, \bibinfo{author}{Koltun, V.},
  \bibinfo{author}{Kr{\"a}henb{\"u}hl, P.}, \bibinfo{year}{2021}.
\newblock \bibinfo{title}{Learning to drive from a world on rails}, in:
  \bibinfo{booktitle}{IEEE/CVF International Conference on Computer Vision
  (ICCV)}, pp. \bibinfo{pages}{15570--15579}.
\newblock \DOIprefix\doi{10.1109/ICCV48922.2021.01530}.
\bibitem[{Chen and Kr{\"a}henb{\"u}hl(2022)}]{chen2022lav}
\bibinfo{author}{Chen, D.}, \bibinfo{author}{Kr{\"a}henb{\"u}hl, P.},
  \bibinfo{year}{2022}.
\newblock \bibinfo{title}{Learning from all vehicles}, in:
  \bibinfo{booktitle}{IEEE/CVF Conference on Computer Vision and Pattern
  Recognition (CVPR)}, pp. \bibinfo{pages}{17201--17210}.
\newblock \DOIprefix\doi{10.1109/CVPR52688.2022.01671}.
\bibitem[{Chen et~al.(2020)Chen, Zhou, Koltun and
  Kr{\"a}henb{\"u}hl}]{chen2020lbc}
\bibinfo{author}{Chen, D.}, \bibinfo{author}{Zhou, B.},
  \bibinfo{author}{Koltun, V.}, \bibinfo{author}{Kr{\"a}henb{\"u}hl, P.},
  \bibinfo{year}{2020}.
\newblock \bibinfo{title}{Learning by cheating}, in:
  \bibinfo{booktitle}{Proceedings of the Conference on Robot Learning (CoRL
  2019)}, pp. \bibinfo{pages}{66--75}.
\bibitem[{Chitta et~al.(2021)Chitta, Prakash and Geiger}]{chitta2021neat}
\bibinfo{author}{Chitta, K.}, \bibinfo{author}{Prakash, A.},
  \bibinfo{author}{Geiger, A.}, \bibinfo{year}{2021}.
\newblock \bibinfo{title}{{NEAT}: Neural attention fields for end-to-end
  autonomous driving}, in: \bibinfo{booktitle}{IEEE/CVF International
  Conference on Computer Vision (ICCV)}, pp. \bibinfo{pages}{15773--15783}.
\newblock \DOIprefix\doi{10.1109/ICCV48922.2021.01550}.
\bibitem[{Chitta et~al.(2023)Chitta, Prakash, Jaeger, Yu, Renz and
  Geiger}]{chitta2023transfuser}
\bibinfo{author}{Chitta, K.}, \bibinfo{author}{Prakash, A.},
  \bibinfo{author}{Jaeger, B.}, \bibinfo{author}{Yu, Z.},
  \bibinfo{author}{Renz, K.}, \bibinfo{author}{Geiger, A.},
  \bibinfo{year}{2023}.
\newblock \bibinfo{title}{{TransFuser}: Imitation with transformer-based sensor
  fusion for autonomous driving}.
\newblock \bibinfo{journal}{IEEE Transactions on Pattern Analysis and Machine
  Intelligence} \bibinfo{volume}{45}, \bibinfo{pages}{12878--12895}.
\newblock \DOIprefix\doi{10.1109/TPAMI.2022.3200245}.
\bibitem[{Chua et~al.(2018)Chua, Calandra, McAllister and
  Levine}]{chua2018pets}
\bibinfo{author}{Chua, K.}, \bibinfo{author}{Calandra, R.},
  \bibinfo{author}{McAllister, R.}, \bibinfo{author}{Levine, S.},
  \bibinfo{year}{2018}.
\newblock \bibinfo{title}{Deep reinforcement learning in a handful of trials
  using probabilistic dynamics models}, in: \bibinfo{booktitle}{Advances in
  Neural Information Processing Systems (NeurIPS)}, pp.
  \bibinfo{pages}{4754--4765}.
\bibitem[{Codevilla et~al.(2018)Codevilla, M{\"u}ller, L{\'o}pez, Koltun and
  Dosovitskiy}]{codevilla2018cil}
\bibinfo{author}{Codevilla, F.}, \bibinfo{author}{M{\"u}ller, M.},
  \bibinfo{author}{L{\'o}pez, A.}, \bibinfo{author}{Koltun, V.},
  \bibinfo{author}{Dosovitskiy, A.}, \bibinfo{year}{2018}.
\newblock \bibinfo{title}{End-to-end driving via conditional imitation
  learning}, in: \bibinfo{booktitle}{IEEE International Conference on Robotics
  and Automation (ICRA)}, pp. \bibinfo{pages}{4693--4700}.
\newblock \DOIprefix\doi{10.1109/ICRA.2018.8460487}.
\bibitem[{Codevilla et~al.(2019)Codevilla, Santana, L{\'o}pez and
  Gaidon}]{codevilla2019limitations}
\bibinfo{author}{Codevilla, F.}, \bibinfo{author}{Santana, E.},
  \bibinfo{author}{L{\'o}pez, A.M.}, \bibinfo{author}{Gaidon, A.},
  \bibinfo{year}{2019}.
\newblock \bibinfo{title}{Exploring the limitations of behavior cloning for
  autonomous driving}, in: \bibinfo{booktitle}{IEEE/CVF International
  Conference on Computer Vision (ICCV)}, pp. \bibinfo{pages}{9328--9337}.
\newblock \DOIprefix\doi{10.1109/ICCV.2019.00942}.
\bibitem[{Curi et~al.(2020)Curi, Berkenkamp and Krause}]{curi2020hucrl}
\bibinfo{author}{Curi, S.}, \bibinfo{author}{Berkenkamp, F.},
  \bibinfo{author}{Krause, A.}, \bibinfo{year}{2020}.
\newblock \bibinfo{title}{Efficient model-based reinforcement learning through
  optimistic policy search and planning}, in: \bibinfo{booktitle}{Advances in
  Neural Information Processing Systems}, pp. \bibinfo{pages}{14156--14170}.
\bibitem[{Ding et~al.(2020)Ding, Chen, Xu and Zhao}]{ding2020learningtocollide}
\bibinfo{author}{Ding, W.}, \bibinfo{author}{Chen, B.}, \bibinfo{author}{Xu,
  M.}, \bibinfo{author}{Zhao, D.}, \bibinfo{year}{2020}.
\newblock \bibinfo{title}{Learning to collide: An adaptive safety-critical
  scenarios generating method}, in: \bibinfo{booktitle}{IEEE/RSJ International
  Conference on Intelligent Robots and Systems (IROS)}, pp.
  \bibinfo{pages}{2243--2250}.
\newblock \DOIprefix\doi{10.1109/IROS45743.2020.9340696}.
\bibitem[{Dosovitskiy et~al.(2017)Dosovitskiy, Ros, Codevilla, L{\'o}pez and
  Koltun}]{dosovitskiy2017carla}
\bibinfo{author}{Dosovitskiy, A.}, \bibinfo{author}{Ros, G.},
  \bibinfo{author}{Codevilla, F.}, \bibinfo{author}{L{\'o}pez, A.},
  \bibinfo{author}{Koltun, V.}, \bibinfo{year}{2017}.
\newblock \bibinfo{title}{{CARLA}: An open urban driving simulator}, in:
  \bibinfo{booktitle}{Proceedings of the 1st Annual Conference on Robot
  Learning (CoRL)}, pp. \bibinfo{pages}{1--16}.
\bibitem[{Ettinger et~al.(2021)Ettinger, Cheng, Caine, Liu, Zhao, Pradhan,
  Chai, Sapp, Qi, Zhou, Yang, Chouard, Sun, Ngiam, Vasudevan, McCauley, Shlens
  and Anguelov}]{ettinger2021womd}
\bibinfo{author}{Ettinger, S.}, \bibinfo{author}{Cheng, S.},
  \bibinfo{author}{Caine, B.}, \bibinfo{author}{Liu, C.},
  \bibinfo{author}{Zhao, H.}, \bibinfo{author}{Pradhan, S.},
  \bibinfo{author}{Chai, Y.}, \bibinfo{author}{Sapp, B.}, \bibinfo{author}{Qi,
  C.R.}, \bibinfo{author}{Zhou, Y.}, \bibinfo{author}{Yang, Z.},
  \bibinfo{author}{Chouard, A.}, \bibinfo{author}{Sun, P.},
  \bibinfo{author}{Ngiam, J.}, \bibinfo{author}{Vasudevan, V.},
  \bibinfo{author}{McCauley, A.}, \bibinfo{author}{Shlens, J.},
  \bibinfo{author}{Anguelov, D.}, \bibinfo{year}{2021}.
\newblock \bibinfo{title}{Large scale interactive motion forecasting for
  autonomous driving: The {Waymo} open motion dataset}, in:
  \bibinfo{booktitle}{IEEE/CVF International Conference on Computer Vision
  (ICCV)}, pp. \bibinfo{pages}{9690--9699}.
\newblock \DOIprefix\doi{10.1109/ICCV48922.2021.00957}.
\bibitem[{Fu et~al.(2018)Fu, Luo and Levine}]{fu2018airl}
\bibinfo{author}{Fu, J.}, \bibinfo{author}{Luo, K.}, \bibinfo{author}{Levine,
  S.}, \bibinfo{year}{2018}.
\newblock \bibinfo{title}{Learning robust rewards with adversarial inverse
  reinforcement learning}, in: \bibinfo{booktitle}{International Conference on
  Learning Representations (ICLR)}.
\bibitem[{Fujimoto and Gu(2021)}]{fujimoto2021td3bc}
\bibinfo{author}{Fujimoto, S.}, \bibinfo{author}{Gu, S.S.},
  \bibinfo{year}{2021}.
\newblock \bibinfo{title}{A minimalist approach to offline reinforcement
  learning}, in: \bibinfo{booktitle}{Advances in Neural Information Processing
  Systems (NeurIPS)}, pp. \bibinfo{pages}{20132--20145}.
\bibitem[{Fujimoto et~al.(2018)Fujimoto, van Hoof and Meger}]{fujimoto2018td3}
\bibinfo{author}{Fujimoto, S.}, \bibinfo{author}{van Hoof, H.},
  \bibinfo{author}{Meger, D.}, \bibinfo{year}{2018}.
\newblock \bibinfo{title}{Addressing function approximation error in
  actor-critic methods}, in: \bibinfo{booktitle}{Proceedings of the 35th
  International Conference on Machine Learning (ICML)}, pp.
  \bibinfo{pages}{1587--1596}.
\bibitem[{Gleave et~al.(2022)Gleave, Taufeeque, Rocamonde, Jenner, Wang, Toyer,
  Ernestus, Belrose, Emmons and Russell}]{gleave2022imitation}
\bibinfo{author}{Gleave, A.}, \bibinfo{author}{Taufeeque, M.},
  \bibinfo{author}{Rocamonde, J.}, \bibinfo{author}{Jenner, E.},
  \bibinfo{author}{Wang, S.H.}, \bibinfo{author}{Toyer, S.},
  \bibinfo{author}{Ernestus, M.}, \bibinfo{author}{Belrose, N.},
  \bibinfo{author}{Emmons, S.}, \bibinfo{author}{Russell, S.},
  \bibinfo{year}{2022}.
\newblock \bibinfo{title}{imitation: Clean imitation learning implementations}.
\newblock \bibinfo{journal}{arXiv preprint arXiv:2211.11972} .
\bibitem[{Guo and Langari(2026)}]{guo2026corrriskwm}
\bibinfo{author}{Guo, T.}, \bibinfo{author}{Langari, R.}, \bibinfo{year}{2026}.
\newblock \bibinfo{title}{{CorrRisk-WM}: Corridor-conditioned risk world
  modeling for safety-critical trajectory planning}.
\newblock \bibinfo{journal}{arXiv preprint arXiv:2609.16724} .
\bibitem[{Ha and Schmidhuber(2018)}]{ha2018worldmodels}
\bibinfo{author}{Ha, D.}, \bibinfo{author}{Schmidhuber, J.},
  \bibinfo{year}{2018}.
\newblock \bibinfo{title}{World models}.
\newblock \bibinfo{journal}{arXiv preprint arXiv:1803.10122} .
\bibitem[{de~Haan et~al.(2019)de~Haan, Jayaraman and Levine}]{dehaan2019causal}
\bibinfo{author}{de~Haan, P.}, \bibinfo{author}{Jayaraman, D.},
  \bibinfo{author}{Levine, S.}, \bibinfo{year}{2019}.
\newblock \bibinfo{title}{Causal confusion in imitation learning}, in:
  \bibinfo{booktitle}{Advances in Neural Information Processing Systems
  (NeurIPS)}, pp. \bibinfo{pages}{11698--11709}.
\bibitem[{Haarnoja et~al.(2018)Haarnoja, Zhou, Abbeel and
  Levine}]{haarnoja2018sac}
\bibinfo{author}{Haarnoja, T.}, \bibinfo{author}{Zhou, A.},
  \bibinfo{author}{Abbeel, P.}, \bibinfo{author}{Levine, S.},
  \bibinfo{year}{2018}.
\newblock \bibinfo{title}{Soft actor-critic: Off-policy maximum entropy deep
  reinforcement learning with a stochastic actor}, in:
  \bibinfo{booktitle}{Proceedings of the 35th International Conference on
  Machine Learning (ICML)}, pp. \bibinfo{pages}{1861--1870}.
\bibitem[{Hafner et~al.(2020)Hafner, Lillicrap, Ba and
  Norouzi}]{hafner2020dreamer}
\bibinfo{author}{Hafner, D.}, \bibinfo{author}{Lillicrap, T.},
  \bibinfo{author}{Ba, J.}, \bibinfo{author}{Norouzi, M.},
  \bibinfo{year}{2020}.
\newblock \bibinfo{title}{Dream to control: Learning behaviors by latent
  imagination}, in: \bibinfo{booktitle}{International Conference on Learning
  Representations (ICLR)}.
\bibitem[{Hafner et~al.(2019)Hafner, Lillicrap, Fischer, Villegas, Ha, Lee and
  Davidson}]{hafner2019planet}
\bibinfo{author}{Hafner, D.}, \bibinfo{author}{Lillicrap, T.},
  \bibinfo{author}{Fischer, I.}, \bibinfo{author}{Villegas, R.},
  \bibinfo{author}{Ha, D.}, \bibinfo{author}{Lee, H.},
  \bibinfo{author}{Davidson, J.}, \bibinfo{year}{2019}.
\newblock \bibinfo{title}{Learning latent dynamics for planning from pixels},
  in: \bibinfo{booktitle}{Proceedings of the 36th International Conference on
  Machine Learning (ICML)}, pp. \bibinfo{pages}{2555--2565}.
\bibitem[{Hafner et~al.(2021)Hafner, Lillicrap, Norouzi and
  Ba}]{hafner2021dreamerv2}
\bibinfo{author}{Hafner, D.}, \bibinfo{author}{Lillicrap, T.},
  \bibinfo{author}{Norouzi, M.}, \bibinfo{author}{Ba, J.},
  \bibinfo{year}{2021}.
\newblock \bibinfo{title}{Mastering {Atari} with discrete world models}, in:
  \bibinfo{booktitle}{International Conference on Learning Representations
  (ICLR)}.
\bibitem[{Hafner et~al.(2025a)Hafner, Pasukonis, Ba and
  Lillicrap}]{hafner2023dreamerv3}
\bibinfo{author}{Hafner, D.}, \bibinfo{author}{Pasukonis, J.},
  \bibinfo{author}{Ba, J.}, \bibinfo{author}{Lillicrap, T.},
  \bibinfo{year}{2025}a.
\newblock \bibinfo{title}{Mastering diverse control tasks through world
  models}.
\newblock \bibinfo{journal}{Nature} \bibinfo{volume}{640},
  \bibinfo{pages}{647--653}.
\newblock \DOIprefix\doi{10.1038/s41586-025-08744-2}.
\bibitem[{Hafner et~al.(2025b)Hafner, Yan and Lillicrap}]{hafner2025dreamer4}
\bibinfo{author}{Hafner, D.}, \bibinfo{author}{Yan, W.},
  \bibinfo{author}{Lillicrap, T.}, \bibinfo{year}{2025}b.
\newblock \bibinfo{title}{Training agents inside of scalable world models}.
\newblock \bibinfo{journal}{arXiv preprint arXiv:2509.24527} .
\bibitem[{Hansen et~al.(2022)Hansen, Su and Wang}]{hansen2022tdmpc}
\bibinfo{author}{Hansen, N.}, \bibinfo{author}{Su, H.}, \bibinfo{author}{Wang,
  X.}, \bibinfo{year}{2022}.
\newblock \bibinfo{title}{Temporal difference learning for model predictive
  control}, in: \bibinfo{booktitle}{Proceedings of the 39th International
  Conference on Machine Learning (ICML)}, pp. \bibinfo{pages}{8387--8406}.
\bibitem[{Hansen et~al.(2024)Hansen, Su and Wang}]{hansen2024tdmpc2}
\bibinfo{author}{Hansen, N.}, \bibinfo{author}{Su, H.}, \bibinfo{author}{Wang,
  X.}, \bibinfo{year}{2024}.
\newblock \bibinfo{title}{{TD-MPC2}: Scalable, robust world models for
  continuous control}, in: \bibinfo{booktitle}{International Conference on
  Learning Representations (ICLR)}.
\bibitem[{Hills et~al.(2015)Hills, Todd, Lazer, Redish and
  Couzin}]{hills2015exploration}
\bibinfo{author}{Hills, T.T.}, \bibinfo{author}{Todd, P.M.},
  \bibinfo{author}{Lazer, D.}, \bibinfo{author}{Redish, A.D.},
  \bibinfo{author}{Couzin, I.D.}, \bibinfo{year}{2015}.
\newblock \bibinfo{title}{Exploration versus exploitation in space, mind, and
  society}.
\newblock \bibinfo{journal}{Trends in Cognitive Sciences} \bibinfo{volume}{19},
  \bibinfo{pages}{46--54}.
\newblock \DOIprefix\doi{10.1016/j.tics.2014.10.004}.
\bibitem[{Ho and Ermon(2016)}]{ho2016gail}
\bibinfo{author}{Ho, J.}, \bibinfo{author}{Ermon, S.}, \bibinfo{year}{2016}.
\newblock \bibinfo{title}{Generative adversarial imitation learning}, in:
  \bibinfo{booktitle}{Advances in Neural Information Processing Systems
  (NeurIPS)}, pp. \bibinfo{pages}{4565--4573}.
\bibitem[{Hu et~al.(2022)Hu, Corrado, Griffiths, Murez, Gurau, Yeo, Kendall,
  Cipolla and Shotton}]{hu2022mile}
\bibinfo{author}{Hu, A.}, \bibinfo{author}{Corrado, G.},
  \bibinfo{author}{Griffiths, N.}, \bibinfo{author}{Murez, Z.},
  \bibinfo{author}{Gurau, C.}, \bibinfo{author}{Yeo, H.},
  \bibinfo{author}{Kendall, A.}, \bibinfo{author}{Cipolla, R.},
  \bibinfo{author}{Shotton, J.}, \bibinfo{year}{2022}.
\newblock \bibinfo{title}{Model-based imitation learning for urban driving},
  in: \bibinfo{booktitle}{Advances in Neural Information Processing Systems
  (NeurIPS)}, pp. \bibinfo{pages}{20703--20716}.
\newblock \DOIprefix\doi{10.52202/068431-1505}.
\bibitem[{Hu et~al.(2023)Hu, Russell, Yeo, Murez, Fedoseev, Kendall, Shotton
  and Corrado}]{hu2023gaia1}
\bibinfo{author}{Hu, A.}, \bibinfo{author}{Russell, L.}, \bibinfo{author}{Yeo,
  H.}, \bibinfo{author}{Murez, Z.}, \bibinfo{author}{Fedoseev, G.},
  \bibinfo{author}{Kendall, A.}, \bibinfo{author}{Shotton, J.},
  \bibinfo{author}{Corrado, G.}, \bibinfo{year}{2023}.
\newblock \bibinfo{title}{{GAIA-1}: A generative world model for autonomous
  driving}.
\newblock \bibinfo{journal}{arXiv preprint arXiv:2309.17080} .
\bibitem[{Huang et~al.(2025)Huang, Sheng, Qu, You and Chen}]{huang2025vlmrl}
\bibinfo{author}{Huang, Z.}, \bibinfo{author}{Sheng, Z.}, \bibinfo{author}{Qu,
  Y.}, \bibinfo{author}{You, J.}, \bibinfo{author}{Chen, S.},
  \bibinfo{year}{2025}.
\newblock \bibinfo{title}{{VLM-RL}: A unified vision language models and
  reinforcement learning framework for safe autonomous driving}.
\newblock \bibinfo{journal}{Transportation Research Part C: Emerging
  Technologies} \bibinfo{volume}{180}, \bibinfo{pages}{105321}.
\newblock \DOIprefix\doi{10.1016/j.trc.2025.105321}.
\bibitem[{Huang et~al.(2026)Huang, Sheng, Wan, Qu, You, Jiang and
  Chen}]{huang2026drivevlmrl}
\bibinfo{author}{Huang, Z.}, \bibinfo{author}{Sheng, Z.}, \bibinfo{author}{Wan,
  Z.}, \bibinfo{author}{Qu, Y.}, \bibinfo{author}{You, J.},
  \bibinfo{author}{Jiang, S.}, \bibinfo{author}{Chen, S.},
  \bibinfo{year}{2026}.
\newblock \bibinfo{title}{{DriveVLM-RL}: Neuroscience-inspired reinforcement
  learning with vision-language models for safe and deployable autonomous
  driving}.
\newblock \bibinfo{journal}{arXiv preprint arXiv:2603.18315} .
\bibitem[{Jaeger et~al.(2023)Jaeger, Chitta and
  Geiger}]{jaeger2023hiddenbiases}
\bibinfo{author}{Jaeger, B.}, \bibinfo{author}{Chitta, K.},
  \bibinfo{author}{Geiger, A.}, \bibinfo{year}{2023}.
\newblock \bibinfo{title}{Hidden biases of end-to-end driving models}, in:
  \bibinfo{booktitle}{IEEE/CVF International Conference on Computer Vision
  (ICCV)}, pp. \bibinfo{pages}{8206--8215}.
\newblock \DOIprefix\doi{10.1109/ICCV51070.2023.00757}.
\bibitem[{Janner et~al.(2019)Janner, Fu, Zhang and Levine}]{janner2019mbpo}
\bibinfo{author}{Janner, M.}, \bibinfo{author}{Fu, J.}, \bibinfo{author}{Zhang,
  M.}, \bibinfo{author}{Levine, S.}, \bibinfo{year}{2019}.
\newblock \bibinfo{title}{When to trust your model: Model-based policy
  optimization}, in: \bibinfo{booktitle}{Advances in Neural Information
  Processing Systems (NeurIPS)}, pp. \bibinfo{pages}{12519--12530}.
\bibitem[{Khanzada and Kwon(2025)}]{khanzada2025indrive}
\bibinfo{author}{Khanzada, F.K.}, \bibinfo{author}{Kwon, J.},
  \bibinfo{year}{2025}.
\newblock \bibinfo{title}{{InDRiVE}: Intrinsic disagreement based reinforcement
  for vehicle exploration through curiosity driven generalized world model}.
\newblock \bibinfo{journal}{arXiv preprint arXiv:2503.05573} .
\bibitem[{Kostrikov et~al.(2022)Kostrikov, Nair and Levine}]{kostrikov2022iql}
\bibinfo{author}{Kostrikov, I.}, \bibinfo{author}{Nair, A.},
  \bibinfo{author}{Levine, S.}, \bibinfo{year}{2022}.
\newblock \bibinfo{title}{Offline reinforcement learning with implicit
  {Q}-learning}, in: \bibinfo{booktitle}{International Conference on Learning
  Representations (ICLR)}.
\bibitem[{Laskey et~al.(2017)Laskey, Lee, Fox, Dragan and
  Goldberg}]{laskey2017dart}
\bibinfo{author}{Laskey, M.}, \bibinfo{author}{Lee, J.}, \bibinfo{author}{Fox,
  R.}, \bibinfo{author}{Dragan, A.}, \bibinfo{author}{Goldberg, K.},
  \bibinfo{year}{2017}.
\newblock \bibinfo{title}{{DART}: Noise injection for robust imitation
  learning}, in: \bibinfo{booktitle}{Proceedings of the 1st Annual Conference
  on Robot Learning (CoRL)}, pp. \bibinfo{pages}{143--156}.
\bibitem[{Levy et~al.(2026)Levy, Westenbroek, Huang, Palafox, Yin, Omidshafiei,
  Kim, Gupta and Fridovich-Keil}]{levy2026simdist}
\bibinfo{author}{Levy, J.}, \bibinfo{author}{Westenbroek, T.},
  \bibinfo{author}{Huang, K.}, \bibinfo{author}{Palafox, F.},
  \bibinfo{author}{Yin, P.}, \bibinfo{author}{Omidshafiei, S.},
  \bibinfo{author}{Kim, D.K.}, \bibinfo{author}{Gupta, A.},
  \bibinfo{author}{Fridovich-Keil, D.}, \bibinfo{year}{2026}.
\newblock \bibinfo{title}{Simulation distillation: Pretraining world models in
  simulation for rapid real-world adaptation}, in:
  \bibinfo{booktitle}{Robotics: Science and Systems XXII},
  \bibinfo{address}{Sydney, Australia}.
\newblock \DOIprefix\doi{10.15607/RSS.2026.XXII.017}.
\bibitem[{Li et~al.(2024)Li, Jia, Wang and Yan}]{li2024think2drive}
\bibinfo{author}{Li, Q.}, \bibinfo{author}{Jia, X.}, \bibinfo{author}{Wang,
  S.}, \bibinfo{author}{Yan, J.}, \bibinfo{year}{2024}.
\newblock \bibinfo{title}{{Think2Drive}: Efficient reinforcement learning by
  thinking with latent world model for autonomous driving (in {CARLA-v2})}, in:
  \bibinfo{booktitle}{European Conference on Computer Vision (ECCV)},
  \bibinfo{publisher}{Springer}. pp. \bibinfo{pages}{142--158}.
\newblock \DOIprefix\doi{10.1007/978-3-031-72995-9_9}.
\bibitem[{Li et~al.(2023)Li, Peng, Feng, Zhang, Xue and Zhou}]{li2023metadrive}
\bibinfo{author}{Li, Q.}, \bibinfo{author}{Peng, Z.}, \bibinfo{author}{Feng,
  L.}, \bibinfo{author}{Zhang, Q.}, \bibinfo{author}{Xue, Z.},
  \bibinfo{author}{Zhou, B.}, \bibinfo{year}{2023}.
\newblock \bibinfo{title}{{MetaDrive}: Composing diverse driving scenarios for
  generalizable reinforcement learning}.
\newblock \bibinfo{journal}{IEEE Transactions on Pattern Analysis and Machine
  Intelligence} \bibinfo{volume}{45}, \bibinfo{pages}{3461--3475}.
\newblock \DOIprefix\doi{10.1109/TPAMI.2022.3190471}.
\bibitem[{Li et~al.(2026)Li, Chen, Wang, Liu, Chitta, Yang, Lu, Ye, Qiu, Wang,
  Zou, Peng, Pan, Su, Bursuc, Li, Geiger, Su and Li}]{li2026worldengine}
\bibinfo{author}{Li, T.}, \bibinfo{author}{Chen, L.}, \bibinfo{author}{Wang,
  C.}, \bibinfo{author}{Liu, H.}, \bibinfo{author}{Chitta, K.},
  \bibinfo{author}{Yang, Z.}, \bibinfo{author}{Lu, Y.}, \bibinfo{author}{Ye,
  N.}, \bibinfo{author}{Qiu, Y.}, \bibinfo{author}{Wang, Y.},
  \bibinfo{author}{Zou, L.}, \bibinfo{author}{Peng, J.}, \bibinfo{author}{Pan,
  J.}, \bibinfo{author}{Su, Z.}, \bibinfo{author}{Bursuc, A.},
  \bibinfo{author}{Li, S.E.}, \bibinfo{author}{Geiger, A.},
  \bibinfo{author}{Su, P.}, \bibinfo{author}{Li, H.}, \bibinfo{year}{2026}.
\newblock \bibinfo{title}{World engine: Towards the era of post-training for
  autonomous driving}.
\newblock \bibinfo{journal}{arXiv preprint arXiv:2606.19836} .
\bibitem[{Mnih et~al.(2015)Mnih, Kavukcuoglu, Silver, Rusu, Veness, Bellemare,
  Graves, Riedmiller, Fidjeland, Ostrovski, Petersen, Beattie, Sadik,
  Antonoglou, King, Kumaran, Wierstra, Legg and Hassabis}]{mnih2015dqn}
\bibinfo{author}{Mnih, V.}, \bibinfo{author}{Kavukcuoglu, K.},
  \bibinfo{author}{Silver, D.}, \bibinfo{author}{Rusu, A.A.},
  \bibinfo{author}{Veness, J.}, \bibinfo{author}{Bellemare, M.G.},
  \bibinfo{author}{Graves, A.}, \bibinfo{author}{Riedmiller, M.},
  \bibinfo{author}{Fidjeland, A.K.}, \bibinfo{author}{Ostrovski, G.},
  \bibinfo{author}{Petersen, S.}, \bibinfo{author}{Beattie, C.},
  \bibinfo{author}{Sadik, A.}, \bibinfo{author}{Antonoglou, I.},
  \bibinfo{author}{King, H.}, \bibinfo{author}{Kumaran, D.},
  \bibinfo{author}{Wierstra, D.}, \bibinfo{author}{Legg, S.},
  \bibinfo{author}{Hassabis, D.}, \bibinfo{year}{2015}.
\newblock \bibinfo{title}{Human-level control through deep reinforcement
  learning}.
\newblock \bibinfo{journal}{Nature} \bibinfo{volume}{518},
  \bibinfo{pages}{529--533}.
\newblock \DOIprefix\doi{10.1038/nature14236}.
\bibitem[{Niu et~al.(2024)Niu, Xu, Jiang and Hu}]{niu2024clic}
\bibinfo{author}{Niu, H.}, \bibinfo{author}{Xu, Y.}, \bibinfo{author}{Jiang,
  X.}, \bibinfo{author}{Hu, J.}, \bibinfo{year}{2024}.
\newblock \bibinfo{title}{Continual driving policy optimization with
  closed-loop individualized curricula}, in: \bibinfo{booktitle}{IEEE
  International Conference on Robotics and Automation (ICRA)}, pp.
  \bibinfo{pages}{6850--6857}.
\newblock \DOIprefix\doi{10.1109/ICRA57147.2024.10611578}.
\bibitem[{Pinneri et~al.(2021)Pinneri, Sawant, Blaes, Achterhold, Stueckler,
  Rolinek and Martius}]{pinneri2020icem}
\bibinfo{author}{Pinneri, C.}, \bibinfo{author}{Sawant, S.},
  \bibinfo{author}{Blaes, S.}, \bibinfo{author}{Achterhold, J.},
  \bibinfo{author}{Stueckler, J.}, \bibinfo{author}{Rolinek, M.},
  \bibinfo{author}{Martius, G.}, \bibinfo{year}{2021}.
\newblock \bibinfo{title}{Sample-efficient cross-entropy method for real-time
  planning}, in: \bibinfo{booktitle}{Proceedings of the 2020 Conference on
  Robot Learning (CoRL)}, pp. \bibinfo{pages}{1049--1065}.
\bibitem[{Prakash et~al.(2021)Prakash, Chitta and
  Geiger}]{prakash2021transfuser}
\bibinfo{author}{Prakash, A.}, \bibinfo{author}{Chitta, K.},
  \bibinfo{author}{Geiger, A.}, \bibinfo{year}{2021}.
\newblock \bibinfo{title}{Multi-modal fusion transformer for end-to-end
  autonomous driving}, in: \bibinfo{booktitle}{IEEE/CVF Conference on Computer
  Vision and Pattern Recognition (CVPR)}, pp. \bibinfo{pages}{7073--7083}.
\newblock \DOIprefix\doi{10.1109/CVPR46437.2021.00700}.
\bibitem[{Qu et~al.(2026)Qu, Sheng, Huang, Chen, Luo, Wang, Feng, Labi and
  Chen}]{qu2026foundrl}
\bibinfo{author}{Qu, Y.}, \bibinfo{author}{Sheng, Z.}, \bibinfo{author}{Huang,
  Z.}, \bibinfo{author}{Chen, J.}, \bibinfo{author}{Luo, Y.},
  \bibinfo{author}{Wang, T.}, \bibinfo{author}{Feng, Y.},
  \bibinfo{author}{Labi, S.}, \bibinfo{author}{Chen, S.}, \bibinfo{year}{2026}.
\newblock \bibinfo{title}{{Found-RL}: Foundation model-enhanced reinforcement
  learning via asynchronous {VLM} feedback for autonomous driving}.
\newblock \bibinfo{journal}{Communications in Transportation Research}
  \bibinfo{volume}{6}, \bibinfo{pages}{9640027}.
\newblock \DOIprefix\doi{10.26599/COMMTR.2026.9640027}.
\bibitem[{Ray et~al.(2019)Ray, Achiam and Amodei}]{ray2019safetygym}
\bibinfo{author}{Ray, A.}, \bibinfo{author}{Achiam, J.},
  \bibinfo{author}{Amodei, D.}, \bibinfo{year}{2019}.
\newblock \bibinfo{title}{Benchmarking Safe Exploration in Deep Reinforcement
  Learning}.
\newblock \bibinfo{type}{Technical Report}. OpenAI.
\newblock \URLprefix \url{https://cdn.openai.com/safexp-short.pdf}.
\bibitem[{Reddy et~al.(2020)Reddy, Dragan and Levine}]{reddy2020sqil}
\bibinfo{author}{Reddy, S.}, \bibinfo{author}{Dragan, A.D.},
  \bibinfo{author}{Levine, S.}, \bibinfo{year}{2020}.
\newblock \bibinfo{title}{{SQIL}: Imitation learning via reinforcement learning
  with sparse rewards}, in: \bibinfo{booktitle}{International Conference on
  Learning Representations (ICLR)}.
\bibitem[{Ross et~al.(2011)Ross, Gordon and Bagnell}]{ross2011dagger}
\bibinfo{author}{Ross, S.}, \bibinfo{author}{Gordon, G.},
  \bibinfo{author}{Bagnell, D.}, \bibinfo{year}{2011}.
\newblock \bibinfo{title}{A reduction of imitation learning and structured
  prediction to no-regret online learning}, in: \bibinfo{booktitle}{Proceedings
  of the Fourteenth International Conference on Artificial Intelligence and
  Statistics (AISTATS)}, pp. \bibinfo{pages}{627--635}.
\bibitem[{Schulman et~al.(2017)Schulman, Wolski, Dhariwal, Radford and
  Klimov}]{schulman2017ppo}
\bibinfo{author}{Schulman, J.}, \bibinfo{author}{Wolski, F.},
  \bibinfo{author}{Dhariwal, P.}, \bibinfo{author}{Radford, A.},
  \bibinfo{author}{Klimov, O.}, \bibinfo{year}{2017}.
\newblock \bibinfo{title}{Proximal policy optimization algorithms}.
\newblock \bibinfo{journal}{arXiv preprint arXiv:1707.06347} .
\bibitem[{Sekar et~al.(2020)Sekar, Rybkin, Daniilidis, Abbeel, Hafner and
  Pathak}]{sekar2020plan2explore}
\bibinfo{author}{Sekar, R.}, \bibinfo{author}{Rybkin, O.},
  \bibinfo{author}{Daniilidis, K.}, \bibinfo{author}{Abbeel, P.},
  \bibinfo{author}{Hafner, D.}, \bibinfo{author}{Pathak, D.},
  \bibinfo{year}{2020}.
\newblock \bibinfo{title}{Planning to explore via self-supervised world
  models}, in: \bibinfo{booktitle}{Proceedings of the 37th International
  Conference on Machine Learning}, pp. \bibinfo{pages}{8583--8592}.
\bibitem[{Shao et~al.(2023)Shao, Wang, Chen, Li and Liu}]{shao2023interfuser}
\bibinfo{author}{Shao, H.}, \bibinfo{author}{Wang, L.}, \bibinfo{author}{Chen,
  R.}, \bibinfo{author}{Li, H.}, \bibinfo{author}{Liu, Y.},
  \bibinfo{year}{2023}.
\newblock \bibinfo{title}{Safety-enhanced autonomous driving using
  interpretable sensor fusion transformer}, in: \bibinfo{booktitle}{Proceedings
  of the 6th Conference on Robot Learning (CoRL)}, pp.
  \bibinfo{pages}{726--737}.
\bibitem[{Sheng et~al.(2026a)Sheng, Huang, Qu, Leng, Bhavanam and
  Chen}]{sheng2026curricuvlm}
\bibinfo{author}{Sheng, Z.}, \bibinfo{author}{Huang, Z.}, \bibinfo{author}{Qu,
  Y.}, \bibinfo{author}{Leng, Y.}, \bibinfo{author}{Bhavanam, S.},
  \bibinfo{author}{Chen, S.}, \bibinfo{year}{2026}a.
\newblock \bibinfo{title}{{CurricuVLM}: Towards safe autonomous driving via
  personalized safety-critical curriculum learning with vision-language
  models}.
\newblock \bibinfo{journal}{Transportation Research Part C: Emerging
  Technologies} \bibinfo{volume}{185}, \bibinfo{pages}{105549}.
\newblock \DOIprefix\doi{10.1016/j.trc.2026.105549}.
\bibitem[{Sheng et~al.(2025)Sheng, Huang, Qu, Leng and
  Chen}]{sheng2025talk2traffic}
\bibinfo{author}{Sheng, Z.}, \bibinfo{author}{Huang, Z.}, \bibinfo{author}{Qu,
  Y.}, \bibinfo{author}{Leng, Y.}, \bibinfo{author}{Chen, S.},
  \bibinfo{year}{2025}.
\newblock \bibinfo{title}{{Talk2Traffic}: Interactive and editable traffic
  scenario generation for autonomous driving with multimodal large language
  model}, in: \bibinfo{booktitle}{IEEE/CVF Conference on Computer Vision and
  Pattern Recognition Workshops (CVPRW)}, pp. \bibinfo{pages}{3788--3797}.
\bibitem[{Sheng et~al.(2026b)Sheng, Ye, Luo, Chen and
  Ren}]{sheng2026explorevla}
\bibinfo{author}{Sheng, Z.}, \bibinfo{author}{Ye, X.}, \bibinfo{author}{Luo,
  J.}, \bibinfo{author}{Chen, S.}, \bibinfo{author}{Ren, L.},
  \bibinfo{year}{2026}b.
\newblock \bibinfo{title}{{ExploreVLA}: Dense world modeling and exploration
  for end-to-end autonomous driving}, in: \bibinfo{booktitle}{European
  Conference on Computer Vision (ECCV)}, \bibinfo{publisher}{Springer}. pp.
  \bibinfo{pages}{266--284}.
\newblock \DOIprefix\doi{10.1007/978-3-032-37261-1_16}.
\bibitem[{Smith and Winkler(2006)}]{smith2006optimizerscurse}
\bibinfo{author}{Smith, J.E.}, \bibinfo{author}{Winkler, R.L.},
  \bibinfo{year}{2006}.
\newblock \bibinfo{title}{The optimizer's curse: Skepticism and postdecision
  surprise in decision analysis}.
\newblock \bibinfo{journal}{Management Science} \bibinfo{volume}{52},
  \bibinfo{pages}{311--322}.
\newblock \DOIprefix\doi{10.1287/mnsc.1050.0451}.
\bibitem[{Stooke et~al.(2020)Stooke, Achiam and Abbeel}]{stooke2020pid}
\bibinfo{author}{Stooke, A.}, \bibinfo{author}{Achiam, J.},
  \bibinfo{author}{Abbeel, P.}, \bibinfo{year}{2020}.
\newblock \bibinfo{title}{Responsive safety in reinforcement learning by {PID}
  {Lagrangian} methods}, in: \bibinfo{booktitle}{Proceedings of the 37th
  International Conference on Machine Learning (ICML)}, pp.
  \bibinfo{pages}{9133--9143}.
\bibitem[{Sun et~al.(2026)Sun, Xue, Long, Liu, Hu, Zheng and
  Sebe}]{sun2026rawmpc}
\bibinfo{author}{Sun, J.}, \bibinfo{author}{Xue, F.}, \bibinfo{author}{Long,
  T.}, \bibinfo{author}{Liu, C.}, \bibinfo{author}{Hu, J.F.},
  \bibinfo{author}{Zheng, W.S.}, \bibinfo{author}{Sebe, N.},
  \bibinfo{year}{2026}.
\newblock \bibinfo{title}{Risk-aware world model predictive control for
  generalizable end-to-end autonomous driving}.
\newblock \bibinfo{journal}{arXiv preprint arXiv:2602.23259} .
\bibitem[{Toromanoff et~al.(2020)Toromanoff, Wirbel and
  Moutarde}]{toromanoff2020affordances}
\bibinfo{author}{Toromanoff, M.}, \bibinfo{author}{Wirbel, E.},
  \bibinfo{author}{Moutarde, F.}, \bibinfo{year}{2020}.
\newblock \bibinfo{title}{End-to-end model-free reinforcement learning for
  urban driving using implicit affordances}, in: \bibinfo{booktitle}{IEEE/CVF
  Conference on Computer Vision and Pattern Recognition (CVPR)}, pp.
  \bibinfo{pages}{7151--7160}.
\newblock \DOIprefix\doi{10.1109/CVPR42600.2020.00718}.
\bibitem[{Vaswani et~al.(2017)Vaswani, Shazeer, Parmar, Uszkoreit, Jones,
  Gomez, Kaiser and Polosukhin}]{vaswani2017attention}
\bibinfo{author}{Vaswani, A.}, \bibinfo{author}{Shazeer, N.},
  \bibinfo{author}{Parmar, N.}, \bibinfo{author}{Uszkoreit, J.},
  \bibinfo{author}{Jones, L.}, \bibinfo{author}{Gomez, A.N.},
  \bibinfo{author}{Kaiser, {\L}.}, \bibinfo{author}{Polosukhin, I.},
  \bibinfo{year}{2017}.
\newblock \bibinfo{title}{Attention is all you need}, in:
  \bibinfo{booktitle}{Advances in Neural Information Processing Systems
  (NeurIPS)}, pp. \bibinfo{pages}{5998--6008}.
\bibitem[{Wang et~al.(2025)Wang, Ye, Tao, Pan, Mallik, Yaman, Ren and
  Zhang}]{wang2025adawm}
\bibinfo{author}{Wang, H.}, \bibinfo{author}{Ye, X.}, \bibinfo{author}{Tao,
  F.}, \bibinfo{author}{Pan, C.}, \bibinfo{author}{Mallik, A.},
  \bibinfo{author}{Yaman, B.}, \bibinfo{author}{Ren, L.},
  \bibinfo{author}{Zhang, J.}, \bibinfo{year}{2025}.
\newblock \bibinfo{title}{{AdaWM}: Adaptive world model based planning for
  autonomous driving}, in: \bibinfo{booktitle}{International Conference on
  Learning Representations (ICLR)}, pp. \bibinfo{pages}{85591--85615}.
\bibitem[{Wang et~al.(2021)Wang, Pun, Tu, Manivasagam, Sadat, Casas, Ren and
  Urtasun}]{wang2021advsim}
\bibinfo{author}{Wang, J.}, \bibinfo{author}{Pun, A.}, \bibinfo{author}{Tu,
  J.}, \bibinfo{author}{Manivasagam, S.}, \bibinfo{author}{Sadat, A.},
  \bibinfo{author}{Casas, S.}, \bibinfo{author}{Ren, M.},
  \bibinfo{author}{Urtasun, R.}, \bibinfo{year}{2021}.
\newblock \bibinfo{title}{{AdvSim}: Generating safety-critical scenarios for
  self-driving vehicles}, in: \bibinfo{booktitle}{IEEE/CVF Conference on
  Computer Vision and Pattern Recognition (CVPR)}, pp.
  \bibinfo{pages}{9904--9913}.
\newblock \DOIprefix\doi{10.1109/CVPR46437.2021.00978}.
\bibitem[{Wang et~al.(2024a)Wang, Zhu, Huang, Chen, Zhu and
  Lu}]{wang2023drivedreamer}
\bibinfo{author}{Wang, X.}, \bibinfo{author}{Zhu, Z.}, \bibinfo{author}{Huang,
  G.}, \bibinfo{author}{Chen, X.}, \bibinfo{author}{Zhu, J.},
  \bibinfo{author}{Lu, J.}, \bibinfo{year}{2024}a.
\newblock \bibinfo{title}{{DriveDreamer}: Towards real-world-drive world models
  for autonomous driving}, in: \bibinfo{booktitle}{European Conference on
  Computer Vision (ECCV)}, \bibinfo{publisher}{Springer}. pp.
  \bibinfo{pages}{55--72}.
\newblock \DOIprefix\doi{10.1007/978-3-031-73195-2_4}.
\bibitem[{Wang et~al.(2024b)Wang, He, Fan, Li, Chen and
  Zhang}]{wang2024drivewm}
\bibinfo{author}{Wang, Y.}, \bibinfo{author}{He, J.}, \bibinfo{author}{Fan,
  L.}, \bibinfo{author}{Li, H.}, \bibinfo{author}{Chen, Y.},
  \bibinfo{author}{Zhang, Z.}, \bibinfo{year}{2024}b.
\newblock \bibinfo{title}{Driving into the future: Multiview visual forecasting
  and planning with world model for autonomous driving}, in:
  \bibinfo{booktitle}{IEEE/CVF Conference on Computer Vision and Pattern
  Recognition (CVPR)}, pp. \bibinfo{pages}{14749--14759}.
\newblock \DOIprefix\doi{10.1109/CVPR52733.2024.01397}.
\bibitem[{Williams et~al.(2017)Williams, Wagener, Goldfain, Drews, Rehg, Boots
  and Theodorou}]{williams2017mppi}
\bibinfo{author}{Williams, G.}, \bibinfo{author}{Wagener, N.},
  \bibinfo{author}{Goldfain, B.}, \bibinfo{author}{Drews, P.},
  \bibinfo{author}{Rehg, J.M.}, \bibinfo{author}{Boots, B.},
  \bibinfo{author}{Theodorou, E.A.}, \bibinfo{year}{2017}.
\newblock \bibinfo{title}{Information theoretic {MPC} for model-based
  reinforcement learning}, in: \bibinfo{booktitle}{IEEE International
  Conference on Robotics and Automation (ICRA)}, pp.
  \bibinfo{pages}{1714--1721}.
\newblock \DOIprefix\doi{10.1109/ICRA.2017.7989202}.
\bibitem[{Wilson et~al.(2021)Wilson, Bonawitz, Costa and
  Ebitz}]{wilson2021balancing}
\bibinfo{author}{Wilson, R.C.}, \bibinfo{author}{Bonawitz, E.},
  \bibinfo{author}{Costa, V.D.}, \bibinfo{author}{Ebitz, R.B.},
  \bibinfo{year}{2021}.
\newblock \bibinfo{title}{Balancing exploration and exploitation with
  information and randomization}.
\newblock \bibinfo{journal}{Current Opinion in Behavioral Sciences}
  \bibinfo{volume}{38}, \bibinfo{pages}{49--56}.
\newblock \DOIprefix\doi{10.1016/j.cobeha.2020.10.001}.
\bibitem[{Wu et~al.(2022)Wu, Jia, Chen, Yan, Li and Qiao}]{wu2022tcp}
\bibinfo{author}{Wu, P.}, \bibinfo{author}{Jia, X.}, \bibinfo{author}{Chen,
  L.}, \bibinfo{author}{Yan, J.}, \bibinfo{author}{Li, H.},
  \bibinfo{author}{Qiao, Y.}, \bibinfo{year}{2022}.
\newblock \bibinfo{title}{Trajectory-guided control prediction for end-to-end
  autonomous driving: A simple yet strong baseline}, in:
  \bibinfo{booktitle}{Advances in Neural Information Processing Systems
  (NeurIPS)}, pp. \bibinfo{pages}{6119--6132}.
\newblock \DOIprefix\doi{10.52202/068431-0443}.
\bibitem[{Yin et~al.(2023)Yin, Zhang and Tsiotras}]{yin2023cvarmppi}
\bibinfo{author}{Yin, J.}, \bibinfo{author}{Zhang, Z.},
  \bibinfo{author}{Tsiotras, P.}, \bibinfo{year}{2023}.
\newblock \bibinfo{title}{Risk-aware model predictive path integral control
  using conditional value-at-risk}, in: \bibinfo{booktitle}{IEEE International
  Conference on Robotics and Automation (ICRA)}, pp.
  \bibinfo{pages}{7937--7943}.
\newblock \DOIprefix\doi{10.1109/ICRA48891.2023.10161100}.
\bibitem[{Yu et~al.(2020)Yu, Thomas, Yu, Ermon, Zou, Levine, Finn and
  Ma}]{yu2020mopo}
\bibinfo{author}{Yu, T.}, \bibinfo{author}{Thomas, G.}, \bibinfo{author}{Yu,
  L.}, \bibinfo{author}{Ermon, S.}, \bibinfo{author}{Zou, J.},
  \bibinfo{author}{Levine, S.}, \bibinfo{author}{Finn, C.},
  \bibinfo{author}{Ma, T.}, \bibinfo{year}{2020}.
\newblock \bibinfo{title}{{MOPO}: Model-based offline policy optimization}, in:
  \bibinfo{booktitle}{Advances in Neural Information Processing Systems
  (NeurIPS)}, pp. \bibinfo{pages}{14129--14142}.
\bibitem[{Zhang et~al.(2026)Zhang, Cui, Xu, Ye, Zhang and
  Wang}]{zhang2026counterfactual}
\bibinfo{author}{Zhang, J.}, \bibinfo{author}{Cui, C.}, \bibinfo{author}{Xu,
  Y.}, \bibinfo{author}{Ye, X.}, \bibinfo{author}{Zhang, R.},
  \bibinfo{author}{Wang, Z.}, \bibinfo{year}{2026}.
\newblock \bibinfo{title}{How can driving world models do counterfactual
  prediction?}
\newblock \bibinfo{journal}{arXiv preprint arXiv:2608.11601} .
\bibitem[{Zhang et~al.(2023)Zhang, Peng, Li and Zhou}]{zhang2023cat}
\bibinfo{author}{Zhang, L.}, \bibinfo{author}{Peng, Z.}, \bibinfo{author}{Li,
  Q.}, \bibinfo{author}{Zhou, B.}, \bibinfo{year}{2023}.
\newblock \bibinfo{title}{{CAT}: Closed-loop adversarial training for safe
  end-to-end driving}, in: \bibinfo{booktitle}{Proceedings of the 7th
  Conference on Robot Learning (CoRL)}, pp. \bibinfo{pages}{2357--2372}.
\bibitem[{Zhang et~al.(2022)Zhang, Tang, Geng, Chen, Xin and
  Wang}]{zhang2022mmfn}
\bibinfo{author}{Zhang, Q.}, \bibinfo{author}{Tang, M.}, \bibinfo{author}{Geng,
  R.}, \bibinfo{author}{Chen, F.}, \bibinfo{author}{Xin, R.},
  \bibinfo{author}{Wang, L.}, \bibinfo{year}{2022}.
\newblock \bibinfo{title}{{MMFN}: Multi-modal-fusion-net for end-to-end
  driving}, in: \bibinfo{booktitle}{IEEE/RSJ International Conference on
  Intelligent Robots and Systems (IROS)}, pp. \bibinfo{pages}{8638--8643}.
\newblock \DOIprefix\doi{10.1109/IROS47612.2022.9981775}.
\bibitem[{Zhang et~al.(2021)Zhang, Liniger, Dai, Yu and
  Van~Gool}]{zhang2021roach}
\bibinfo{author}{Zhang, Z.}, \bibinfo{author}{Liniger, A.},
  \bibinfo{author}{Dai, D.}, \bibinfo{author}{Yu, F.},
  \bibinfo{author}{Van~Gool, L.}, \bibinfo{year}{2021}.
\newblock \bibinfo{title}{End-to-end urban driving by imitating a reinforcement
  learning coach}, in: \bibinfo{booktitle}{IEEE/CVF International Conference on
  Computer Vision (ICCV)}, pp. \bibinfo{pages}{15202--15212}.
\newblock \DOIprefix\doi{10.1109/ICCV48922.2021.01494}.
\bibitem[{Zheng et~al.(2025)Zheng, Yang, Xing, Zhang, Zheng, Gao, Li, Zhang,
  Xia, Jia, Lang and Zhao}]{zheng2025world4drive}
\bibinfo{author}{Zheng, Y.}, \bibinfo{author}{Yang, P.}, \bibinfo{author}{Xing,
  Z.}, \bibinfo{author}{Zhang, Q.}, \bibinfo{author}{Zheng, Y.},
  \bibinfo{author}{Gao, Y.}, \bibinfo{author}{Li, P.}, \bibinfo{author}{Zhang,
  T.}, \bibinfo{author}{Xia, Z.}, \bibinfo{author}{Jia, P.},
  \bibinfo{author}{Lang, X.}, \bibinfo{author}{Zhao, D.}, \bibinfo{year}{2025}.
\newblock \bibinfo{title}{{World4Drive}: End-to-end autonomous driving via
  intention-aware physical latent world model}, in:
  \bibinfo{booktitle}{IEEE/CVF International Conference on Computer Vision
  (ICCV)}, pp. \bibinfo{pages}{28632--28642}.
\newblock \DOIprefix\doi{10.1109/ICCV51701.2025.02659}.

\end{thebibliography}

\end{document}